\documentclass[singlecolumn]{ModernAcademic-EN}

\usepackage{makecell}
\usepackage{pifont}
\usepackage{float}

\graphicspath{{./}}

\newcolumntype{T}[1]{>{\raggedright\arraybackslash}p{#1}}
\newcolumntype{M}[1]{>{\centering\arraybackslash}m{#1}}
\newcolumntype{N}[1]{>{\raggedright\arraybackslash}m{#1}}
\newcommand{\cmark}{\textcolor{green!60!black}{\ding{51}}}
\newcommand{\xmark}{\textcolor{red}{\ding{55}}}
\newcommand{\bench}{Evo-DexBench}

\setthemecolor{1a2a6c}
\setlinkcolor{1a2a6c}
\setcitecolor{1a2a6c}
\seturlcolor{1a2a6c}
\settitlepreskip{0pt}
\makeatletter
\renewcommand{\MA@title@font}{\LARGE\bfseries\sffamily}
\renewcommand{\@maketitleblock}{%
  \begin{center}
    {\MA@title@font\color{academicblue}\@title\par}
    \vspace{\MA@title@author@skip}
    {\MA@author@font\@author\par}
    \ifx\@affiliation\@empty\else
      \vspace{\MA@author@affiliation@skip}
      {\MA@affiliation@font\@affiliation\par}%
    \fi
    \ifx\@email\@empty\else
      \vspace{\MA@affiliation@email@skip}
      {\MA@email@font\@email\par}%
    \fi
  \end{center}
}
\makeatother
\hidelogoinheader
\hidelogointitle
\title{Evo-DexBench: Benchmarking Generalization across Related Tasks for Dexterous Manipulation}
\author{Bing Cheng\textsuperscript{1,*}, Yiran Mao\textsuperscript{1,2,*}, Zewei Ye\textsuperscript{1,*}, Yifan Jin\textsuperscript{1,3}, Xinyi Chen\textsuperscript{1}, Ruifeng Tu\textsuperscript{1}, Shuhuai Wang\textsuperscript{1,4}, Tao Lin\textsuperscript{1}, Zhimu Zhou\textsuperscript{1}, Yuxin Du\textsuperscript{1}, Fangwei Zhong\textsuperscript{5}, Junchi Yan\textsuperscript{1},\\
Bo Zhao\textsuperscript{1,\ensuremath{\dagger}}}
\shorttitle{Evo-DexBench}
\shortauthor{Cheng et~al.}
\affiliation{\footnotesize\textsuperscript{1}School of AI, Shanghai Jiao Tong University; \textsuperscript{2}Nanjing University; \textsuperscript{3}Central South University\\
\textsuperscript{4}Beijing University of Posts and Telecommunications; \textsuperscript{5}Beijing Normal University}
\email{\textsuperscript{*}Equal contribution; \textsuperscript{\ensuremath{\dagger}}Corresponding author\\[0.35em]
{Contact:  sjtu8620588@sjtu.edu.cn; bo.zhao@sjtu.edu.cn}\\[0.35em]
\normalfont\large
\href{https://mint-sjtu.github.io/Evo-DexBench.io/}{ \textcolor{urlpurple}{[Project Page]}}\hspace{1em}
\href{https://huggingface.co/collections/MINT-SJTU/evo-dexbench}{ \textcolor{urlpurple}{[Hugging Face]}}\hspace{1em}
\href{https://github.com/MINT-SJTU/Evo-DexBench}{ \textcolor{urlpurple}{[GitHub]}}}
\begin{document}
\maketitle

\begin{figure}[H]
    \centering
    \includegraphics[width=\linewidth]{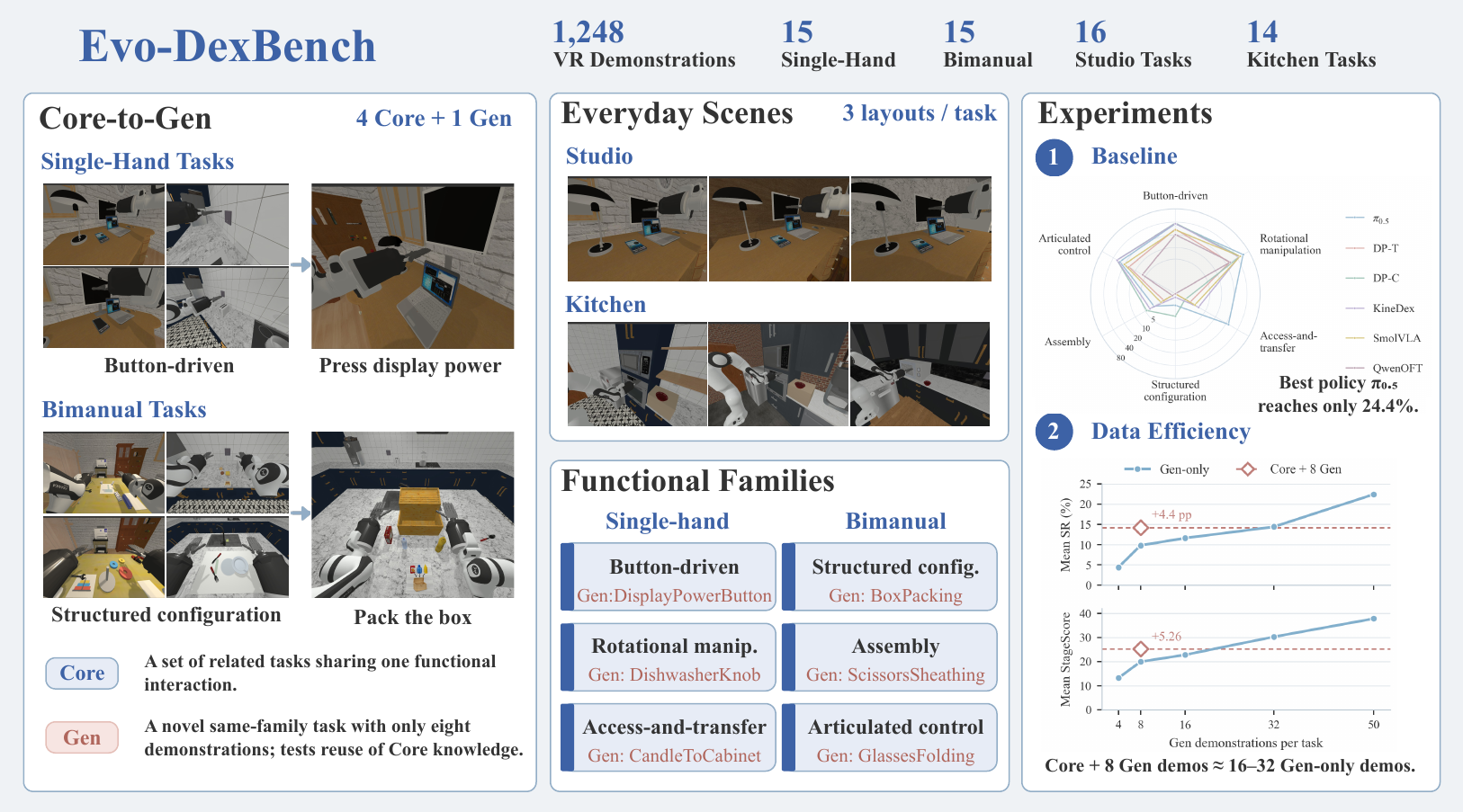}
    \caption{\textbf{Overview of Evo-DexBench.} The benchmark comprises six functional families, each containing four Core tasks and one Gen task. For each task, one of the three layouts in its scene is sampled at random. The radar plot shows each baseline's mean success rate per family. The data-efficiency study uses $\pi_{0.5}$, where the red dashed line denotes joint training with Core data and 8 Gen demonstrations, and the blue curve denotes Gen-only training.}
    \label{fig:evo-dexbench-overview}
\end{figure}

\begin{tcolorbox}[
  enhanced,
  breakable,
  width=\textwidth,
  center,
  colback=abstractbg,
  colframe=abstractbg,
  boxrule=0pt,
  sharp corners,
  toptitle=8pt,
  bottomtitle=3pt,
  left=12pt,
  right=12pt,
  top=5pt,
  bottom=10pt,
  fontupper=\small,
  before skip=1em,
  after skip=1.5em,
  title={Abstract},
  fonttitle=\bfseries,
  coltitle=academicblue,
  halign title=center
]
\noindent General-purpose dexterous manipulation requires policies to transfer knowledge from related tasks to new tasks with only a few target demonstrations. Existing benchmarks often provide too many target demonstrations and leave the differences between source and target tasks uncontrolled. As a result, they cannot clearly distinguish knowledge transfer from task memorization. Our key insight is the \textbf{Core-Gen principle}. \textbf{Core tasks} are related tasks that share a functional interaction and provide source knowledge, while a \textbf{Gen task} is a novel target within the same functional family that has only a few demonstrations. This setup tests whether a policy can reuse functional interaction knowledge. Based on this principle, we propose \textbf{\bench{}}, which controls both source-task relatedness and target-task exposure. It comprises 30 tasks in six functional families, three everyday layouts per task, and 1,248 multimodal demonstrations collected via VR teleoperation with simulated Shadow Hands equipped with 123 taxels. We first compare few-shot adaptation strategies and find that joint Core + Gen training makes the most effective use of Core knowledge. Under this setting, we evaluate six representative policies. The best-performing policy, $\pi_{0.5}$, achieves a mean success rate of only 24.4\%, revealing the limitations of existing methods in transferring functional interaction knowledge. Further analysis shows that on the tested Gen targets, functionally matched Core data yields larger success-rate gains than mismatched data, and that with matched Core data, 8 Gen demonstrations reach the performance of training on 16--32 Gen demonstrations alone. These results indicate that \bench{} separates knowledge transfer from task memorization under controlled conditions and effectively evaluates the generalization of dexterous manipulation policies across related tasks.
\end{tcolorbox}

\section{Introduction}
\label{sec:introduction}

General-purpose dexterous manipulation requires policies to extract functional interactions shared across existing tasks and transfer them to new tasks given only a few target demonstrations~\citep{fu2026survey,tsuji2025survey,an2025dexterous}. Many tasks use different objects and scenes but share the same functional interaction. For example, turning a knob, opening a bottle cap, and twisting a tool all require a stable grasp and proper contact. A policy should not merely memorize specific objects, trajectories, or instructions. Dexterous hands make this requirement more pronounced: Their many degrees of freedom and multi-finger coordination~\citep{li2022survey} make demonstrations costly to collect, so gathering ample data for every new task is impractical. Moreover, even tasks that share a functional interaction require object-specific finger configurations and contact relationships~\citep{zhao2026towards,wang2026dexjoco}, so replaying memorized trajectories often fails. Hence, policies must transfer the functional interaction itself: which component to engage, how to contact it, and how to move under its motion constraints.

Recent progress in Vision-Language-Action (VLA) models~\citep{physicalintelligence2026pi07,luo2025beingh0,zheng2026egoscale,yuan2026qwen} and World Action Models (WAMs)~\citep{ye2026dreamzero,li2026causal,liao2026genie} suggests that policies are increasingly able to reuse functional interactions from multi-task demonstrations. However, current evaluations~\citep{liu2023libero,yao2026dexverse,yu2020meta} often struggle to assess this transfer ability: with too many target-task demonstrations, a model may rely on task memorization; with large differences between source and target tasks, it becomes difficult to determine whether performance differences result from functional interaction transfer or from other factors.

To address these issues, we propose the \textbf{Core-Gen principle.} Core tasks are a set of related tasks that share the same type of functional interaction, and they provide transferable knowledge for Gen tasks. Gen tasks are new related tasks with only a few demonstrations, and they test whether a policy can reuse the functional interaction knowledge it has learned. Based on this principle, we propose \bench{}, a benchmark that controls both the functional relatedness between source and target tasks and the target task's direct exposure. This lets us test whether a policy truly extracts functional knowledge from related source tasks, and whether experience with more related tasks leads to greater transfer gains.

Our contributions can be summarized as:
\begin{itemize}
    \item We introduce \bench{}, a benchmark based on the Core-Gen principle. It evaluates how well dexterous policies generalize across related tasks. Each evaluation unit has a Core task set and a Gen task that share functional interaction. Only a few target demonstrations are provided for the Gen task, allowing us to test whether a policy can transfer interaction knowledge to a new task.

    \item We build an integrated simulation environment and a VR teleoperation data collection pipeline. The environment uses simulated Shadow Hands with 123 taxels each and provides three everyday layouts per task. With this pipeline, we collect 1,248 multimodal demonstrations that include tactile signals.

    \item We evaluate six representative policies on \bench{}. Under joint Core + Gen training, the best-performing policy, $\pi_{0.5}$, achieves a mean success rate of only 24.4\%, showing that current methods still struggle to reuse functional interaction knowledge. We further analyze the role of Core data in transfer: functionally matched Core data consistently yields larger gains than mismatched data and allows 8 Gen demonstrations to reach the performance of 16--32 demonstrations.
\end{itemize}

\section{Related Work}
\label{sec:related}

\begin{table}[tb]
\centering
\small
\setlength{\tabcolsep}{0pt}
\caption{Comparison of embodied manipulation benchmarks.}
\label{tab:benchmark-comparison}
\begin{tabular*}{\linewidth}{@{\extracolsep{\fill}}lcccccc@{}}
\toprule
Benchmark & Arm--hand & Bimanual & Tactile & Layouts/task & Core-Gen & Data Collection \\
\midrule
LIBERO~\citep{liu2023libero} & \xmark & \xmark & \xmark & 1 & \xmark & SpaceMouse teleoperation \\
RoboTwin 2.0~\citep{chen2025robotwin} & \xmark & \cmark & \xmark & 1 & \xmark & MLLM-based generation \\
Bi-DexHands~\citep{chen2022bidexhands} & \xmark & \cmark & \xmark & 1 & \xmark & RL policy \\
DexArt~\citep{bao2023dexart} & \cmark & \xmark & \xmark & 1 & \xmark & RL policy \\
DexCraft~\citep{yuan2026dexcraft} & \cmark & \xmark & \xmark & 1 & \xmark & Keyboard teleoperation \\
HANDFUL~\citep{foong2026handful} & \cmark & \xmark & \xmark & 1 & \xmark & RL policy \\
DexJoCo~\citep{wang2026dexjoco} & \cmark & \cmark & \xmark & 1 & \xmark & Motion-capture teleoperation \\
DexVerse~\citep{yao2026dexverse} & \cmark & \cmark & Optional & 1 & \xmark & VR teleoperation \\
\midrule
\textbf{Evo-DexBench} & \cmark & \cmark & 123 taxels & 3 & \cmark & VR teleoperation \\
\bottomrule
\end{tabular*}
\end{table}

\paragraph{Evaluation for Generalization.}
Existing manipulation benchmarks mainly assess generalization to unseen environments~\citep{mees2022calvin,zhang2025vlabench,chen2025robotwin}. Others evaluate policies on held-out tasks, such as the ML10/ML45 splits of Meta-World~\citep{yu2020meta} and the ML4 split of Bi-DexHands~\citep{chen2022bidexhands}, or on knowledge-partitioned task suites in LIBERO~\citep{liu2023libero}. However, such splits merely designate a subset of tasks as targets, while the relatedness between source and target tasks is neither specified nor adjustable, so a transfer gain cannot be attributed to a shared functional interaction. The number of target demonstrations is likewise fixed at one of two extremes, either none at all, with adaptation from rewards alone via meta-reinforcement learning, or abundant demonstrations (50 per task in LIBERO), making it hard to tell whether a policy reuses prior knowledge or simply fits the target task. A further line of work learns reusable interaction structures, such as Where2Act~\citep{mo2021where2act} and VAT-Mart~\citep{wu2022vatmart}, but only for individual articulated objects.

\paragraph{Dexterous Manipulation Benchmarks.}
Table~\ref{tab:benchmark-comparison} shows that benchmarks for dexterous manipulation cover functional grasping, manipulation of unseen objects and bimanual coordination~\citep{bao2023dexart,foong2026handful}. DexCraft~\citep{yuan2026dexcraft} scales grasp pretraining toward tool-oriented manipulation; DexJoCo~\citep{wang2026dexjoco} targets task-oriented functional manipulation; and DexVerse~\citep{yao2026dexverse} emphasizes multi-task, multi-embodiment operation. These efforts extend evaluation beyond single-step grasping, but they typically ask whether a policy can complete predefined tasks, rather than whether contact-level interaction knowledge transfers across tasks.

\paragraph{Dexterous Manipulation Policies.}
Prior work has advanced vision-based dexterous manipulation and generalization to unseen objects and scenes~\citep{rajeswaran2017dapg,andrychowicz2020shadowdexteroushand,qi2023general,xu2023unidexgrasp,wan2023unidexgrasp++,qin2023dexpoint,lum2024dextrah}, while other studies have scaled up human demonstration data and incorporated tactile information~\citep{qin2022dexmv,wang2024dexcap,zhang2025kinedexlearningtactileinformedvisuomotor}. More recent models such as DexVLA~\citep{wen2025dexvla}, DexGraspVLA~\citep{zhong2026dexgraspvla}, and $\pi_{0.5}$~\citep{intelligence2025pi05visionlanguageactionmodelopenworld} seek behavior priors across tasks, objects, and embodiments. As policy research moves toward extracting reusable knowledge, Evo-DexBench provides the corresponding evaluation, measuring whether a policy can transfer from multiple related source tasks to a novel target task given only a few demonstrations.

\section{Benchmark Design}
\label{sec:benchmark}
\begin{figure}[t]
    \centering
    \includegraphics[width=\linewidth]{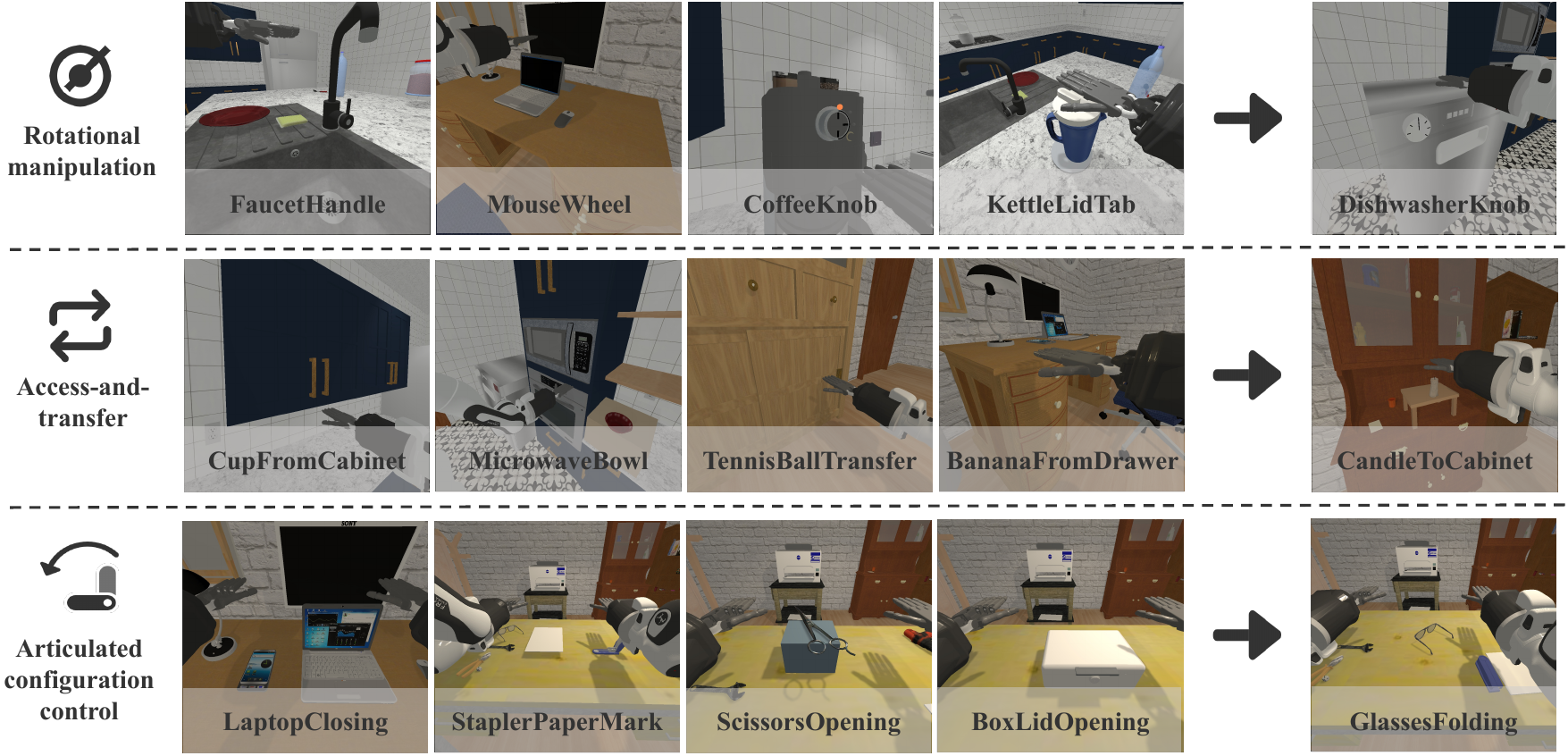}
    \caption{\textbf{Core-Gen task examples.}  The first and second rows show single-hand Rotational manipulation and Access-and-transfer tasks, respectively, while the third row shows bimanual articulated configuration control tasks. The first four columns show Core tasks, and the last column shows Gen tasks.}
    \label{fig:core-gen}
\end{figure}

Unlike prior benchmarks that largely measure completion of individual tasks, Evo-DexBench evaluates whether a policy can distill the functional interaction knowledge shared across \emph{Core} tasks and transfer it to a \emph{Gen} task given only a few demonstrations. Section~\ref{sec:design_principles} introduces the Core-Gen design principle and task composition, Section~\ref{sec:implementation} details the shared environment, observation, and tactile interfaces used for collection and evaluation, and Section~\ref{sec:data_collection} presents the demonstration-collection pipeline.

\subsection{Design Principles and Task Composition}
\label{sec:design_principles}

To systematically assess whether models can learn reusable knowledge from functionally related tasks and transfer it to a target task with only a few target demonstrations, we propose the Core-Gen design principle.
As shown in Figure~\ref{fig:core-gen}, the evaluation unit in Evo-DexBench is a Core-Gen task pair. Core tasks within a family instantiate the same functional relation across different objects and motion conditions, creating diverse source tasks from which a policy can learn stable functional relations rather than rely on a single object. Gen extends this relation to a new object or usage context. Because Gen provides only eight demonstrations, a policy cannot succeed by memorizing target-specific trajectories. It must extract shared functional interaction from Core and transfer it to Gen.
For example, Core may include pressing a desk-lamp button and a smartphone's home button, while Gen requires transferring this interaction to pressing a monitor power button.

\begin{figure}[t]
    \centering
    \includegraphics[width=\linewidth]{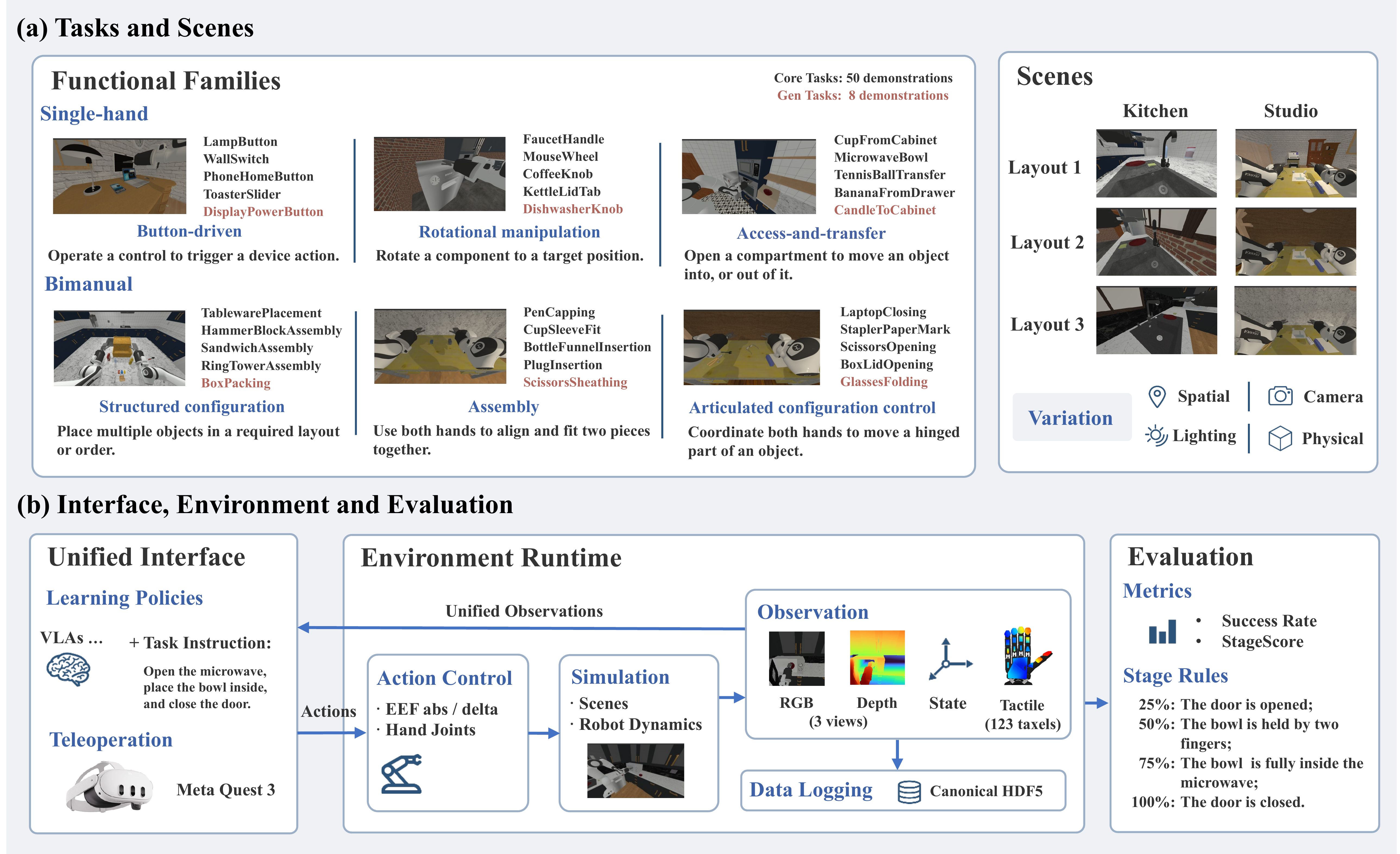}
    \caption{\textbf{Evo-DexBench framework.} (a): Core (black) and Gen (red) tasks in each family share one functional interaction, and task logic is decoupled from scene layout. (b) Using the MicrowaveBowl task as an example, policies and VR teleoperation share observation and action spaces, so demonstrations are directly usable for training. RGB is captured by the context, interaction, and wrist cameras.}
    \label{fig:dexbench-overview}
\end{figure}

Formally, each task is defined as
\begin{equation}
  T = (\Omega, S_0, O, A, G),
  \label{eq:task}
\end{equation}
where $\Omega$ denotes the objects and actuated components involved in the task, $S_0$ the initial environment state, $O$ and $A$ the observations available to and actions executed by the policy, and $G$ the task goal.
We group tasks by their primary functional interaction, i.e., the manipulated component, contact mode, motion constraint, and action order required to achieve $G$.
For example, the FaucetHandle and DishwasherKnob tasks involve different objects, but both require applying directed rotation to an actuated component, so they are grouped into the same family.

Following this grouping, Evo-DexBench organizes its 30 tasks into six functional families, each with four Core tasks and one Gen task.
As summarized in Figure~\ref{fig:dexbench-overview}(a), the six families cover Button-driven, Rotational manipulation, Access-and-transfer, Structured configuration, Assembly, and Articulated configuration control. The first three use single-hand setups, while the last three use bimanual setups. Access-and-transfer and Structured configuration tasks are long-horizon, involving multiple sequential stages (Figure~\ref{fig:task-visualization}(b)); Assembly requires bimanual coordination to precisely align two complementary parts for insertion or fitting (e.g., inserting a plug).

\subsection{Environment and Sensing Interfaces}
\label{sec:implementation}

\paragraph{Composable Environment Design.}
As shown in Figure~\ref{fig:dexbench-overview}, Evo-DexBench uses a composable environment design: task logic is reusable across layouts; scenes are composable across tasks. Benchmark scenes are organized into two categories, Studio and Kitchen, each with three layouts that vary furniture arrangement, relative object placement, and visual background. Tasks are randomly executed across everyday environments rather than on a fixed workbench, preventing policies from relying on location or appearance cues tied to one layout. Policies must instead understand task requirements and adapt to environmental changes.

\paragraph{Scene Assets.}
Scene assets come from four sources: YCB~\citep{calli2015ycb} provides standardized everyday rigid objects; PartNet-Mobility~\citep{xiang2020sapien} provides articulated devices with realistic joint structures; RoboCasa~\citep{nasiriany2024robocasa} provides semantically meaningful kitchen scenes; and custom assets meet task-specific appearance or functional requirements.

\paragraph{Visual Observations and Tactile.}
Visual observations come from three camera types. The context camera observes the overall environment, the interaction camera focuses on the task workspace, and the wrist camera provides local detail around the hand. The context and interaction cameras are manually calibrated for each task.

In dexterous manipulation, contacts are often occluded by the hand itself, making fine-grained contact states hard to infer from vision alone.
We therefore equip each Shadow Hand with distributed normal-force sensing: each hand outputs a 123-dimensional normal-force signal, with taxels allocated across fingers according to their contact area and functional role in manipulation.

\subsection{Data Collection Pipeline}
\label{sec:data_collection}

\begin{figure}[t]
    \centering
    \includegraphics[width=\linewidth]{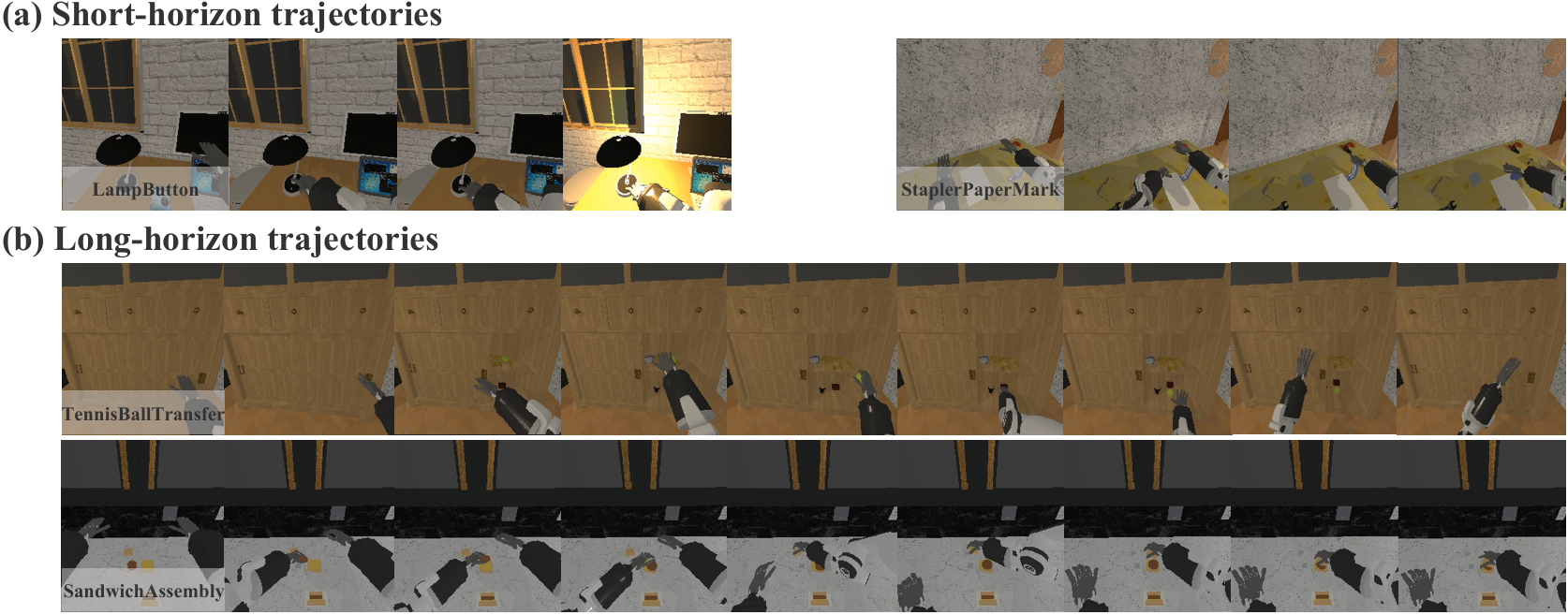}
    \caption{\textbf{Task visualizations.} (a) shows short-horizon tasks: pressing a desk lamp button (left) and stapling paper (right). (b) shows long-horizon tasks: opening a cabinet, placing a tennis ball on the lower shelf, and closing the cabinet; and assembling a sandwich.}
    \label{fig:task-visualization}
\end{figure}

\paragraph{VR Teleoperation.}
Evo-DexBench uses Quest 3 WebXR hand tracking to collect dexterous manipulation demonstrations in simulation. We map the operator's wrist motion to the Panda end-effector and retarget hand keypoints to Shadow Hand joint targets~\citep{qin2023anyteleop}.

\paragraph{Data Collection.}
The demonstrations cover 30 tasks across six functional families, and each demonstration is collected in a layout randomly sampled from the three layouts of its task, reducing dependence on any single scene structure. The released dataset contains 1,248 successful demonstrations, each Core task contributes fifty trajectories and each Gen task contributes eight. Figure~\ref{fig:task-visualization} shows four trajectories. The dataset is provided in two formats: Canonical HDF5, which preserves complete multimodal information, and LeRobot v3.0, which supports direct policy training.

\section{Experiments}
\label{sec:experiments}
\providecommand{\Score}[2]{\makebox[2.8em][r]{#1}\,/\,\makebox[2.4em][l]{#2}}

\begin{figure}[t]
    \centering
    \includegraphics[width=\linewidth]{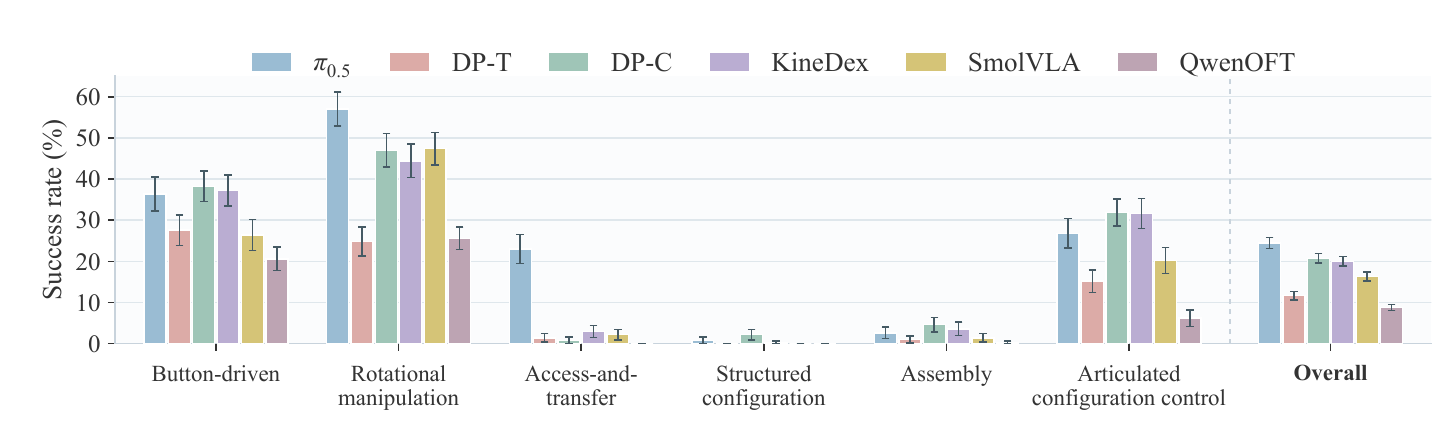}
    \caption{\textbf{Baseline SR across six task families and all 30 tasks (Overall) with 95\% confidence intervals.} Each family average includes four Core tasks and one Gen task.}
    \label{fig:baseline_performance}
\end{figure}

Our main results benchmark six representative policies on Evo-DexBench (Section~\ref{sec:baseline_results}) under the joint Core + Gen training protocol selected in Section~\ref{sec:fewshot_strategy}. We then validate the Core-Gen principle through two analyses: whether Core data from the same functional family transfers better than Core data from other families (Section~\ref{sec:transfer_results}), and how many additional Gen demonstrations can be replaced by matched Core data (Section~\ref{sec:target_data_efficiency}). Finally, we evaluate the contribution of tactile sensing in KineDex (Section~\ref{sec:touch_experiment}).

\begin{table}[t]
    \centering
    \caption{\textbf{Baseline performance by task family.} Each entry reports SR (\%) / StageScore (0--100). Each Core row averages four tasks, and each Gen row reports the target task after joint training with Core data from the same family. The overall rows average the 24 Core tasks and 6 Gen tasks separately. Bold marks the best value in each row, including ties. Red parentheses below the $\pi_{0.5}$ Gen results show the changes relative to Gen-only training.}
    \label{tab:baseline_results}

    \normalsize
    \setlength{\tabcolsep}{1.5pt}
    \renewcommand{\arraystretch}{1.12}
    \definecolor{genrow}{RGB}{235,245,253}
    \definecolor{deltared}{RGB}{170,25,36}

    \newcommand{\baselinefamily}[1]{%
        \multicolumn{1}{@{}l}{%
            \cellcolor{white}\multirow{-2}{*}{\shortstack[l]{#1}}%
        }%
    }

    \newcommand{\baselineresult}[3][]{%
        \makebox[2.6em][r]{#2}%
        \if\relax\detokenize{#1}\relax\else
            \rlap{\makebox[0.55em][c]{%
                \raisebox{-6.5pt}[0pt][8.5pt]{%
                    \textcolor{deltared}{\normalfont\scriptsize(#1)}%
                }%
            }}%
        \fi
        \makebox[0.55em][c]{/}%
        \makebox[2.3em][l]{#3}%
    }

    \newcommand{\baselinegrid}{%
        \small
        \begin{tabular}{@{}ll*{6}{c}@{}}
            \toprule
            Task family & Split & $\pi_{0.5}$ & DP-T & DP-C
                & KineDex & SmolVLA & QwenOFT \\
            \midrule
            & Core
                & \baselineresult{40.5\%}{56.93}
                & \baselineresult{25.0\%}{40.45}
                & \baselineresult{\textbf{45.0\%}}{\textbf{57.08}}
                & \baselineresult{42.8\%}{54.38}
                & \baselineresult{28.5\%}{46.28}
                & \baselineresult{25.5\%}{41.70} \\
            \rowcolor{genrow}
            \baselinefamily{Button-driven} & Gen
                & \baselineresult[+9\%/+9.70]{20\%}{39.40}
                & \baselineresult{\textbf{38\%}}{\textbf{56.60}}
                & \baselineresult{11\%}{33.80}
                & \baselineresult{15\%}{39.60}
                & \baselineresult{18\%}{36.00}
                & \baselineresult{1\%}{7.00} \\
            \addlinespace[2pt]
            & Core
                & \baselineresult{\textbf{59.8\%}}{\textbf{73.50}}
                & \baselineresult{20.3\%}{46.24}
                & \baselineresult{42.0\%}{61.33}
                & \baselineresult{41.8\%}{60.24}
                & \baselineresult{48.8\%}{64.16}
                & \baselineresult{26.8\%}{38.53} \\
            \rowcolor{genrow}
            \baselinefamily{Rotational\\manipulation} & Gen
                & \baselineresult[+10\%/+7.40]{46\%}{62.50}
                & \baselineresult{43\%}{57.80}
                & \baselineresult{\textbf{67\%}}{\textbf{75.10}}
                & \baselineresult{55\%}{67.00}
                & \baselineresult{42\%}{57.30}
                & \baselineresult{21\%}{39.90} \\
            \addlinespace[2pt]
            & Core
                & \baselineresult{\textbf{28.0\%}}{\textbf{52.20}}
                & \baselineresult{1.8\%}{12.13}
                & \baselineresult{1.0\%}{15.89}
                & \baselineresult{3.8\%}{21.45}
                & \baselineresult{2.8\%}{16.31}
                & \baselineresult{0.0\%}{3.13} \\
            \rowcolor{genrow}
            \baselinefamily{Access-and-\\transfer} & Gen
                & \baselineresult[+3\%/+14.75]{\textbf{3\%}}{\textbf{19.25}}
                & \baselineresult{0\%}{0.75}
                & \baselineresult{0\%}{16.50}
                & \baselineresult{0\%}{3.25}
                & \baselineresult{0\%}{1.50}
                & \baselineresult{0\%}{0.25} \\
            \addlinespace[2pt]
            & Core
                & \baselineresult{1.0\%}{8.06}
                & \baselineresult{0.0\%}{2.70}
                & \baselineresult{\textbf{2.8\%}}{\textbf{13.38}}
                & \baselineresult{0.3\%}{4.24}
                & \baselineresult{0.0\%}{4.94}
                & \baselineresult{0.0\%}{0.00} \\
            \rowcolor{genrow}
            \baselinefamily{Structured\\configuration} & Gen
                & \baselineresult[+0\%/+0.40]{\textbf{0\%}}{\textbf{2.40}}
                & \baselineresult{\textbf{0\%}}{0.00}
                & \baselineresult{\textbf{0\%}}{0.40}
                & \baselineresult{\textbf{0\%}}{0.20}
                & \baselineresult{\textbf{0\%}}{2.20}
                & \baselineresult{\textbf{0\%}}{0.00} \\
            \addlinespace[2pt]
            & Core
                & \baselineresult{3.0\%}{27.73}
                & \baselineresult{1.0\%}{17.60}
                & \baselineresult{\textbf{5.3\%}}{27.95}
                & \baselineresult{4.5\%}{\textbf{30.24}}
                & \baselineresult{1.8\%}{21.23}
                & \baselineresult{0.0\%}{11.18} \\
            \rowcolor{genrow}
            \baselinefamily{Assembly} & Gen
                & \baselineresult[+0\%/+2.20]{1\%}{7.60}
                & \baselineresult{1\%}{11.60}
                & \baselineresult{\textbf{2\%}}{6.40}
                & \baselineresult{0\%}{7.00}
                & \baselineresult{0\%}{\textbf{14.00}}
                & \baselineresult{1\%}{8.20} \\
            \addlinespace[2pt]
            & Core
                & \baselineresult{32.5\%}{54.60}
                & \baselineresult{17.5\%}{42.65}
                & \baselineresult{38.3\%}{\textbf{60.03}}
                & \baselineresult{\textbf{38.5\%}}{58.45}
                & \baselineresult{25.3\%}{46.90}
                & \baselineresult{7.3\%}{33.55} \\
            \rowcolor{genrow}
            \baselinefamily{Articulated\\control} & Gen
                & \baselineresult[+3\%/+6.60]{4\%}{14.00}
                & \baselineresult{\textbf{6\%}}{16.80}
                & \baselineresult{\textbf{6\%}}{\textbf{20.40}}
                & \baselineresult{4\%}{16.40}
                & \baselineresult{0\%}{5.20}
                & \baselineresult{2\%}{10.00} \\
            \midrule
            & Core
                & \baselineresult{\textbf{27.5\%}}{\textbf{45.50}}
                & \baselineresult{10.9\%}{26.96}
                & \baselineresult{22.4\%}{39.27}
                & \baselineresult{21.9\%}{38.16}
                & \baselineresult{17.8\%}{33.30}
                & \baselineresult{9.9\%}{21.35} \\
            \rowcolor{genrow}
            \baselinefamily{\textbf{Overall}} & Gen
                & \baselineresult[+4.2\%/+6.84]{12.3\%}{24.19}
                & \baselineresult{\textbf{14.7\%}}{23.93}
                & \baselineresult{14.3\%}{\textbf{25.43}}
                & \baselineresult{12.3\%}{22.24}
                & \baselineresult{10.0\%}{19.37}
                & \baselineresult{4.2\%}{10.89} \\
            \bottomrule
        \end{tabular}%
    }
    \sbox{0}{\baselinegrid}%
    \setlength{\tabcolsep}{%
        \dimexpr\tabcolsep+(\linewidth-\wd0)/14\relax
    }%
    \baselinegrid
\end{table}

\subsection{Experimental Setup and Protocol}
\label{sec:setup}

\paragraph{Policies and Training Data.}
We compare $\pi_{0.5}$~\citep{intelligence2025pi05visionlanguageactionmodelopenworld}, convolutional and transformer-based Diffusion Policies (DP-C and DP-T)~\citep{chi2024diffusionpolicyvisuomotorpolicy}, KineDex~\citep{zhang2025kinedexlearningtactileinformedvisuomotor}, SmolVLA~\citep{shukor2025smolvlavisionlanguageactionmodelaffordable}, and QwenOFT (Qwen3-VL-4B backbone) implemented in StarVLA~\citep{community2026starvlalegolikecodebasevisionlanguageaction}. We adapt the released implementations to the benchmark observation and action interfaces. In the standard baseline setting, following Section~\ref{sec:fewshot_strategy}, we jointly train on 50 demonstrations from each of the four Core tasks and eight demonstrations from the Gen task in each family. Model-specific adaptations and evaluation settings are detailed in Appendix~\ref{app:training}.

\paragraph{Evaluation.}
Each task is evaluated over 100 episodes: 20 nominal episodes and 20 for each of four single-factor variations (camera, lighting, physical, and spatial). The three predefined scene layouts receive 34, 33, and 33 episodes, respectively.

\paragraph{Metrics.}
Let $y_e\in\{0,1\}$ denote the final task success of episode $e$, and let $k_e\in[0,1]$ denote the highest task-specific stage score credited in this episode. We report the success rate (SR) and StageScore:

\begin{equation}
\mathrm{SR}=\frac{1}{N}\sum_{e=1}^{N}y_e \times 100\%,
\qquad
\mathrm{StageScore}=\frac{100}{N}\sum_{e=1}^{N}k_e.
\label{eq}
\end{equation}

StageScore summarizes the highest progress score under the task-specific criteria. When aggregating either metric across tasks, we weight each task equally.

\subsection{Baseline Policy Performance}
\label{sec:baseline_results}

We benchmark six representative policies to establish reference performance on Evo-DexBench.
Figure~\ref{fig:baseline_performance} summarizes success rates by task family and across all 30 tasks, and Table~\ref{tab:baseline_results} reports Core and Gen results separately, with StageScore measuring partial progress. Overall, Evo-DexBench remains challenging for current policies. Across all tasks, the best-performing policy, $\pi_{0.5}$, achieves a success rate of only 24.4\% and a StageScore of 41.24. No policy exceeds a mean success rate of 15\% on Gen tasks.

The long-horizon Access-and-transfer and Structured configuration tasks are the hardest: except for $\pi_{0.5}$ on Access-and-transfer Core tasks (28.0\%), all policies achieve success rates below 4\%.
Their considerably higher StageScores indicate partial progress without task completion.
Success rates also remain low on Assembly tasks, which require precise alignment.

Policies achieve lower average success rates on bimanual tasks than on single-hand tasks, possibly because coordinating two arms and two dexterous hands enlarges the action space.

Compared with Gen-only training, adding Core demonstrations from the matched family improves $\pi_{0.5}$'s mean success rate by 4.2 percentage points and mean StageScore by 6.84 points.
Success-rate gains range from 0 to 10 percentage points across Gen tasks, suggesting that $\pi_{0.5}$ can benefit from related demonstrations but does not yet transfer functional interaction knowledge reliably across tasks.

\subsection{Few-Shot Adaptation with Core Data}
\label{sec:fewshot_strategy}

\begin{table}[ht]
    \caption{\textbf{Few-shot adaptation with $\pi_{0.5}$.} Results are reported at 50 epochs of Gen training as SR (\%) / StageScore (0--100). Bold marks the best result for each target.}
    \label{tab:fewshot_strategy}
    \centering
    \setlength{\tabcolsep}{5pt}
    \small
    \begin{tabular}{@{}lcc@{}}
        \toprule
        Strategy & DisplayPowerButton & DishwasherKnob \\
        \midrule
        Gen-only & \Score{11\%}{29.70} & \Score{36\%}{55.10} \\
        Core $\rightarrow$ Gen (final) & \Score{8\%}{23.60} & \Score{8\%}{29.30} \\
        Core $\rightarrow$ Gen (early) & \Score{9\%}{24.60} & \Score{22\%}{41.20} \\
        Core + Gen (joint) & \textbf{\Score{20\%}{39.40}} & \textbf{\Score{46\%}{62.50}} \\
        \bottomrule
    \end{tabular}
\end{table}

A central question in few-shot adaptation is how demonstrations from related manipulation tasks should be incorporated when only a few target demonstrations are available.

We compare four strategies to train $\pi_{0.5}$ on two Gen tasks with eight demonstrations per target (Table~\ref{tab:fewshot_strategy}): Gen-only training; sequential fine-tuning on the Gen demonstrations from the final (25 epochs) or an early (5 epochs) Core checkpoint; and joint training uses both Core and Gen demonstrations throughout.

Joint Core + Gen training performs best on both target tasks. This suggests that, under limited target supervision, retaining Core data during adaptation is more effective than using it only for initialization.
We therefore adopt joint training in all other experiments.

\subsection{Functional Transfer across Task Families}
\label{sec:transfer_results}

\begin{figure*}[ht]
    \centering
    \begin{minipage}[t]{0.46\textwidth}
        \centering
        \includegraphics[width=\linewidth]{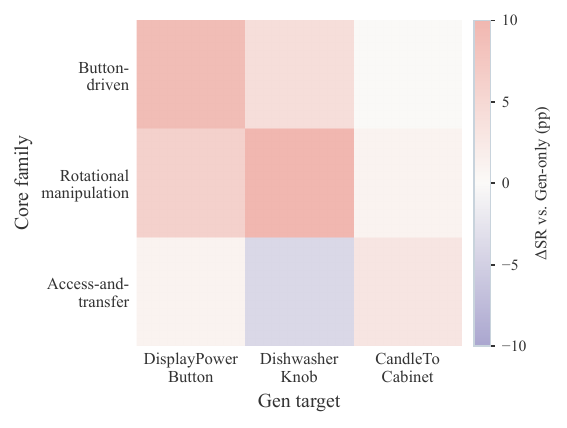}
    \end{minipage}\hfill
    \begin{minipage}[t]{0.52\textwidth}
        \centering
        \includegraphics[width=\linewidth]{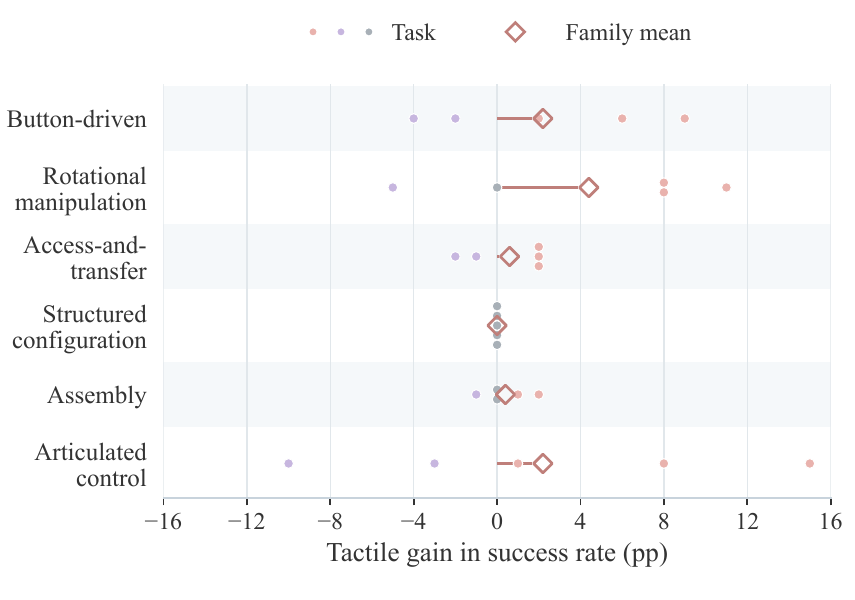}
    \end{minipage}
    \caption{\textbf{Transfer heatmap and tactile ablation in KineDex.} \textbf{Left:} Each cell reports the success rate changes of joint training with Core data from the indicated family in $\pi_{0.5}$, compared with Gen-only training. The diagonal pairs each Gen target with Core data from its own family. \textbf{Right:} The plot shows the success rate gains from the No-Touch variants to full-tactile KineDex (Full minus No-Touch). Circles show individual tasks, and diamonds show task-family means.
    }
    \label{fig:baseline_coregen}
    \label{fig:tactile_ablation}
\end{figure*}

We first examine whether functional relatedness between source and target tasks predicts transfer quality.
For each of the three Gen targets in Figure~\ref{fig:baseline_coregen}, we vary the source family while keeping the Gen demonstrations and training settings fixed. For $\pi_{0.5}$, the matching family yields the largest success rate gain over Gen-only training on all three targets.

\begin{table}[ht]
    \caption{\textbf{Transfer to DishwasherKnob across three policies.} Each cell reports SR (\%) / StageScore, with changes relative to Gen-only training below. Red indicates a gain and blue indicates a drop. Bold marks the best value in each row.}
    \label{tab:coregen_comparison}
    \centering
    \setlength{\tabcolsep}{3pt}
    \definecolor{coregengain}{RGB}{170,25,36}
    \definecolor{coregenloss}{RGB}{35,75,155}
    \newcommand{\CoreGenGain}[1]{\textcolor{coregengain}{#1}}
    \newcommand{\CoreGenLoss}[1]{\textcolor{coregenloss}{#1}}
    \newcommand{\CoreGenCell}[3]{%
        \ensuremath{%
            \vcenter{%
                \offinterlineskip
                \halign{\hfil##\hfil\cr
                    \strut #1\cr
                    \noalign{\kern 1pt}%
                    {\scriptsize(#2 / #3)}\cr
                }%
            }%
        }%
    }
    \small
    \begin{tabular}{@{}lccc@{}}
        \toprule
        Policy & \makecell{Button-driven}
               & \makecell{Rotational manipulation}
               & \makecell{Access-and-transfer} \\
        \midrule
        $\pi_{0.5}$
            & \CoreGenCell{\Score{40\%}{61.00}}
                {\CoreGenGain{+4\%}}{\CoreGenGain{+5.90}}
            & \CoreGenCell{\textbf{\Score{46\%}{62.50}}}
                {\CoreGenGain{+10\%}}{\CoreGenGain{+7.40}}
            & \CoreGenCell{\Score{32\%}{52.40}}
                {\CoreGenLoss{$-4$\%}}{\CoreGenLoss{$-2.70$}} \\
        \addlinespace[2pt]
        DP-T
            & \CoreGenCell{\Score{27\%}{51.00}}
                {\CoreGenLoss{$-1$\%}}{\CoreGenGain{+8.90}}
            & \CoreGenCell{\textbf{\Score{43\%}{57.80}}}
                {\CoreGenGain{+15\%}}{\CoreGenGain{+15.70}}
            & \CoreGenCell{\Score{41\%}{56.00}}
                {\CoreGenGain{+13\%}}{\CoreGenGain{+13.90}} \\
        \addlinespace[2pt]
        KineDex
            & \CoreGenCell{\Score{39\%}{48.90}}
                {\CoreGenLoss{$-3$\%}}{\CoreGenLoss{$-9.00$}}
            & \CoreGenCell{\textbf{\Score{55\%}{67.00}}}
                {\CoreGenGain{+13\%}}{\CoreGenGain{+9.10}}
            & \CoreGenCell{\Score{51\%}{63.90}}
                {\CoreGenGain{+9\%}}{\CoreGenGain{+6.00}} \\
        \bottomrule
    \end{tabular}
\end{table}

We then test whether the transfer gains from same-family Core demonstrations extend across policy architectures.
On DishwasherKnob, we repeat the source-family comparison with $\pi_{0.5}$, DP-T, and KineDex. As shown in Table~\ref{tab:coregen_comparison}, for all three policies, Core data from the matching Rotational manipulation family yields the highest success rate and StageScore among the source families tested, improving success rates by at least 10 percentage points over Gen-only training.

Finally, we vary the breadth and composition of the matched Core data to evaluate whether training on a more diverse set of Core tasks improves transfer to the Gen task.
We compare different subsets of Core tasks from the matching family. Using all four Core tasks generally gives better results on the Gen targets than using smaller subsets. Performance also varies with which Core tasks are included. We report the results in Appendix Table~\ref{tab:core-breadth-composition}.

\subsection{Target-Data Efficiency with Core Demonstrations}
\label{sec:target_data_efficiency}

\begin{figure}[ht]
    \centering
    \includegraphics[width=0.96\linewidth]{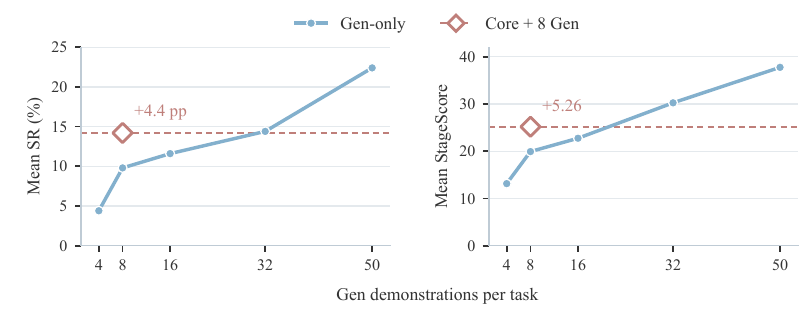}
    \caption{\textbf{Target-data efficiency with $\pi_{0.5}$.} We report mean SR (top) and mean StageScore (bottom) on five Gen tasks (except CandleToCabinet). Blue curves represent Gen-only training. Red diamonds and horizontal dashed lines mark joint training with eight Gen demonstrations and matched Core demonstrations. Red annotations show the gains over Gen-only training with eight demonstrations.}
    \label{fig:gen_data_efficiency}
\end{figure}

Same-family Core data yields the largest transfer gains, but how many target demonstrations are these gains worth?

For each target, we compare Gen-only training with 4, 8, 16, 32, or 50 demonstrations against joint training with 8 Gen demonstrations and demonstrations from the four Core tasks in the same family.
As shown in Figure~\ref{fig:gen_data_efficiency}, joint training with 8 Gen demonstrations places both mean success rate and mean StageScore between the Gen-only results obtained with 16 and 32 demonstrations. These results suggest that related Core experience can partially compensate for limited target supervision, improving the success rate of few-shot adaptation on target Gen tasks.

\subsection{Tactile Sensing Ablation}
\label{sec:touch_experiment}

Finally, we evaluate whether the tactile observations provided by Evo-DexBench improve policy performance.
We compare Full and No-Touch variants of KineDex across all 30 tasks. Following the official KineDex tactile-ablation protocol, we disable both tactile observations and force-target supervision in the No-Touch setting.

As shown in Figure~\ref{fig:tactile_ablation}, tactile sensing improves the success rate or StageScore on most tasks, yielding positive mean gains for five task families. The benefits are particularly evident in contact-rich families such as Rotational manipulation and Articulated configuration control. These results suggest that tactile sensing can provide useful information for dexterous control and yield performance gains when incorporated into a tactile-aware policy.

\section{Discussion and Conclusion}
\label{sec:conclusion}
We introduce Evo-DexBench, which uses Core-Gen task pairs sharing functional interaction requirements as evaluation units. It systematically assesses how well dexterous manipulation policies transfer to new objects and tasks. We evaluate representative models and find that current methods remain limited in their ability to extract and reuse functional interaction knowledge. Evo-DexBench provides a new benchmark for the design and evaluation of general-purpose dexterous manipulation policies.

\textbf{Limitations and Future Work.} All few-shot adaptation strategies considered here require updating model parameters. Future work could evaluate in-context learning under the Core-Gen principle, where a few Gen demonstrations are provided as context without updating model parameters, and examine whether policies can reuse functional interaction knowledge from Core tasks. The evaluation could also be extended to multiple robot embodiments to assess cross-embodiment transfer.

\clearpage

\bibliography{evodexbench_references}

@article{an2025dexterous,
  title={Dexterous manipulation through imitation learning: A survey},
  author={An, Shan and Meng, Ziyu and Tang, Chao and Zhou, Yuning and Liu, Tengyu and Ding, Fangqiang and Zhang, Shufang and Mu, Yao and Song, Ran and Zhang, Wei and others},
  journal={IEEE Transactions on Automation Science and Engineering},
  volume={23},
  pages={1760--1792},
  year={2025},
  publisher={IEEE}
}

@article{zhao2026towards,
  title={Towards Robotic Dexterous Hand Intelligence: A Survey},
  author={Zhao, Weiguang and Liang, Tian and Guo, Xihao and Zhang, Rui and King, Irwin and Huang, Kaizhu},
  journal={arXiv preprint arXiv:2605.13925},
  year={2026}
}

@article{zheng2026egoscale,
  title={Egoscale: Scaling dexterous manipulation with diverse egocentric human data},
  author={Zheng, Ruijie and Niu, Dantong and Xie, Yuqi and Wang, Jing and Xu, Mengda and Jiang, Yunfan and Casta{\~n}eda, Fernando and Hu, Fengyuan and Tan, You Liang and Fu, Letian and others},
  journal={arXiv preprint arXiv:2602.16710},
  year={2026}
}

@article{luo2025beingh0,
  title={Being-h0: vision-language-action pretraining from large-scale human videos},
  author={Luo, Hao and Feng, Yicheng and Zhang, Wanpeng and Zheng, Sipeng and Wang, Ye and Yuan, Haoqi and Liu, Jiazheng and Xu, Chaoyi and Jin, Qin and Lu, Zongqing},
  journal={arXiv preprint arXiv:2507.15597},
  year={2025}
}

@article{mees2022calvin,
author = {Oier Mees and Lukas Hermann and Erick Rosete-Beas and Wolfram Burgard},
title = {CALVIN: A Benchmark for Language-Conditioned Policy Learning for Long-Horizon Robot Manipulation Tasks},
journal={IEEE Robotics and Automation Letters (RA-L)},
volume={7},
number={3},
pages={7327-7334},
year={2022}
}

@article{liu2023libero,
  title={LIBERO: Benchmarking Knowledge Transfer for Lifelong Robot Learning},
  author={Liu, Bo and Zhu, Yifeng and Gao, Chongkai and Feng, Yihao and Liu, Qiang and Zhu, Yuke and Stone, Peter},
  journal={arXiv preprint arXiv:2306.03310},
  year={2023}
}

@article{chen2025robotwin,
  title={Robotwin 2.0: A scalable data generator and benchmark with strong domain randomization for robust bimanual robotic manipulation},
  author={Chen, Tianxing and Chen, Zanxin and Chen, Baijun and Cai, Zijian and Liu, Yibin and Li, Zixuan and Liang, Qiwei and Lin, Xianliang and Ge, Yiheng and Gu, Zhenyu and others},
  journal={arXiv preprint arXiv:2506.18088},
  year={2025}
}

@inproceedings{zhang2025vlabench,
  title={Vlabench: A large-scale benchmark for language-conditioned robotics manipulation with long-horizon reasoning tasks},
  author={Zhang, Shiduo and Xu, Zhe and Liu, Peiju and Yu, Xiaopeng and Li, Yuan and Gao, Qinghui and Fei, Zhaoye and Yin, Zhangyue and Wu, Zuxuan and Jiang, Yu-Gang and others},
  booktitle={2025 IEEE/CVF International Conference on Computer Vision (ICCV)},
  pages={11142--11152},
  year={2025},
  organization={IEEE}
}

@inproceedings{mo2021where2act,
  title={Where2act: From pixels to actions for articulated 3d objects},
  author={Mo, Kaichun and Guibas, Leonidas J and Mukadam, Mustafa and Gupta, Abhinav and Tulsiani, Shubham},
  booktitle={Proceedings of the IEEE/CVF International Conference on Computer Vision},
  pages={6813--6823},
  year={2021}
}

@inproceedings{
wu2022vatmart,
title={{VAT}-Mart: Learning Visual Action Trajectory Proposals for Manipulating 3D {ART}iculated Objects},
author={Ruihai Wu and Yan Zhao and Kaichun Mo and Zizheng Guo and Yian Wang and Tianhao Wu and Qingnan Fan and Xuelin Chen and Leonidas Guibas and Hao Dong},
booktitle={International Conference on Learning Representations},
year={2022},
url={https://openreview.net/forum?id=iEx3PiooLy}
}

@inproceedings{bao2023dexart,
  title={DexArt: Benchmarking Generalizable Dexterous Manipulation with Articulated Objects},
  author={Bao, Chen and Xu, Helin and Qin, Yuzhe and Wang, Xiaolong},
  booktitle={Proceedings of the IEEE/CVF Conference on Computer Vision and Pattern Recognition},
  pages={21190--21200},
  year={2023}
}

@article{foong2026handful,
  title={HANDFUL: Sequential Grasp-Conditioned Dexterous Manipulation with Resource Awareness},
  author={Foong, Ethan and Li, Yunshuang and Jiang, Hao and Sukhatme, Gaurav S and Seita, Daniel},
  journal={arXiv preprint arXiv:2604.25126},
  year={2026}
}

@article{yuan2026dexcraft,
  title={From Grasps to Dexterity: Large-Scale Grasp Pretraining for Dexterous Manipulation},
  author={Yuan, Ying and Liu, Xinyu and Krishna, Sriram and Held, David},
  journal={arXiv preprint arXiv:2606.30749},
  year={2026}
}

@article{wang2026dexjoco,
  title={Dexjoco: A benchmark and toolkit for task-oriented dexterous manipulation on mujoco},
  author={Wang, Hanwen and Zhao, Weizhi and Wang, Xiangyu and Huang, Siyuan and Lin, He and Zheng, Boyuan and Xu, Rongtao and Wang, Gang and Mu, Yao and Wang, He and others},
  journal={arXiv preprint arXiv:2605.16257},
  year={2026}
}

@article{yao2026dexverse,
  title={DexVerse: A Modular Benchmark for Multi-Task, Multi-Embodiment Dexterous Manipulation},
  author={Yao, Yunchao and Xu, Zhuxiu and Zhang, Tianqi and Liu, Zixian and Li, Sikai and Wei, Zhenyu and Chen, Feng and Huang, Dihong and Wan, Kechang and Ma, Chenyang and others},
  journal={arXiv preprint arXiv:2607.08751},
  year={2026}
}

@article{andrychowicz2020shadowdexteroushand,
  title={Learning dexterous in-hand manipulation},
  author={Andrychowicz, OpenAI: Marcin and Baker, Bowen and Chociej, Maciek and Jozefowicz, Rafal and McGrew, Bob and Pachocki, Jakub and Petron, Arthur and Plappert, Matthias and Powell, Glenn and Ray, Alex and others},
  journal={The International Journal of Robotics Research},
  volume={39},
  number={1},
  pages={3--20},
  year={2020},
  publisher={SAGE Publications Sage UK: London, England}
}

@article{rajeswaran2017dapg,
  title={Learning complex dexterous manipulation with deep reinforcement learning and demonstrations},
  author={Rajeswaran, Aravind and Kumar, Vikash and Gupta, Abhishek and Vezzani, Giulia and Schulman, John and Todorov, Emanuel and Levine, Sergey},
  journal={arXiv preprint arXiv:1709.10087},
  year={2017}
}

@inproceedings{qi2023general,
  title={General in-hand object rotation with vision and touch},
  author={Qi, Haozhi and Yi, Brent and Suresh, Sudharshan and Lambeta, Mike and Ma, Yi and Calandra, Roberto and Malik, Jitendra},
  booktitle={Conference on Robot Learning},
  pages={2549--2564},
  year={2023},
  organization={PMLR}
}

@inproceedings{xu2023unidexgrasp,
  title={Unidexgrasp: Universal robotic dexterous grasping via learning diverse proposal generation and goal-conditioned policy},
  author={Xu, Yinzhen and Wan, Weikang and Zhang, Jialiang and Liu, Haoran and Shan, Zikang and Shen, Hao and Wang, Ruicheng and Geng, Haoran and Weng, Yijia and Chen, Jiayi and others},
  booktitle={2023 IEEE/CVF Conference on Computer Vision and Pattern Recognition (CVPR)},
  pages={4737--4746},
  year={2023},
  organization={IEEE}
}

@inproceedings{wan2023unidexgrasp++,
  title={Unidexgrasp++: Improving dexterous grasping policy learning via geometry-aware curriculum and iterative generalist-specialist learning},
  author={Wan, Weikang and Geng, Haoran and Liu, Yun and Shan, Zikang and Yang, Yaodong and Yi, Li and Wang, He},
  booktitle={2023 IEEE/CVF International Conference on Computer Vision (ICCV)},
  pages={3868--3879},
  year={2023},
  organization={IEEE}
}

@inproceedings{qin2023dexpoint,
  title={Dexpoint: Generalizable point cloud reinforcement learning for sim-to-real dexterous manipulation},
  author={Qin, Yuzhe and Huang, Binghao and Yin, Zhao-Heng and Su, Hao and Wang, Xiaolong},
  booktitle={Conference on Robot Learning},
  pages={594--605},
  year={2023},
  organization={PMLR}
}

@article{lum2024dextrah,
  title={Dextrah-g: Pixels-to-action dexterous arm-hand grasping with geometric fabrics},
  author={Lum, Tyler Ga Wei and Matak, Martin and Makoviychuk, Viktor and Handa, Ankur and Allshire, Arthur and Hermans, Tucker and Ratliff, Nathan D and Van Wyk, Karl},
  journal={arXiv preprint arXiv:2407.02274},
  year={2024}
}

@inproceedings{qin2022dexmv,
  title={Dexmv: Imitation learning for dexterous manipulation from human videos},
  author={Qin, Yuzhe and Wu, Yueh-Hua and Liu, Shaowei and Jiang, Hanwen and Yang, Ruihan and Fu, Yang and Wang, Xiaolong},
  booktitle={European Conference on Computer Vision},
  pages={570--587},
  year={2022},
  organization={Springer}
}

@article{wang2024dexcap,
  title={Dexcap: Scalable and portable mocap data collection system for dexterous manipulation},
  author={Wang, Chen and Shi, Haochen and Wang, Weizhuo and Zhang, Ruohan and Fei-Fei, Li and Liu, C Karen},
  journal={arXiv preprint arXiv:2403.07788},
  year={2024}
}

@article{wen2025dexvla,
  title={Dexvla: Vision-language model with plug-in diffusion expert for general robot control},
  author={Wen, Junjie and Zhu, Yichen and Li, Jinming and Tang, Zhibin and Shen, Chaomin and Feng, Feifei},
  journal={arXiv preprint arXiv:2502.05855},
  year={2025}
}

@inproceedings{zhong2026dexgraspvla,
  title={Dexgraspvla: A vision-language-action framework towards general dexterous grasping},
  author={Zhong, Yifan and Huang, Xuchuan and Li, Ruochong and Zhang, Ceyao and Chen, Zhang and Guan, Tianrui and Zeng, Fanlian and Lui, Ka Nam and Ye, Yuyao and Liang, Yitao and others},
  booktitle={Proceedings of the AAAI Conference on Artificial Intelligence},
  volume={40},
  pages={18836--18844},
  year={2026}
}

@article{ye2026dreamzero,
  title={World action models are zero-shot policies},
  author={Ye, Seonghyeon and Ge, Yunhao and Zheng, Kaiyuan and Gao, Shenyuan and Yu, Sihyun and Kurian, George and Indupuru, Suneel and Tan, You Liang and Zhu, Chuning and Xiang, Jiannan and others},
  journal={arXiv preprint arXiv:2602.15922},
  year={2026}
}

@inproceedings{liao2026genie,
  title={Genie envisioner: A unified world foundation platform for robotic manipulation},
  author={Liao, Yue and Zhou, Pengfei and Huang, Siyuan and Yang, Donglin and Chen, Shengcong and Jiang, Yuxin and Hu, Yue and Liu, Si and Luo, Jianlan and Chen, Liliang and others},
  booktitle={International Conference on Learning Representations},
  volume={2026},
  pages={88446--88463},
  year={2026}
}

@article{li2026causal,
  title={Causal world modeling for robot control},
  author={Li, Lin and Zhang, Qihang and Luo, Yiming and Yang, Shuai and Wang, Ruilin and Han, Fei and Yu, Mingrui and Gao, Zelin and Xue, Nan and Zhu, Xing and others},
  journal={arXiv preprint arXiv:2601.21998},
  year={2026}
}

@article{chen2022bidexhands,
  title={Towards human-level bimanual dexterous manipulation with reinforcement learning},
  author={Chen, Yuanpei and Wu, Tianhao and Wang, Shengjie and Feng, Xidong and Jiang, Jiechuan and Lu, Zongqing and McAleer, Stephen and Dong, Hao and Zhu, Song-Chun and Yang, Yaodong},
  journal={Advances in Neural Information Processing Systems},
  volume={35},
  pages={5150--5163},
  year={2022}
}

@article{physicalintelligence2026pi07,
  title   = {$\pi_{0.7}$: A Steerable Generalist Robotic Foundation Model with Emergent Capabilities},
  author  = {{Physical Intelligence} and
             Ai, Bo and
             Amin, Ali and
             Aniceto, Raichelle and
             Balakrishna, Ashwin and
             Balke, Greg and
             Black, Kevin and
             Bokinsky, George and
             Cao, Shihao and
             Charbonnier, Thomas and
             others},
  journal = {arXiv preprint arXiv:2604.15483},
  year    = {2026},
  url     = {https://arxiv.org/abs/2604.15483}
}

@article{yuan2026qwen,
  title={Qwen-robotmanip technical report: Alignment unlocks scale for robotic manipulation foundation models},
  author={Yuan, Haoqi and Liang, Zhixuan and Chen, Anzhe and Wang, Ye and Li, Haoyang and Lin, Pei and Huang, Yiyang and Lei, Zixing and Zhang, Tong and Zhang, Jiazhao and others},
  journal={arXiv preprint arXiv:2606.17846},
  year={2026}
}

@article{fu2026survey,
  title={A survey on learning an autonomous dynamic system for human--robot skills transfer from demonstration},
  author={Fu, Jiayun and Huang, Haotian and Jin, Zhehao and Liu, Andong and Zhang, Wen-An and Yu, Li and Si, Weiyong and Yang, Chenguang},
  journal={Robotics and Computer-Integrated Manufacturing},
  volume={97},
  pages={103092},
  year={2026},
  publisher={Elsevier}
}

@article{li2022survey,
  title={A survey of multifingered robotic manipulation: Biological results, structural evolvements, and learning methods},
  author={Li, Yinlin and Wang, Peng and Li, Rui and Tao, Mo and Liu, Zhiyong and Qiao, Hong},
  journal={Frontiers in neurorobotics},
  volume={16},
  pages={843267},
  year={2022},
  publisher={Frontiers Media SA}
}

@article{tsuji2025survey,
  title={A survey on imitation learning for contact-rich tasks in robotics},
  author={Tsuji, Toshiaki and Kato, Yasuhiro and Solak, Gokhan and Zhang, Heng and Petri{\v{c}}, Tadej and Nori, Francesco and Ajoudani, Arash},
  journal={The International Journal of Robotics Research},
  pages={02783649261417694},
  year={2025},
  publisher={Sage Publications, Sage UK: London, England}
}

@misc{chi2024diffusionpolicyvisuomotorpolicy,
      title={Diffusion Policy: Visuomotor Policy Learning via Action Diffusion}, 
      author={Cheng Chi and Zhenjia Xu and Siyuan Feng and Eric Cousineau and Yilun Du and Benjamin Burchfiel and Russ Tedrake and Shuran Song},
      year={2024},
      eprint={2303.04137},
      archivePrefix={arXiv},
      primaryClass={cs.RO},
      url={https://arxiv.org/abs/2303.04137}, 
}

@misc{intelligence2025pi05visionlanguageactionmodelopenworld,
      title={$\pi_{0.5}$: a Vision-Language-Action Model with Open-World Generalization}, 
      author={Physical Intelligence and Kevin Black and Noah Brown and James Darpinian and Karan Dhabalia and Danny Driess and Adnan Esmail and Michael Equi and Chelsea Finn and Niccolo Fusai and Manuel Y. Galliker and Dibya Ghosh and Lachy Groom and Karol Hausman and Brian Ichter and Szymon Jakubczak and Tim Jones and Liyiming Ke and Devin LeBlanc and Sergey Levine and Adrian Li-Bell and Mohith Mothukuri and Suraj Nair and Karl Pertsch and Allen Z. Ren and Lucy Xiaoyang Shi and Laura Smith and Jost Tobias Springenberg and Kyle Stachowicz and James Tanner and Quan Vuong and Homer Walke and Anna Walling and Haohuan Wang and Lili Yu and Ury Zhilinsky},
      year={2025},
      eprint={2504.16054},
      archivePrefix={arXiv},
      primaryClass={cs.LG},
      url={https://arxiv.org/abs/2504.16054}, 
}

@misc{zhang2025kinedexlearningtactileinformedvisuomotor,
      title={KineDex: Learning Tactile-Informed Visuomotor Policies via Kinesthetic Teaching for Dexterous Manipulation}, 
      author={Di Zhang and Chengbo Yuan and Chuan Wen and Hai Zhang and Junqiao Zhao and Yang Gao},
      year={2025},
      eprint={2505.01974},
      archivePrefix={arXiv},
      primaryClass={cs.RO},
      url={https://arxiv.org/abs/2505.01974}, 
}

@misc{shukor2025smolvlavisionlanguageactionmodelaffordable,
      title={SmolVLA: A Vision-Language-Action Model for Affordable and Efficient Robotics}, 
      author={Mustafa Shukor and Dana Aubakirova and Francesco Capuano and Pepijn Kooijmans and Steven Palma and Adil Zouitine and Michel Aractingi and Caroline Pascal and Martino Russi and Andres Marafioti and Simon Alibert and Matthieu Cord and Thomas Wolf and Remi Cadene},
      year={2025},
      eprint={2506.01844},
      archivePrefix={arXiv},
      primaryClass={cs.LG},
      url={https://arxiv.org/abs/2506.01844}, 
}

@article{qin2023anyteleop,
  title={Anyteleop: A general vision-based dexterous robot arm-hand teleoperation system},
  author={Qin, Yuzhe and Yang, Wei and Huang, Binghao and Van Wyk, Karl and Su, Hao and Wang, Xiaolong and Chao, Yu-Wei and Fox, Dieter},
  journal={arXiv preprint arXiv:2307.04577},
  year={2023}
}

@inproceedings{calli2015ycb,
  title={The ycb object and model set: Towards common benchmarks for manipulation research},
  author={Calli, Berk and Singh, Arjun and Walsman, Aaron and Srinivasa, Siddhartha and Abbeel, Pieter and Dollar, Aaron M},
  booktitle={2015 international conference on advanced robotics (ICAR)},
  pages={510--517},
  year={2015},
  organization={IEEE}
}

@inproceedings{xiang2020sapien,
  title={Sapien: A simulated part-based interactive environment},
  author={Xiang, Fanbo and Qin, Yuzhe and Mo, Kaichun and Xia, Yikuan and Zhu, Hao and Liu, Fangchen and Liu, Minghua and Jiang, Hanxiao and Yuan, Yifu and Wang, He and others},
  booktitle={2020 IEEE/CVF Conference on Computer Vision and Pattern Recognition (CVPR)},
  pages={11094--11104},
  year={2020},
  organization={IEEE}
}

@article{nasiriany2024robocasa,
  title={Robocasa: Large-scale simulation of everyday tasks for generalist robots},
  author={Nasiriany, Soroush and Maddukuri, Abhiram and Zhang, Lance and Parikh, Adeet and Lo, Aaron and Joshi, Abhishek and Mandlekar, Ajay and Zhu, Yuke},
  journal={arXiv preprint arXiv:2406.02523},
  year={2024}
}

@inproceedings{yu2020meta,
  title={Meta-world: A benchmark and evaluation for multi-task and meta reinforcement learning},
  author={Yu, Tianhe and Quillen, Deirdre and He, Zhanpeng and Julian, Ryan and Hausman, Karol and Finn, Chelsea and Levine, Sergey},
  booktitle={Conference on robot learning},
  pages={1094--1100},
  year={2020},
  organization={PMLR}
}

@misc{community2026starvlalegolikecodebasevisionlanguageaction,
  title         = {StarVLA: A Lego-like Codebase for Vision-Language-Action Model Developing},
  author        = {StarVLA Community},
  year          = {2026},
  eprint        = {2604.05014},
  archiveprefix = {arXiv},
  primaryclass  = {cs.RO},
  url           = {https://arxiv.org/abs/2604.05014}
}

\clearpage
\appendix
\section*{Appendix}
This appendix provides the complete specification, experimental settings,
and per-task results of Evo-DexBench.
Section~\ref{app:task-specification} gives the natural-language description
and stage-wise success criteria of all 30 tasks;
section~\ref{app:scenes} describes the three scene layouts of each task and
the perturbation ranges and sampling of the four generalization axes:
camera, lighting, physical, and spatial;
section~\ref{app:tactile} describes the layout of the 123 virtual taxels on
each Shadow Hand, the computation of tactile force from contact impulses,
and the KineDex tactile interface and No-Touch control;
section~\ref{app:training} details the data format, the implementation and
adaptation of the six policies, the training hyperparameters, the transfer
protocols, and the evaluation protocol;
section~\ref{app:results} reports the complete baseline, transfer,
data-efficiency, and tactile-ablation results.

\section{Task specification}
\label{app:task-specification}

Evo-DexBench comprises 30 tasks organized into six functional families. Each
family consists of four \emph{Core} tasks, each with 50 successful
demonstrations, and one \emph{Gen} task with eight successful demonstrations.  The Core tasks instantiate a shared functional interaction in different task
contexts, whereas the Gen task requires the same interaction in a new context.  The first three families use a single-arm Franka Panda--Shadow Hand setup, and
the last three use a bimanual setup.  \Cref{tab:task-design-button,tab:task-design-rotational,tab:task-design-access,tab:task-design-structured,tab:task-design-assembly,tab:task-design-articulated}
list, for each task, its natural-language description and task-specific stage
success conditions; the 100\% stage specifies the complete terminal success
criterion.
Success and StageScore are determined independently. Success depends only on whether the episode meets the task's success criterion at the end, whereas StageScore is the highest stage credited in the episode. Consequently, a successful episode does not receive a StageScore of 100 if any intermediate stage is never triggered. For example, if DisplayPowerButton is pressed with a knuckle or the palm and the fingertip never comes within 0.05\,m of the button, the 30\% stage is never triggered. In this case, the 60\% and 100\% stages are not credited, even if the contact force exceeds 0.15\,N and the button travel reaches 70\% for 3 control steps.

All stage criteria are computed from simulator state and contact
measurements, not from policy observations. Intermediate stages separate
meaningful task progress from arbitrary motion. Some criteria require a condition to
hold for multiple control steps, so that transient contacts
and momentary threshold crossings are not credited.

\begingroup
\newlength{\StageLabelWidth}
\settowidth{\StageLabelWidth}{100\%:\space}
\newcommand{\StageLabel}[1]{%
  \hangindent=\StageLabelWidth%
  \hangafter=1\relax%
  \makebox[\StageLabelWidth][r]{#1\%:\space}\ignorespaces%
}
\newcommand{\StageConditionSetup}{%
  \setlength{\parindent}{0pt}%
  \renewcommand{\newline}{\par\noindent}%
}
\newcolumntype{V}[1]{>{\raggedright\arraybackslash}m{#1}}
\newcolumntype{S}[1]{>{\raggedright\arraybackslash\StageConditionSetup}m{#1}}

\begin{longtable}{@{}V{0.16\linewidth}V{0.25\linewidth}S{0.55\linewidth}@{}}
\caption{Task definitions for the Button-driven family.}\label{tab:task-design-button}\\
\toprule
Task & Description & Stage success conditions \\
\midrule
\endfirsthead
\multicolumn{3}{@{}l}{\textbf{Table~\thetable{} continued from the previous page}}\\
\toprule
Task & Description & Stage success conditions \\
\midrule
\endhead
\midrule
\multicolumn{3}{r@{}}{Continued on the next page}\\
\endfoot
\bottomrule
\endlastfoot
Lamp\allowbreak{}Button & Press the lamp button and keep the lamp on. & \StageLabel{30} index-finger force $\ge$0.15 N;\newline \StageLabel{60} button travel $\ge$70\%;\newline \StageLabel{100} lamp remains lit for $\ge$1 s after triggering. \\
\midrule
Wall\allowbreak{}Switch & Flip the wall switch on. & \StageLabel{30} hand tip within 0.20 m;\newline \StageLabel{60} angle $\ge$0.061 rad;\newline \StageLabel{100} switch angle $\ge$0.122 rad for 1 s. \\
\midrule
Phone\allowbreak{}Home\allowbreak{}Button & Press and release the phone Home button. & \StageLabel{30} correct contact force $\ge$0.12 N;\newline \StageLabel{70} contact held while travel $\ge$70\%;\newline \StageLabel{100} button released, return travel $\le$20\%, and phone remains safely on the table. \\
\midrule
Toaster\allowbreak{}Slider & Press the toaster slider. & \StageLabel{30} hand tip within 0.40 m of toaster;\newline \StageLabel{60} hand tip within 0.35 m of slider point;\newline \StageLabel{100} toaster slider displacement $\ge$0.062 m. \\
\midrule
Display\allowbreak{}Power\allowbreak{}Button & Press the display power button. & \StageLabel{30} fingertip within 0.05 m;\newline \StageLabel{60} contact force $\ge$0.15 N;\newline \StageLabel{100} contact followed by travel $\ge$70\% and lasted for 3 control steps. \\
\end{longtable}

\begin{longtable}{@{}V{0.15\linewidth}V{0.23\linewidth}S{0.58\linewidth}@{}}
\caption{Task definitions for the Rotational manipulation family.}\label{tab:task-design-rotational}\\
\toprule
Task & Description & Stage success conditions \\
\midrule
\endfirsthead
\multicolumn{3}{@{}l}{\textbf{Table~\thetable{} continued from the previous page}}\\
\toprule
Task & Description & Stage success conditions \\
\midrule
\endhead
\midrule
\multicolumn{3}{r@{}}{Continued on the next page}\\
\endfoot
\bottomrule
\endlastfoot
Faucet\allowbreak{}Handle & Turn the faucet handle open. & \StageLabel{30} handle contact force $\ge$0.40 N;\newline \StageLabel{60} opening travel $\ge$50\%;\newline \StageLabel{100} opening travel $\ge$80\%. \\
\midrule
Mouse\allowbreak{}Wheel & Hold the mouse and scroll its wheel. & \StageLabel{25} mouse contact force $\ge$0.15 N or wheel rotation $>0.03$ rad;\newline \StageLabel{50} wheel contact force $\ge$0.15 N or rotation $>0.03$ rad;\newline \StageLabel{100} wheel rotation $\ge$1.5 rad, mouse linear speed $\le$0.04 m/s, and angular speed $\le$0.60 rad/s. \\
\midrule
Coffee\allowbreak{}Knob & Turn the coffee-machine knob to its designated target setting. & \StageLabel{30} knob contact force $\ge$0.40 N;\newline \StageLabel{50} turning toward the target at $\ge$0.03 rad/s with valid knob contact has been recorded;\newline \StageLabel{75} angle error $\le$20$^\circ$;\newline \StageLabel{100} angle error $\le$6$^\circ$, knob angular speed $\le$0.20 rad/s. \\
\midrule
Kettle\allowbreak{}Lid\allowbreak{}Tab & Rotate the kettle-lid tab to the notch. & \StageLabel{30} thumb and index contacts, each $\ge$0.35 N;\newline \StageLabel{55} positive rotation $\ge$5$^\circ$ at $\ge$0.03 rad/s;\newline \StageLabel{80} angle in 60$^\circ$--90$^\circ$;\newline \StageLabel{100} angle error $\le$9$^\circ$, tab angular speed $\le$0.50 rad/s, kettle displacement $\le$0.01 m, and tilt $\le$3$^\circ$. \\
\midrule
Dishwasher\allowbreak{}Knob & Turn the dishwasher knob to its target setting. & \StageLabel{30} knob contact force $\ge$0.40 N;\newline \StageLabel{60} angle error $\le$45$^\circ$;\newline \StageLabel{100} effective turn recorded, angle error $\le$9$^\circ$, and angular speed $\le$1.00 rad/s. \\
\end{longtable}

\begin{longtable}{@{}V{0.14\linewidth}V{0.2\linewidth}S{0.62\linewidth}@{}}
\caption{Task definitions for the Access-and-transfer family.}\label{tab:task-design-access}\\
\toprule
Task & Description & Stage success conditions \\
\midrule
\endfirsthead
\multicolumn{3}{@{}l}{\textbf{Table~\thetable{} continued from the previous page}}\\
\toprule
Task & Description & Stage success conditions \\
\midrule
\endhead
\midrule
\multicolumn{3}{r@{}}{Continued on the next page}\\
\endfoot
\bottomrule
\endlastfoot
Cup\allowbreak{}From\allowbreak{}Cabinet & Open the cabinet, remove the cup, and place it upright on the counter. & \StageLabel{25} door angle $>45^\circ$;\newline \StageLabel{50} valid cup-body/handle grasp (force $\ge$0.40 N);\newline \StageLabel{75} cup clears the cabinet opening;\newline \StageLabel{100} cup is on the counter, tilt $\le$15$^\circ$, released, with linear speed $\le$0.05 m/s and angular speed $\le$0.60 rad/s for 0.40 s. \\
\midrule
Microwave\allowbreak{}Bowl & Open the microwave, place the bowl inside, and close the door. & \StageLabel{25} door opening $\ge$45\%;\newline \StageLabel{50} two-finger bowl grasp, each force $\ge$0.30 N, held 0.12 s;\newline \StageLabel{75} bowl fully inside, support $\ge$0.20 N, tilt $\le$35$^\circ$, released, linear speed $\le$0.05 m/s and angular speed $\le$0.60 rad/s for 0.35 s;\newline \StageLabel{100} placement confirmed and the same bowl conditions hold with the door closed for 0.30 s. Closed condition is at opening $\le$5\% and speed $\le$0.08 rad/s, and exits above 8\%. \\
\midrule
Tennis\allowbreak{}Ball\allowbreak{}Transfer & Move the tennis ball from shelf 1 to shelf 2 and close the cabinet. & \StageLabel{20} door opening $\ge$55\%;\newline \StageLabel{45} two-finger grasp with force $\ge$0.40 N per finger;\newline \StageLabel{75} ball in shelf-2 target, supported, released, linear speed $\le$0.03 m/s, angular speed $\le$0.60 rad/s for 5 steps;\newline \StageLabel{100} target was placed, the door is confirmed closed with speed $\le$0.01 m/s for 5 steps, and the ball satisfies the placement conditions. \\
\midrule
Banana\allowbreak{}From\allowbreak{}Drawer & Open the drawer, place the banana on the desk, and close the drawer. & \StageLabel{25} drawer opening $\ge$45\%;\newline \StageLabel{40} valid grasp held 0.15 s;\newline \StageLabel{60} banana fully clears drawer;\newline \StageLabel{85} desk support $\ge$0.20 N, released and still for 0.50 s;\newline \StageLabel{100} banana remains released and stable on the desk while the drawer is confirmed closed with speed $\le$0.01 m/s for 0.15 s. \\
\midrule
Candle\allowbreak{}To\allowbreak{}Cabinet & Open the upper glass cabinet, place the candle inside, and release it. & \StageLabel{25} an upper door angle $>60^\circ$;\newline \StageLabel{50} at least two fingers, each $\ge$0.30 N;\newline \StageLabel{75} candle bounding box inside cabinet;\newline \StageLabel{100} candle remains inside and upright with base-height error $\le$0.025 m, released, tilt $\le$20$^\circ$, linear speed $\le$0.03 m/s, and angular speed $\le$0.50 rad/s for 1 s. \\
\end{longtable}

\begin{longtable}{@{}V{0.12\linewidth}V{0.18\linewidth}S{0.65\linewidth}@{}}
\caption{Task definitions for the Structured configuration family.}\label{tab:task-design-structured}\\
\toprule
Task & Description & Stage success conditions \\
\midrule
\endfirsthead
\multicolumn{3}{@{}l}{\textbf{Table~\thetable{} continued from the previous page}}\\
\toprule
Task & Description & Stage success conditions \\
\midrule
\endhead
\midrule
\multicolumn{3}{r@{}}{Continued on the next page}\\
\endfoot
\bottomrule
\endlastfoot
Tableware\allowbreak{}Placement & Place the plate in the center, fork left, and knife right. & \StageLabel{30} any one piece correctly placed;\newline \StageLabel{60} any two pieces correctly placed;\newline \StageLabel{90} all three correctly placed. Correct placement requires position error $\le$0.12 m, tilt $\le$35$^\circ$, utensil yaw error $\le$35$^\circ$, release, linear speed $\le$0.04 m/s and angular speed $\le$0.60 rad/s;\newline \StageLabel{100} all three have valid placements, left/right relations and clearance for 1 s. \\
\midrule
Hammer\allowbreak{}Block\allowbreak{}Assembly & Arrange three blocks into the reference hammer shape. & \StageLabel{25} one block position/height error $\le$0.015 m, yaw/tilt $\le$8$^\circ$, released;\newline \StageLabel{50} at least two blocks have valid poses and releases, with at least one prescribed adjacency pair (center error $\le$0.020 m, edge angle $\le$12$^\circ$);\newline \StageLabel{75} all three target poses and releases valid;\newline \StageLabel{100} all three target poses and prescribed adjacency pairs valid, all blocks released, linear speed $\le$0.03 m/s, and angular speed $\le$0.50 rad/s for 0.50 s. \\
\midrule
Sandwich\allowbreak{}Assembly & Place the patty and top bread on the pre-stacked bottom bread and cheese. & \StageLabel{40} either movable ingredient correctly placed;\newline \StageLabel{80} both movable ingredients correctly placed. Placement requires position error $\le$0.050 m, height error $\le$0.018 m, tilt $\le$25$^\circ$, valid support and overlap, release, linear speed $\le$0.08 m/s and angular speed $\le$1.20 rad/s;\newline \StageLabel{100} all four layers satisfy the ordered support, pose, overlap, release and speed conditions for 0.25 s. \\
\midrule
Ring\allowbreak{}Tower\allowbreak{}Assembly & Stack the red, blue, and yellow rings from largest to smallest. & \StageLabel{30} red ring correctly on bottom;\newline \StageLabel{60} red and blue rings correctly placed; each ring center within 0.028 m of post axis, height error $\le$0.018 m, tilt $\le$22$^\circ$;\newline \StageLabel{100} correct three-ring order and poses, with linear speed $\le$0.06 m/s and angular speed $\le$0.80 rad/s for 20 steps. \\
\midrule
Box\allowbreak{}Packing & Pack three shaped objects into the box and close the lid. & \StageLabel{20} at least one object fully inside;\newline \StageLabel{40} at least two objects fully inside;\newline \StageLabel{75} all three objects fully inside;\newline \StageLabel{100} all three objects fully inside and maximum opening across all lids $\le$4\%. \\
\end{longtable}

\begin{longtable}{@{}V{0.14\linewidth}V{0.18\linewidth}S{0.64\linewidth}@{}}
\caption{Task definitions for the Assembly family.}\label{tab:task-design-assembly}\\
\toprule
Task & Description & Stage success conditions \\
\midrule
\endfirsthead
\multicolumn{3}{@{}l}{\textbf{Table~\thetable{} continued from the previous page}}\\
\toprule
Task & Description & Stage success conditions \\
\midrule
\endhead
\midrule
\multicolumn{3}{r@{}}{Continued on the next page}\\
\endfoot
\bottomrule
\endlastfoot
Pen\allowbreak{}Capping & Remove the pen, grasp its cap, and cap the pen. & \StageLabel{20} either object previously grasped ($\ge$0.15 N per parent link);\newline \StageLabel{40} both objects previously grasped;\newline \StageLabel{70} axis error $\le$18$^\circ$, lateral error $\le$0.015 m, tip inside cap range;\newline \StageLabel{100} axis error $\le$18$^\circ$, lateral error $\le$0.010 m, and insertion depth $\ge$0.015 m. \\
\midrule
Cup\allowbreak{}Sleeve\allowbreak{}Fit & Slide a sleeve from the cup bottom to its middle. & \StageLabel{20} either object previously grasped ($\ge$0.18 N per parent link);\newline \StageLabel{40} both objects previously grasped;\newline \StageLabel{70} axis error $\le$15$^\circ$, lateral error $\le$0.018$\times$scale, center height in [$-0.055$, $0.040$]$\times$scale;\newline \StageLabel{100} center height in [$0.062$, $0.100$]$\times$scale, lateral error $\le$0.010$\times$scale, axis error $\le$15$^\circ$. Here scale $=0.72$. \\
\midrule
Bottle\allowbreak{}Funnel\allowbreak{}Insertion & Align a funnel with a bottle and insert it. & \StageLabel{20} either object previously grasped ($\ge$0.18 N per parent link);\newline \StageLabel{40} both objects previously grasped;\newline \StageLabel{70} axis error $\le$15$^\circ$, lateral error $\le$0.018 m, tip near mouth;\newline \StageLabel{100} axis error $\le$15$^\circ$, insertion depth $\ge$0.010 m, tip height in [0.1031, 0.1660] m in the bottle frame, lateral error $\le$0.010 m, linear speed relative to bottle $\le$0.06 m/s, and relative angular speed $\le$1.00 rad/s for 3 steps. \\
\midrule
Plug\allowbreak{}Insertion & Hold a plug and socket and fully insert the plug. & \StageLabel{25} opposite hands hold both objects, at least two parent contacts per object at $\ge$0.20 N;\newline \StageLabel{50} both actual pins align with the socket channels, direction error $\le$12$^\circ$, tips near the openings;\newline \StageLabel{75} both aligned pins enter their channels;\newline \StageLabel{100} both pins remain aligned with depth $\ge$0.01575 m each, and linear speed relative to socket $\le$0.15 m/s. \\
\midrule
Scissors\allowbreak{}Sheathing & Insert the tips of closed scissors into the sheath. & \StageLabel{20} scissors previously grasped with at least two parent contacts, each $\ge$0.20 N, for 0.10 s;\newline \StageLabel{60} both real blade tips align with the channel cross-section, axis error $\le$12$^\circ$, axial distance from entrance $\le$0.025 m;\newline \StageLabel{100} both blade tips inside the sheath with 0.002 m geometric tolerance, axis error $\le$12$^\circ$, maximum tip speed relative to sheath $\le$0.08 m/s, and relative angular speed $\le$1.00 rad/s for 0.10 s. \\
\end{longtable}

\begin{longtable}{@{}V{0.14\linewidth}V{0.2\linewidth}S{0.62\linewidth}@{}}
\caption{Task definitions for the Articulated configuration control family.}\label{tab:task-design-articulated}\\
\toprule
Task & Description & Stage success conditions \\
\midrule
\endfirsthead
\multicolumn{3}{@{}l}{\textbf{Table~\thetable{} continued from the previous page}}\\
\toprule
Task & Description & Stage success conditions \\
\midrule
\endhead
\midrule
\multicolumn{3}{r@{}}{Continued on the next page}\\
\endfoot
\bottomrule
\endlastfoot
Laptop\allowbreak{}Closing & Stabilize the laptop base and close the screen to about 25$^\circ$. & \StageLabel{30} opposite hands contact base/screen at $\ge$0.40 N, base displacement $\le$0.040 m and tilt $\le$5$^\circ$;\newline \StageLabel{60} screen reaches 60$^\circ$$\pm$8$^\circ$;\newline \StageLabel{100} screen reaches 25$^\circ$$\pm$5$^\circ$, base remains stable, and absolute screen angular speed $\le$1.00 rad/s for 0.25 s. \\
\midrule
Stapler\allowbreak{}Paper\allowbreak{}Mark & Insert paper, press the stapler, then move it away. & \StageLabel{20} paper force $\ge$0.50 N or displacement $\ge$0.03 m;\newline \StageLabel{50} paper enters staple area with tilt $\le$35$^\circ$;\newline \StageLabel{70} paper remains in area, stapler press-arm qpos $\le$0.10 rad, mark created;\newline \StageLabel{100} after a mark is created, the current stapling point and the paper mark remain $\ge$0.06 m apart for 5 steps. \\
\midrule
Scissors\allowbreak{}Opening & Stabilize scissors and open the moving handle. & \StageLabel{40} support-handle contact $\ge$0.15 N or base linear speed $<0.30$ m/s has been recorded, and current base linear/angular speeds are $\le$0.30 m/s and $\le$6.00 rad/s;\newline \StageLabel{70} moving-handle contact $\ge$0.15 N or proximity to target has been recorded, and current angle error $\le$0.50 rad;\newline \StageLabel{100} opening angle 1.263$\pm$0.20 rad, scissors-base linear speed $\le$0.30 m/s, and angular speed $\le$6.00 rad/s for 10 steps. \\
\midrule
Box\allowbreak{}Lid\allowbreak{}Opening & Stabilize a box and open its hinged lid. & \StageLabel{30} task contact $\ge$0.40 N with box displacement $\le$0.10 m, tilt $\le$7$^\circ$, linear speed $\le$0.05 m/s and angular speed $\le$0.50 rad/s has been recorded;\newline \StageLabel{60} opening past $\ge$10$^\circ$ with a stable box, lid contact $\ge$0.40 N and positive opening velocity $\ge$0.02 rad/s has been recorded;\newline \StageLabel{100} lid angle 90$^\circ$--105$^\circ$, box stable, and lid angular speed $\le$0.25 rad/s for 1 s. \\
\midrule
Glasses\allowbreak{}Folding & Fold both eyeglass temples into the storage position. & \StageLabel{40} left temple angle $\ge$0.55 rad has been recorded;\newline \StageLabel{80} right temple angle $\ge$0.459 rad has been recorded;\newline \StageLabel{100} right/left angles satisfy either ($\ge$0.91, $\ge$0.55) rad or ($\ge$0.459, $\ge$1.090) rad. \\
\end{longtable}

\endgroup

\section{Scene layouts and generalization protocol}
\label{app:scenes}

\subsection{Scene-layout variation}
\label{app:scene-layouts}

\newif\ifSceneLayoutsByTaskRow
\SceneLayoutsByTaskRowtrue

Each task is instantiated in three independently constructed scene layouts. These layouts preserve the same task semantics and success criteria but differ in visual context, object arrangement, and overall scene composition, which reduces reliance on any single canonical layout. This design separates scene-level diversity from the continuous single-axis perturbations described below. The panels below show all 90 task--layout combinations.
\ifSceneLayoutsByTaskRow Tasks are grouped by family, with one task per row: within each family, the first four rows show the Core tasks and the fifth row shows the Gen task. Each row presents the three layouts of a task from left to right, with its canonical name centered below.\fi

\ifSceneLayoutsByTaskRow

\begingroup
\newcommand{\LayoutStripImage}[1]{\includegraphics[width=0.316\textwidth]{figures/sceneLayout/#1}}
\newcommand{\TaskLayoutStrip}[5]{%
  \noindent\begin{minipage}{\textwidth}
  \centering
  \LayoutStripImage{#3}\hfill
  \LayoutStripImage{#4}\hfill
  \LayoutStripImage{#5}\\[-1pt]
  {\scriptsize\texttt{#1}\enspace\textit{(#2)}}
  \end{minipage}\par\vspace{3pt}}
\newcommand{\LayoutFamilyStart}[6]{%
  \par\addvspace{8pt}
  \noindent\begin{minipage}{\textwidth}
  \centering
  {\small\bfseries #1}\par\vspace{4pt}
  \LayoutStripImage{#4}\hfill
  \LayoutStripImage{#5}\hfill
  \LayoutStripImage{#6}\\[-1pt]
  {\scriptsize\texttt{#2}\enspace\textit{(#3)}}
  \end{minipage}\par\vspace{3pt}}

\LayoutFamilyStart{Button-driven}{LampButton}{Core}{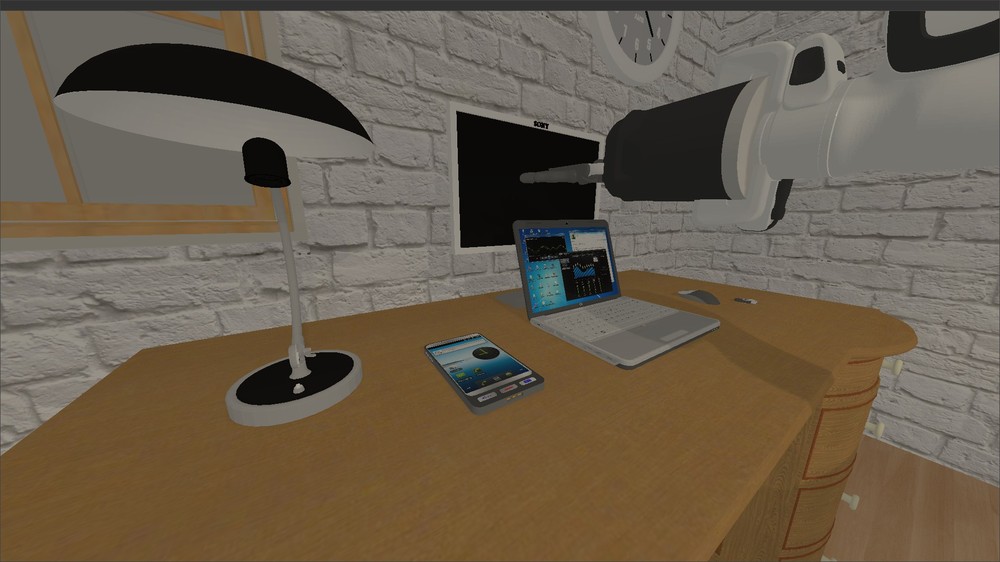}{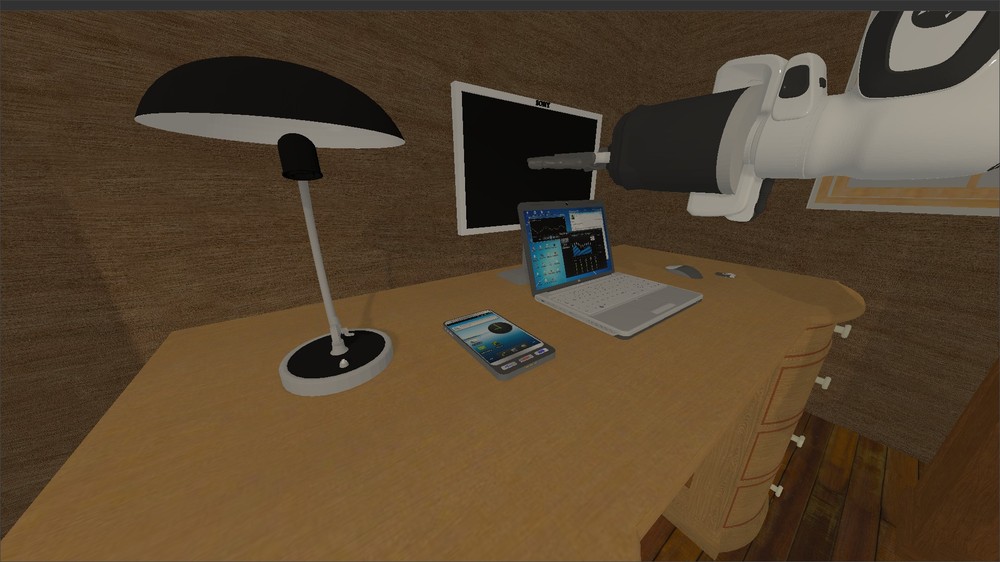}{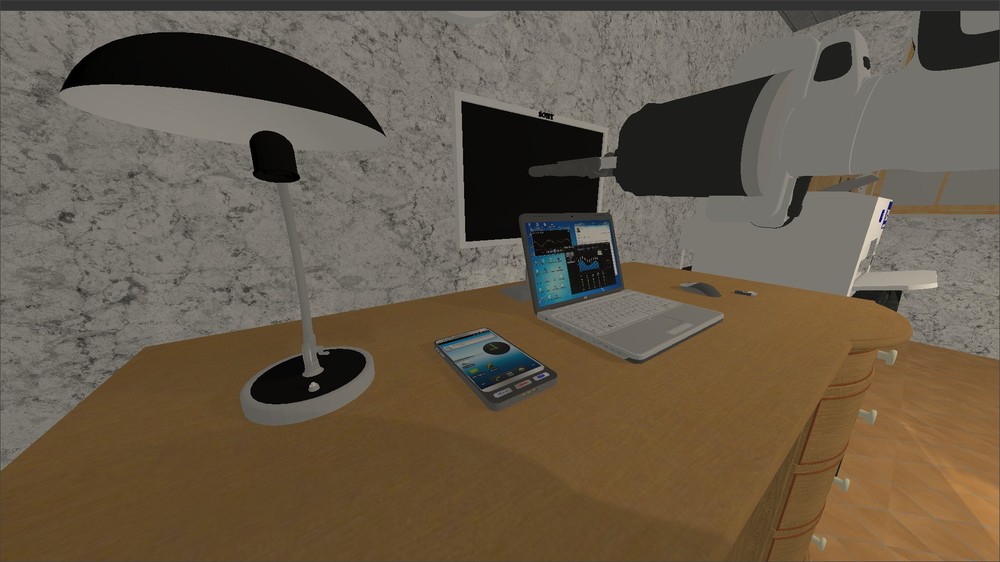}
\TaskLayoutStrip{WallSwitch}{Core}{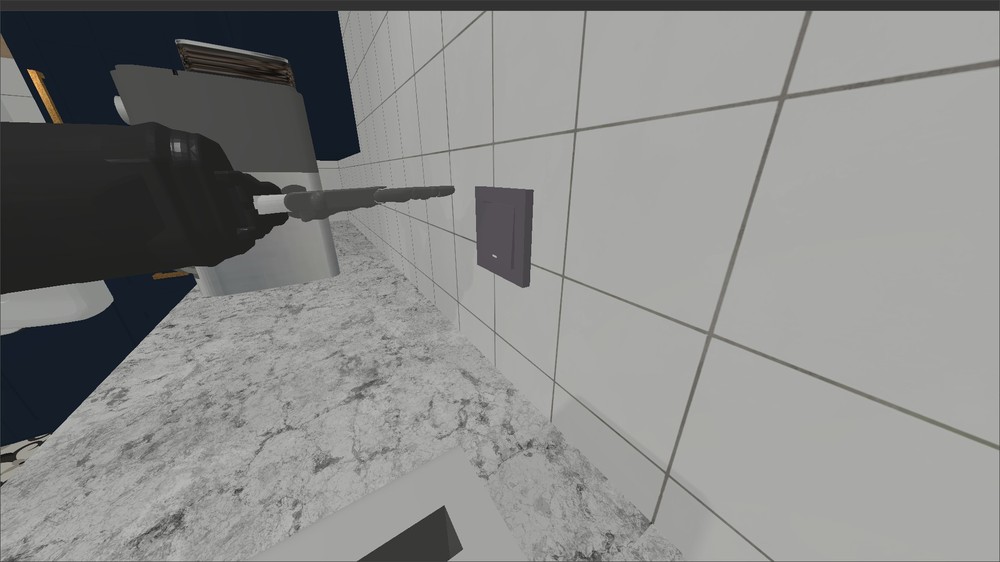}{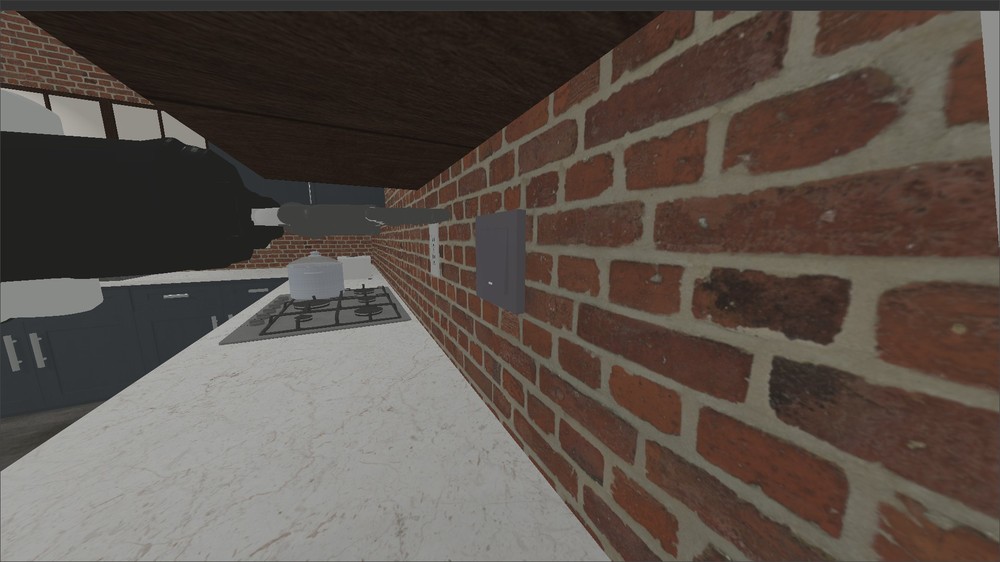}{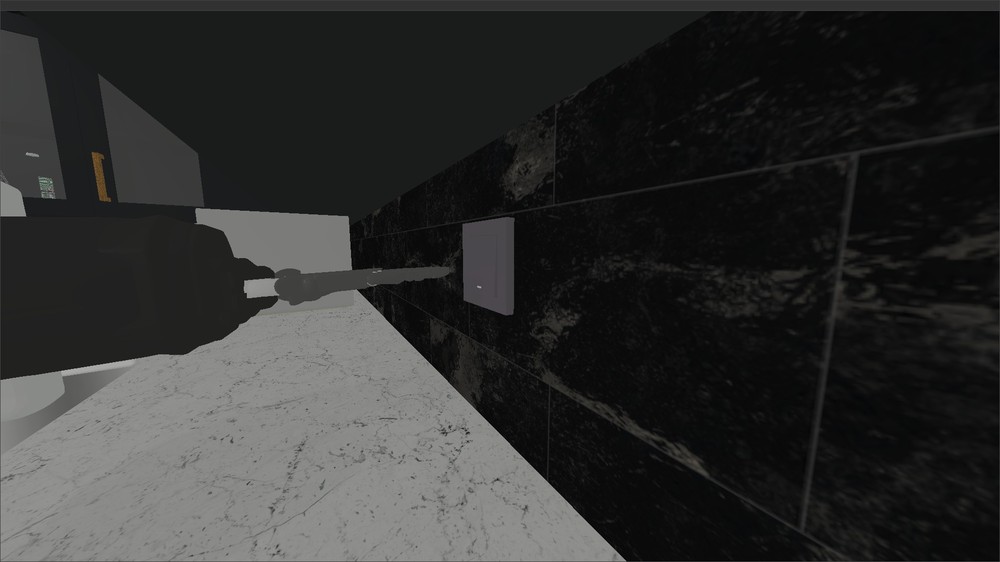}
\TaskLayoutStrip{PhoneHomeButton}{Core}{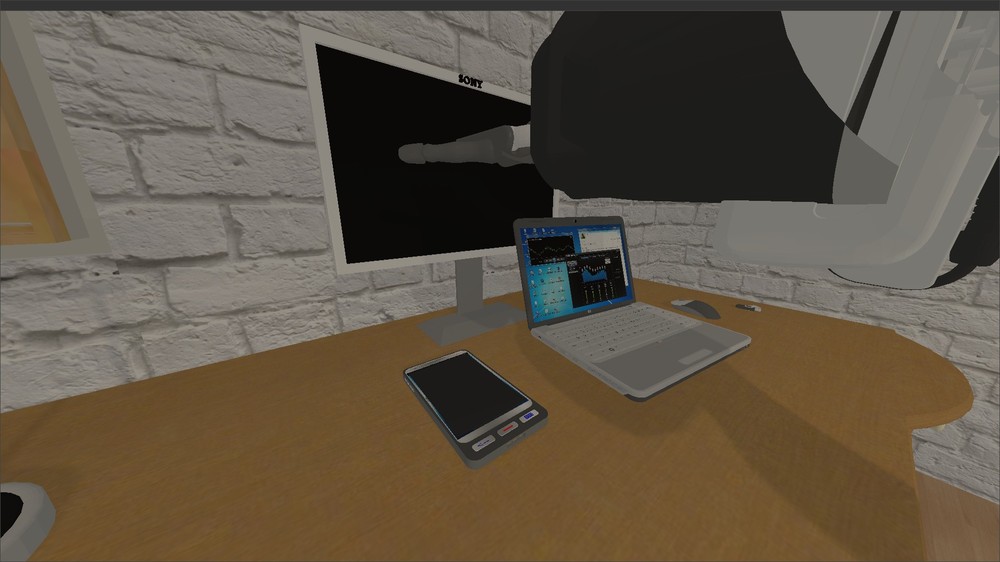}{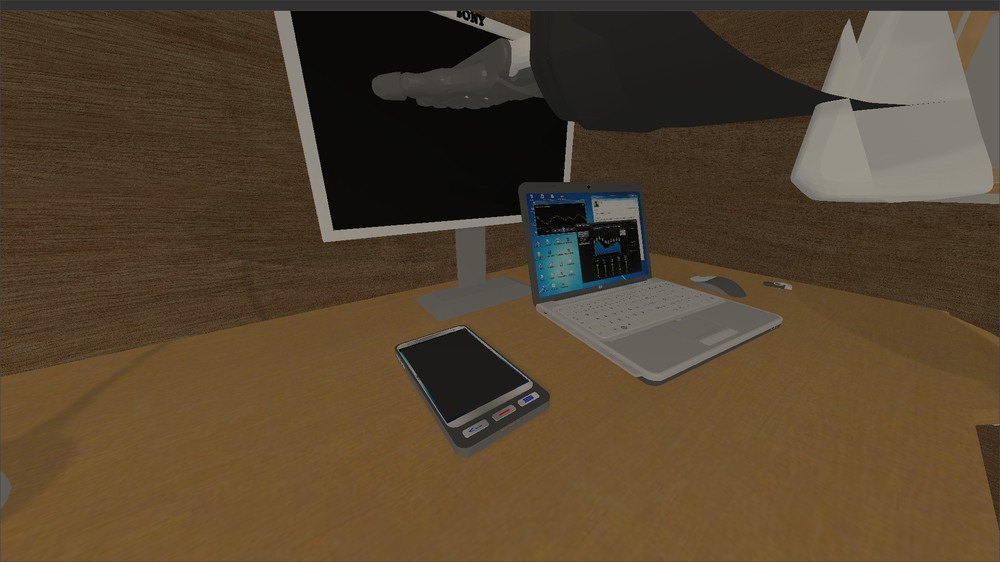}{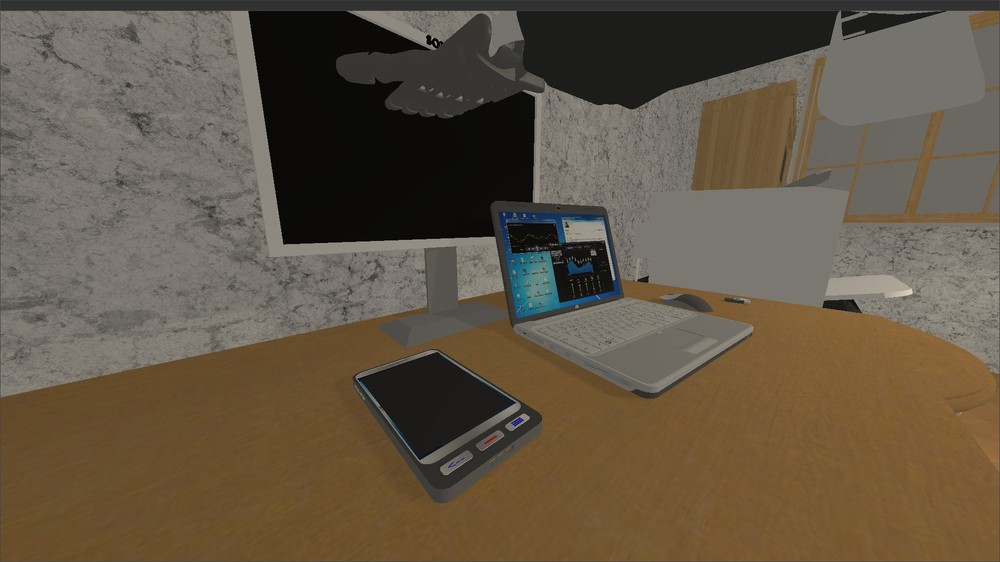}
\TaskLayoutStrip{ToasterSlider}{Core}{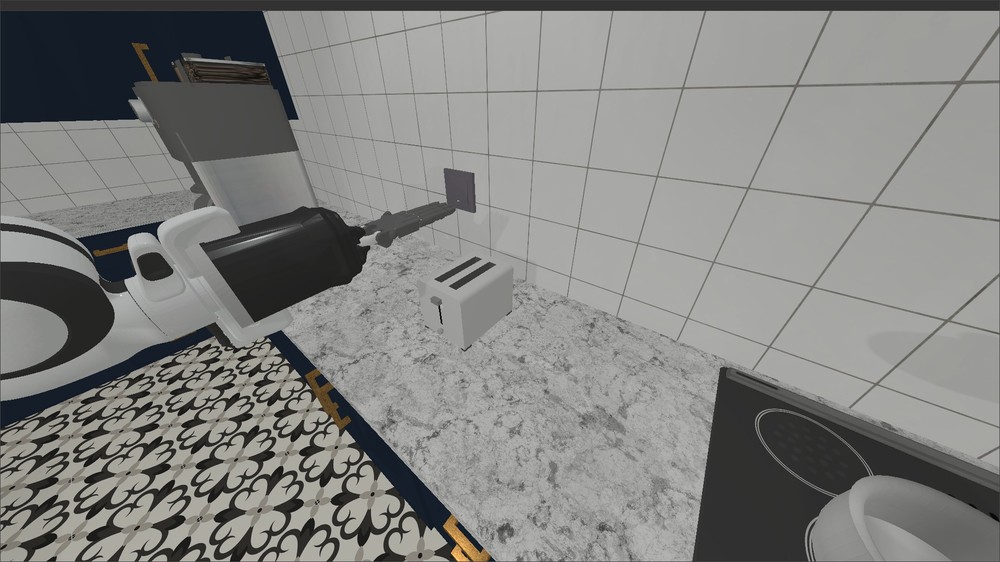}{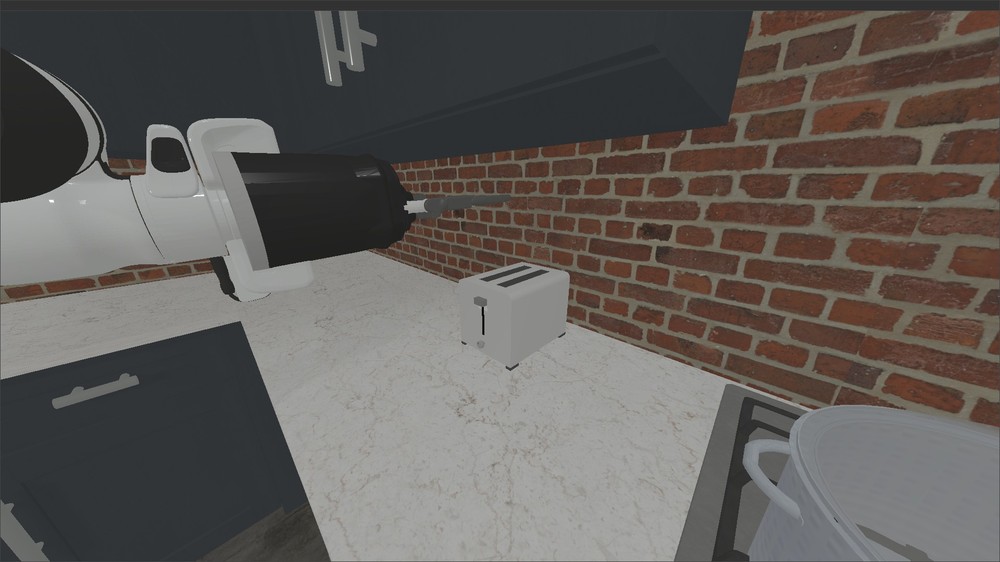}{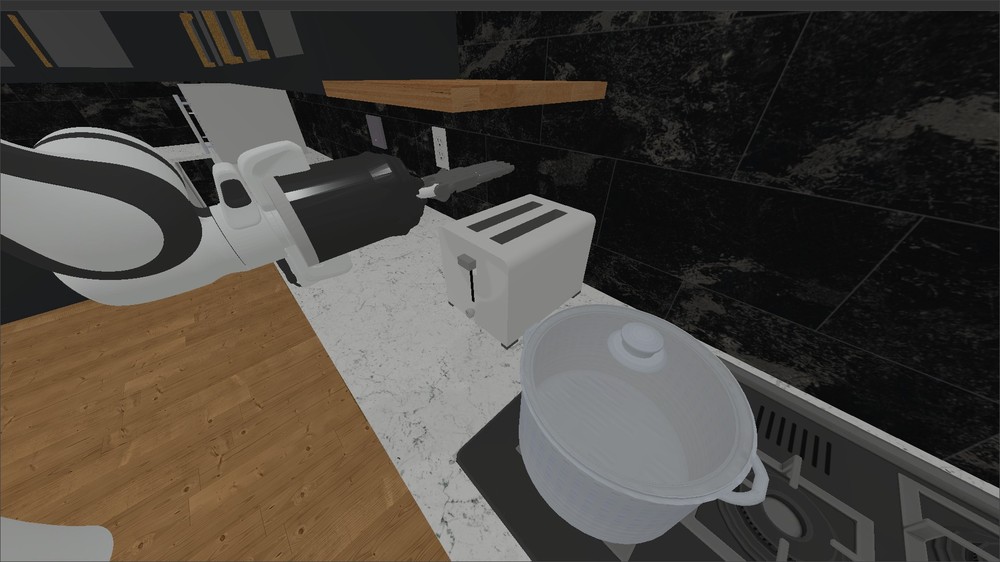}
\TaskLayoutStrip{DisplayPowerButton}{Gen}{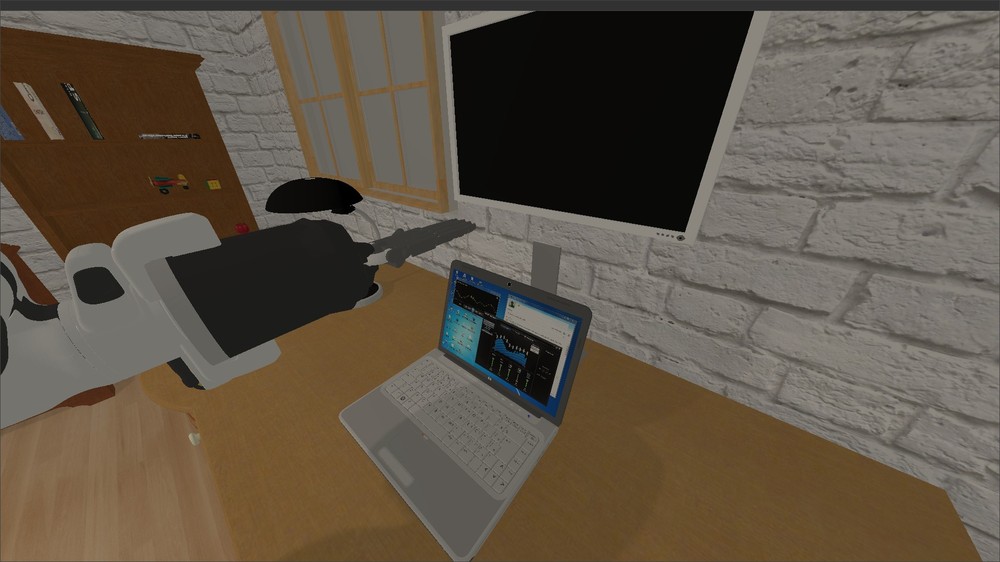}{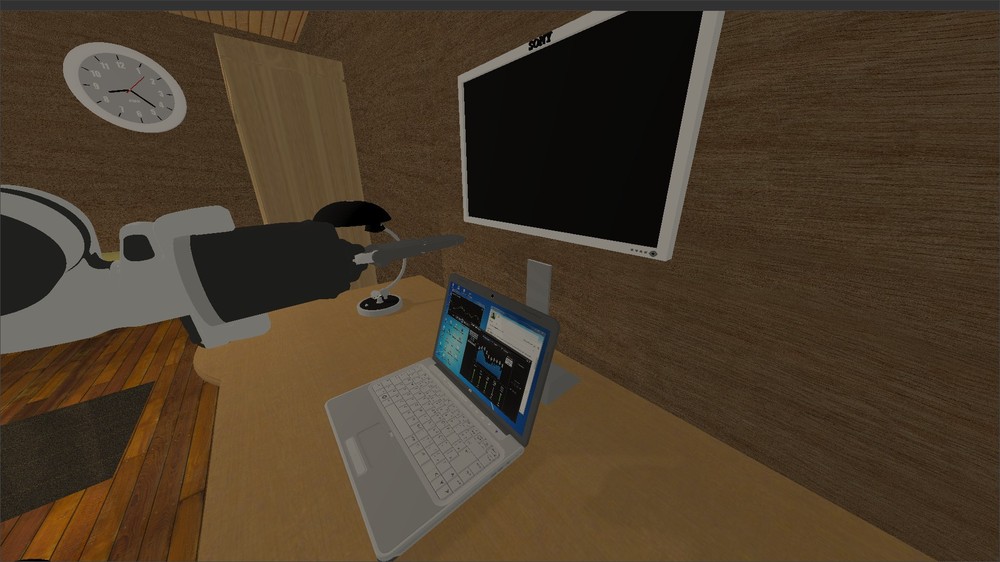}{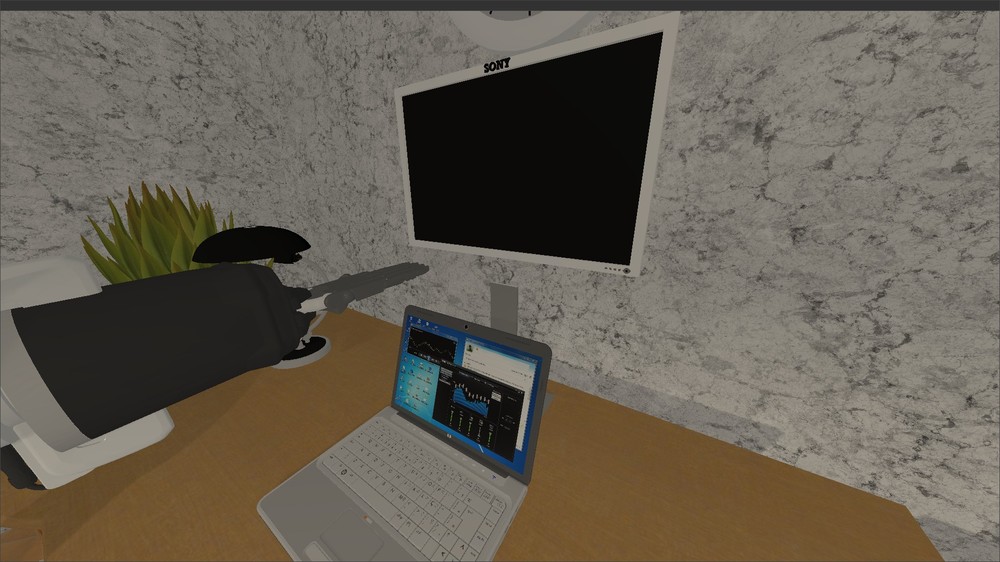}

\LayoutFamilyStart{Rotational manipulation}{FaucetHandle}{Core}{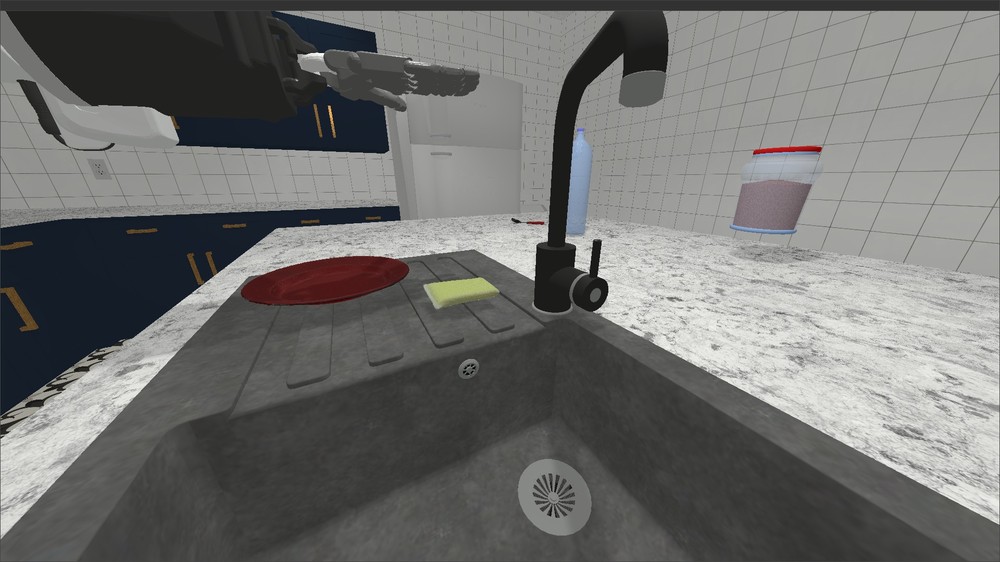}{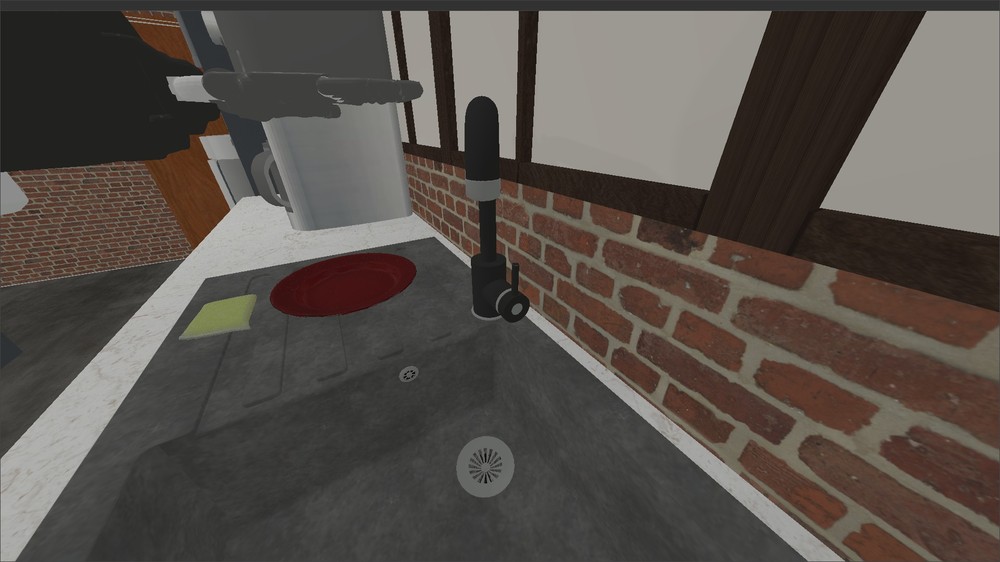}{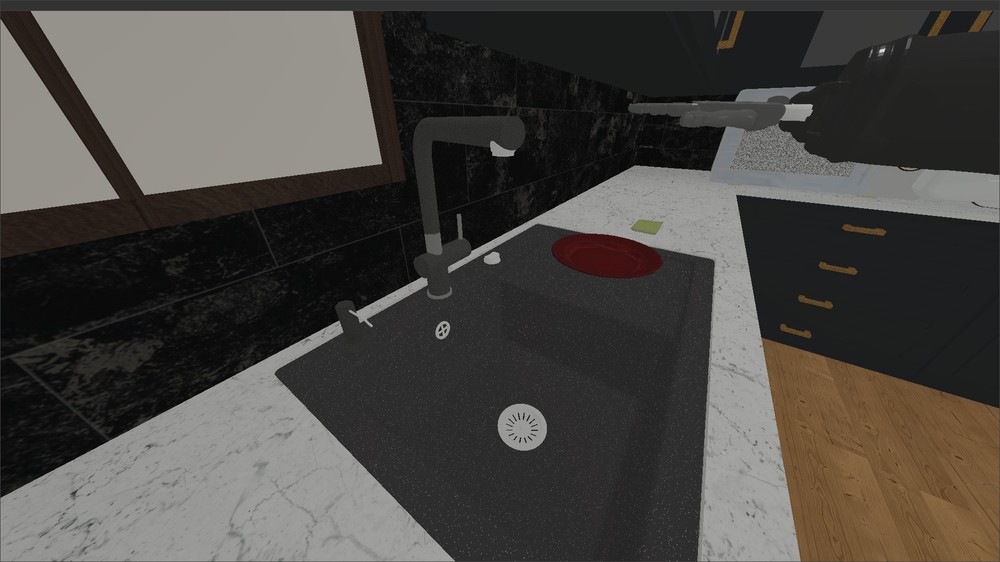}
\TaskLayoutStrip{MouseWheel}{Core}{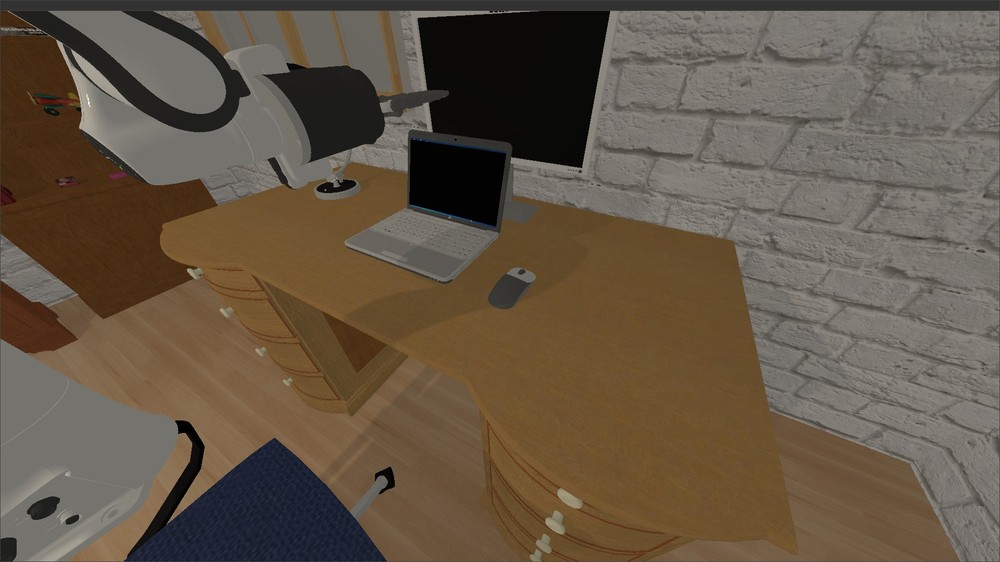}{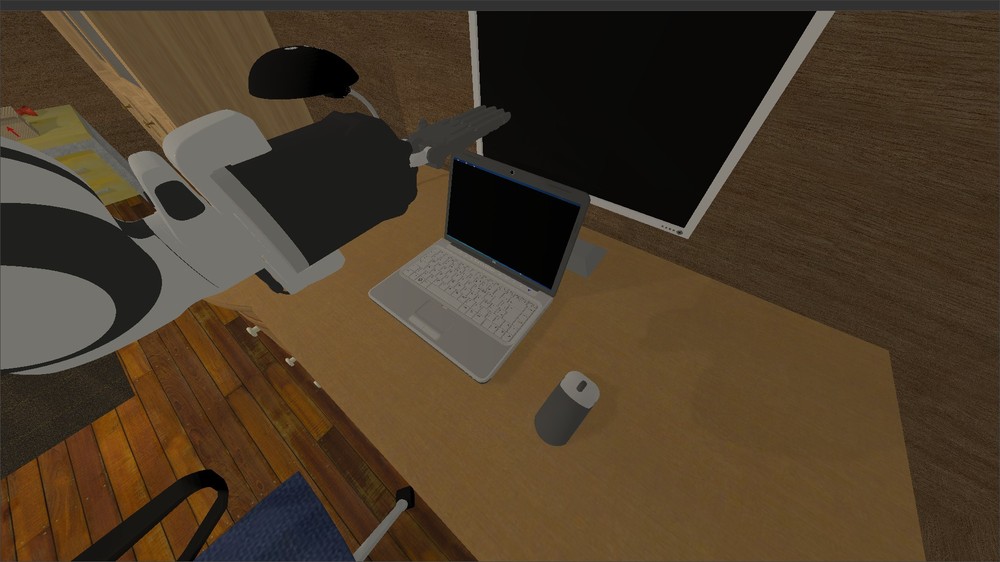}{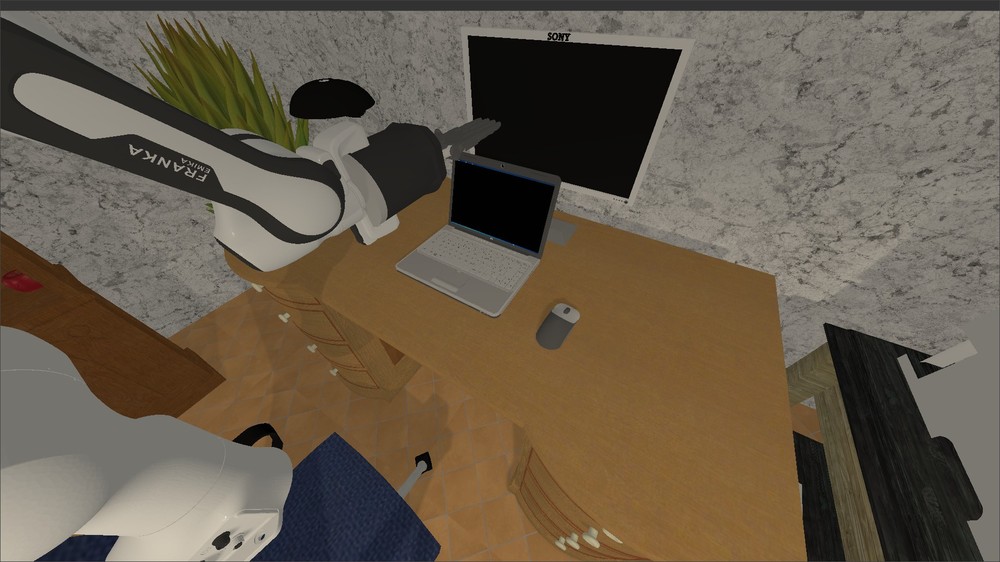}
\TaskLayoutStrip{CoffeeKnob}{Core}{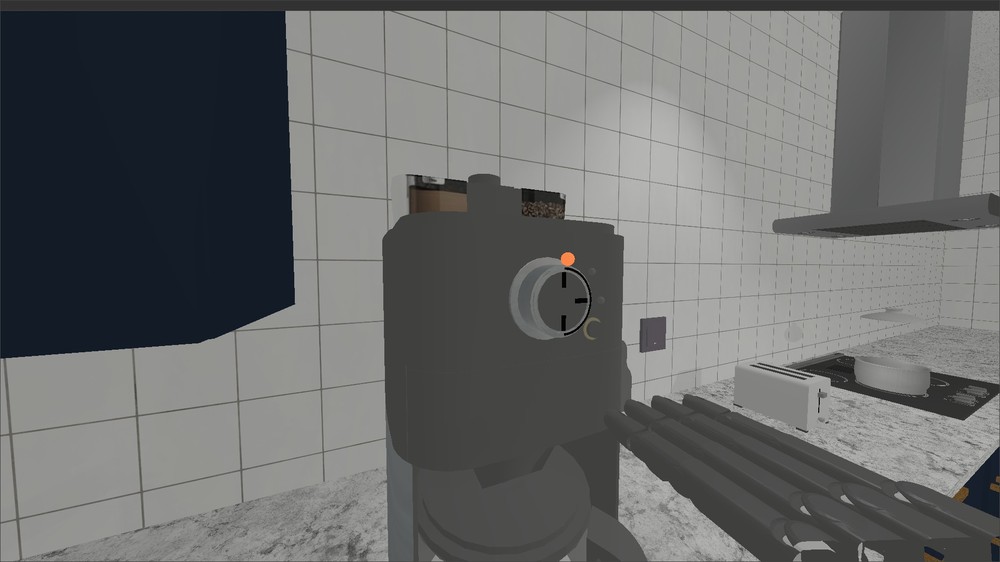}{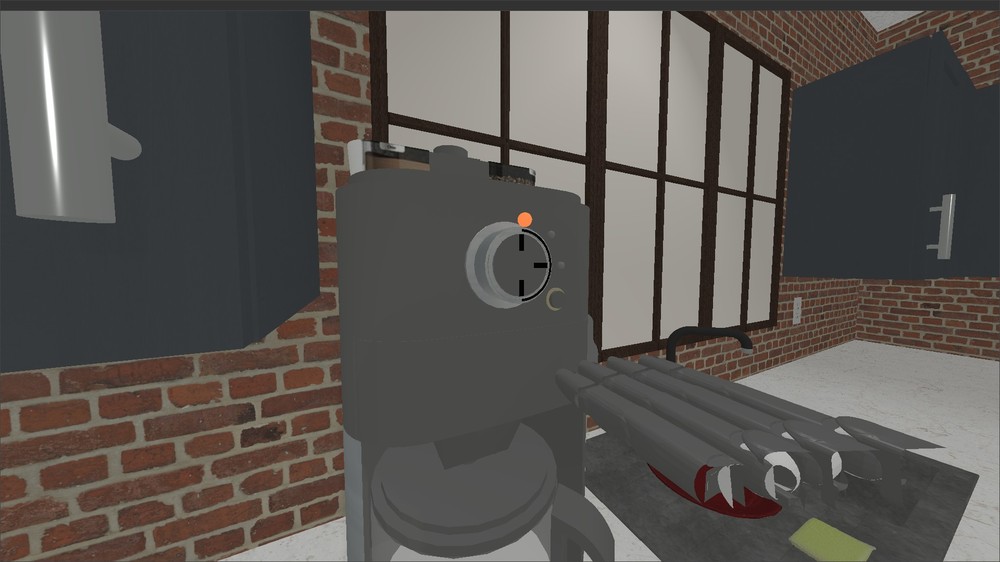}{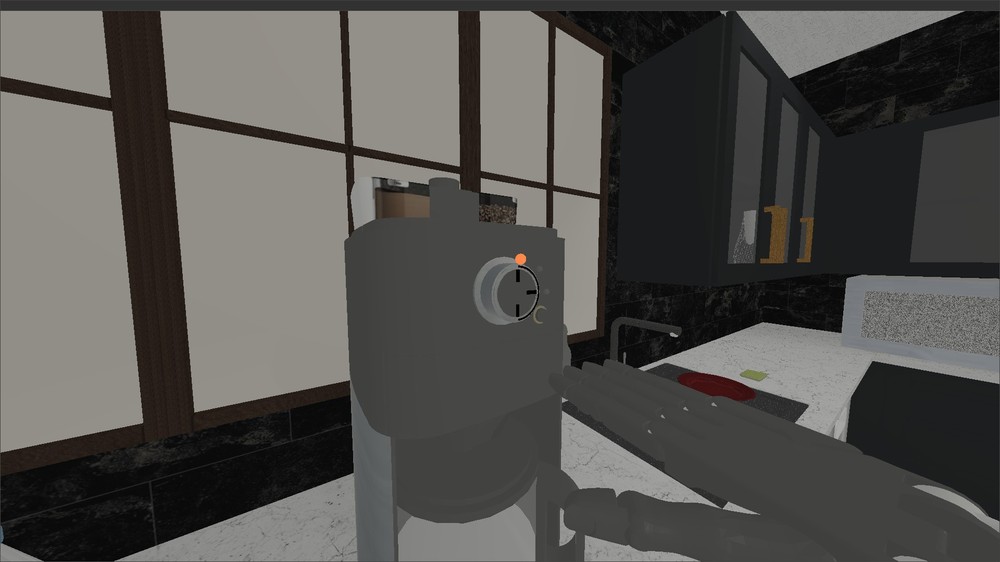}
\TaskLayoutStrip{KettleLidTab}{Core}{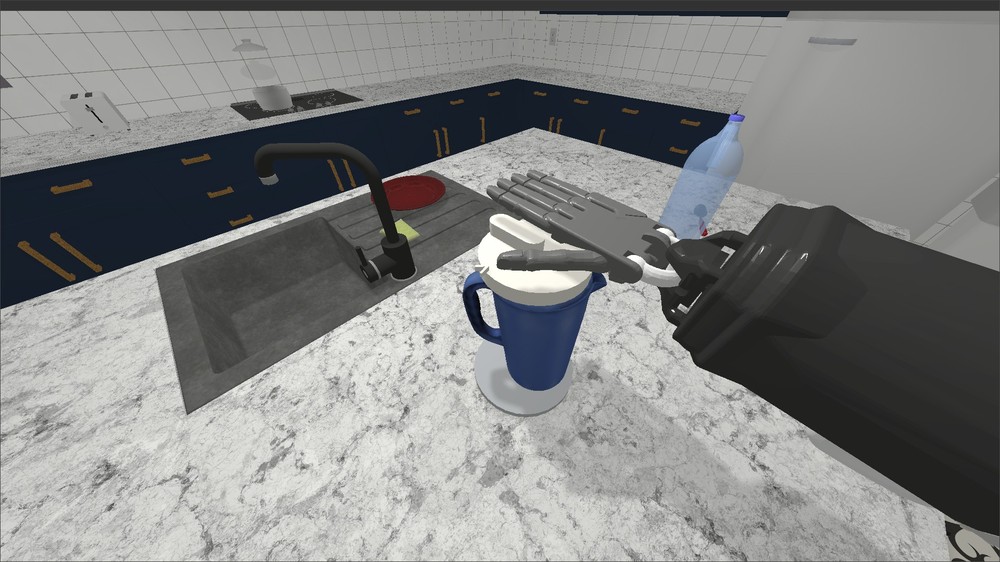}{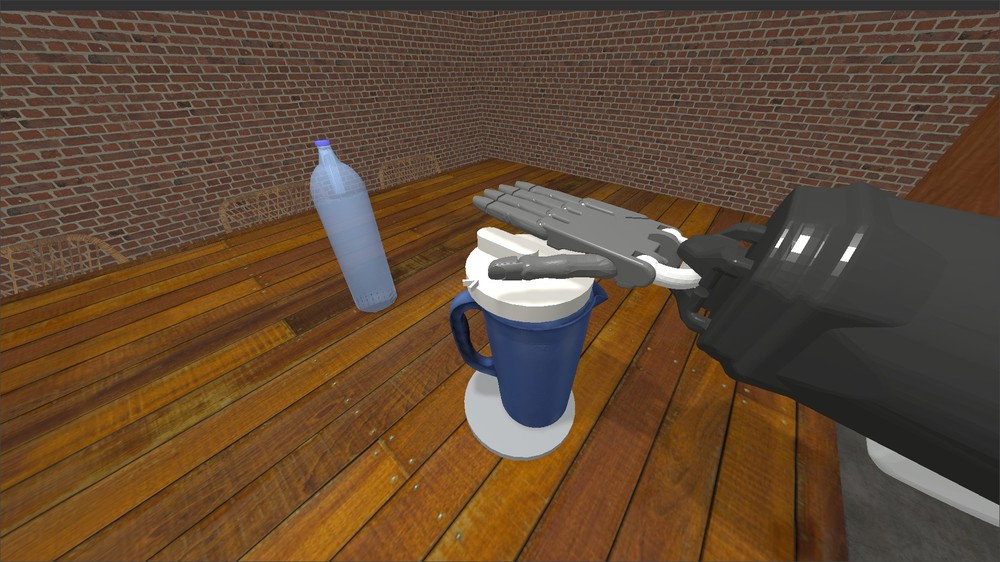}{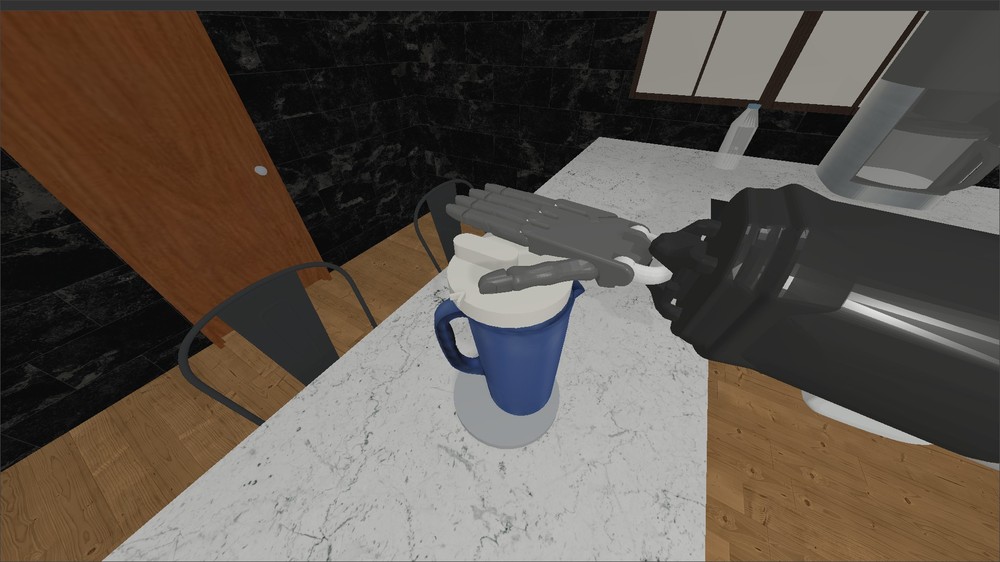}
\TaskLayoutStrip{DishwasherKnob}{Gen}{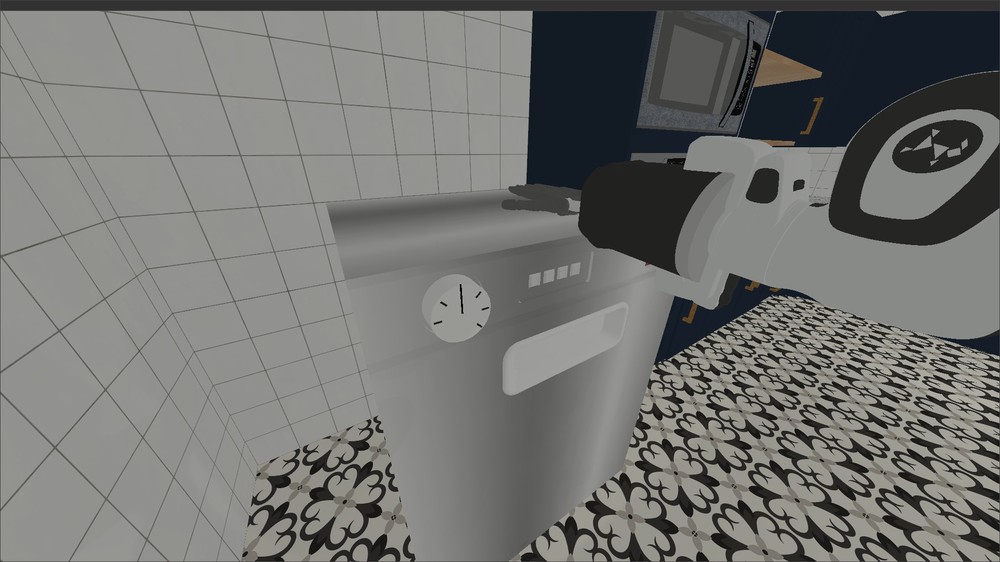}{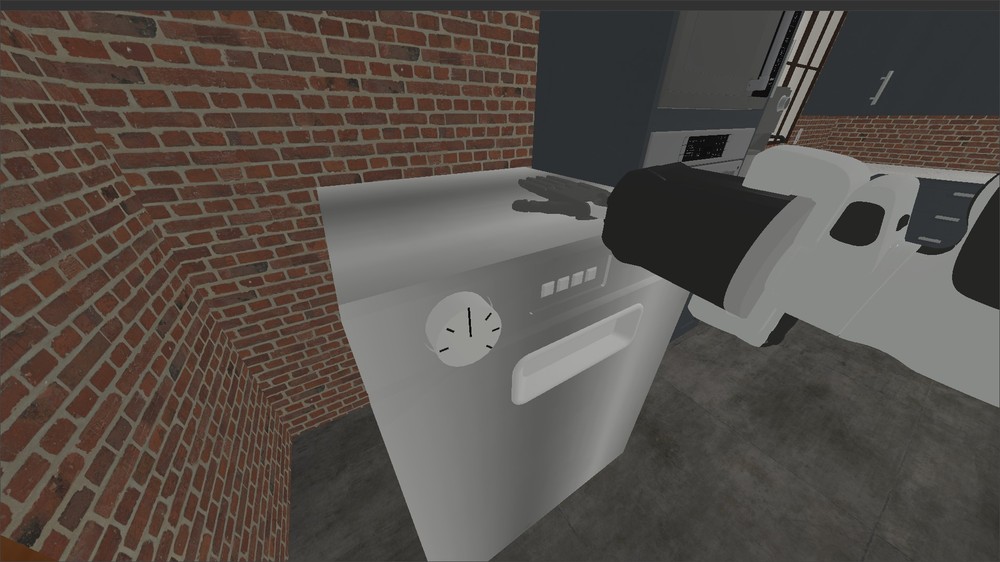}{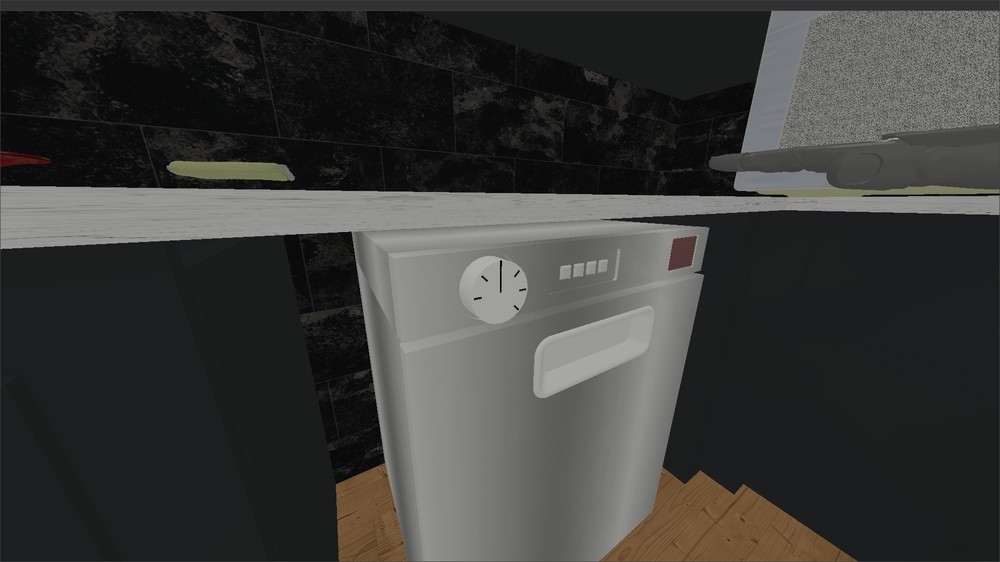}

\LayoutFamilyStart{Access-and-transfer}{CupFromCabinet}{Core}{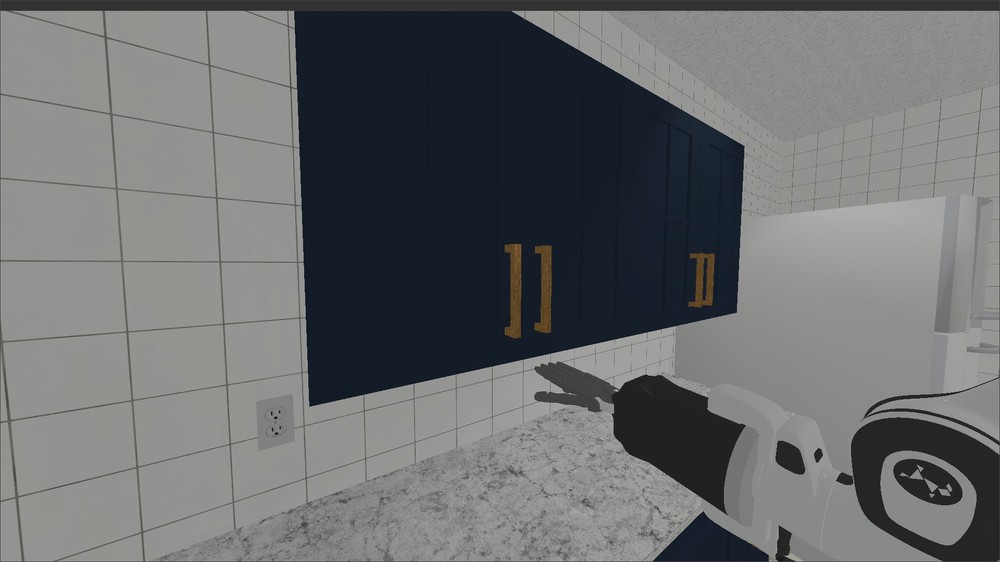}{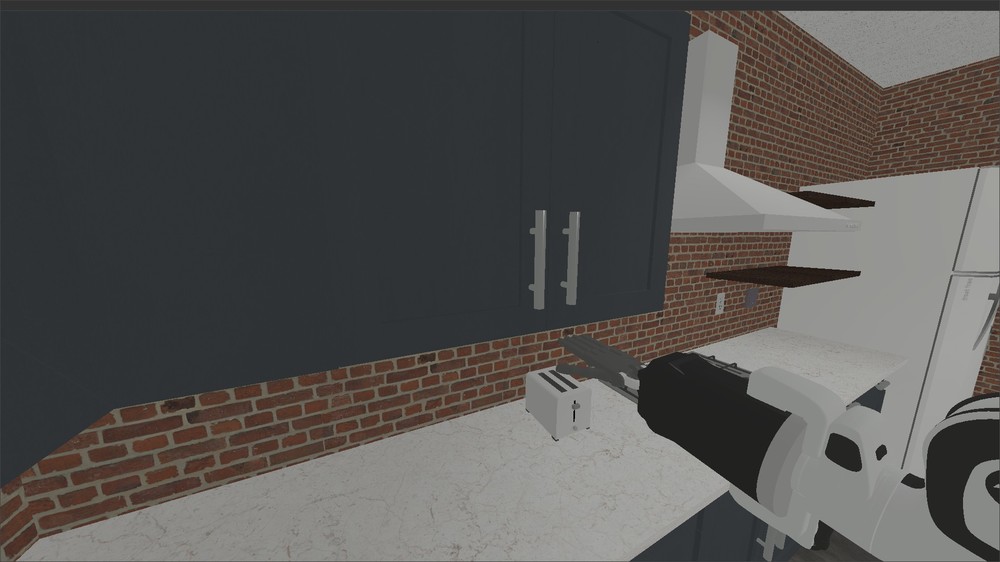}{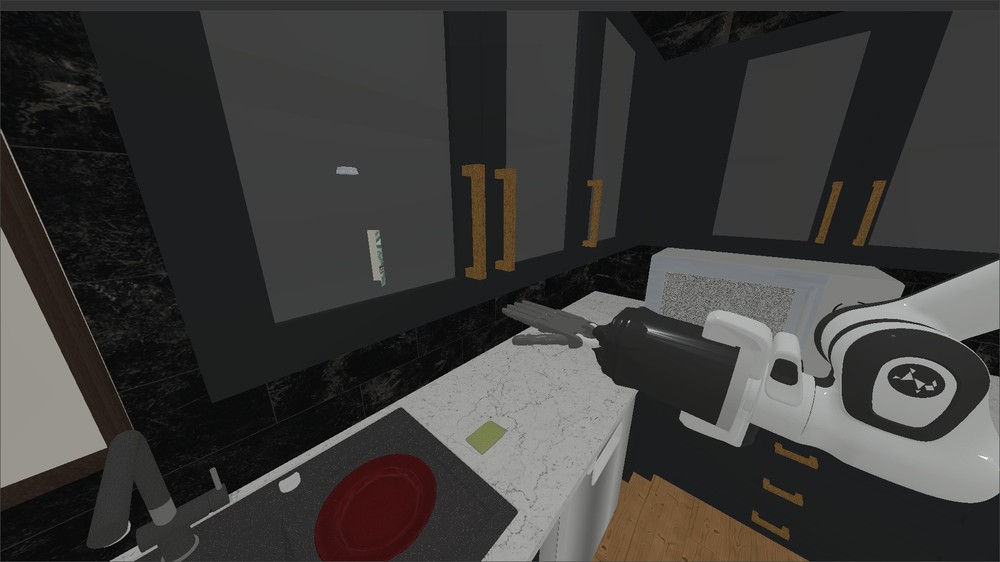}
\TaskLayoutStrip{MicrowaveBowl}{Core}{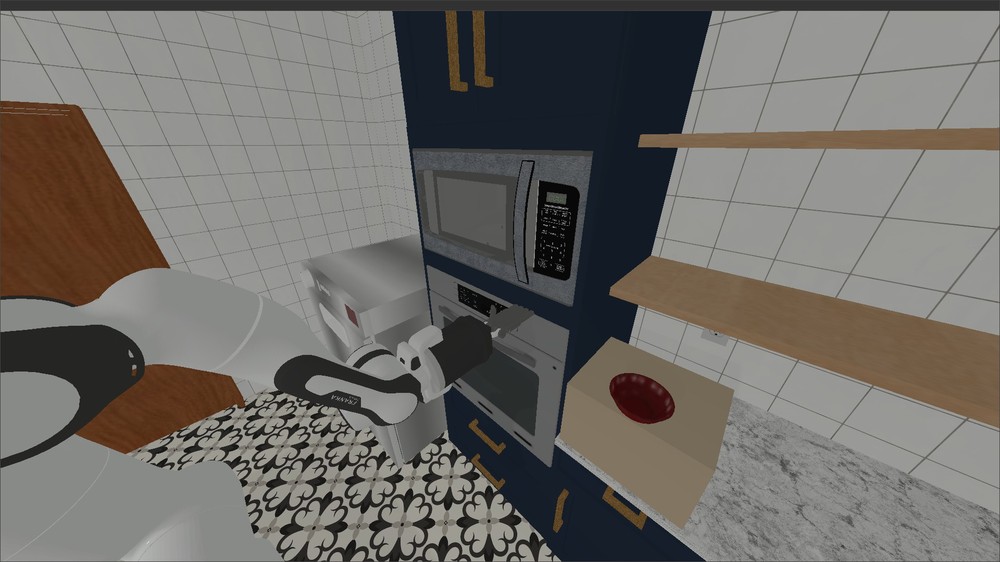}{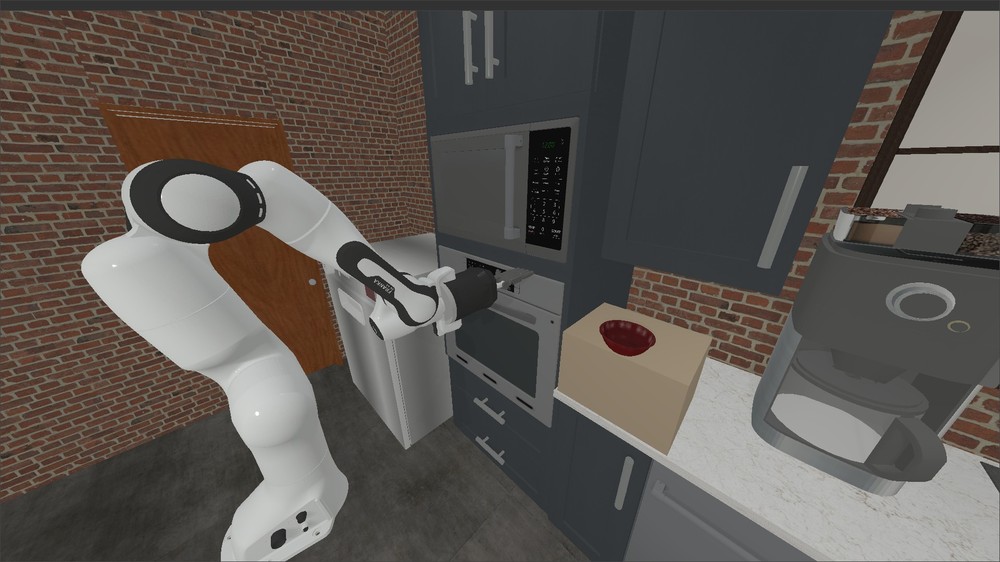}{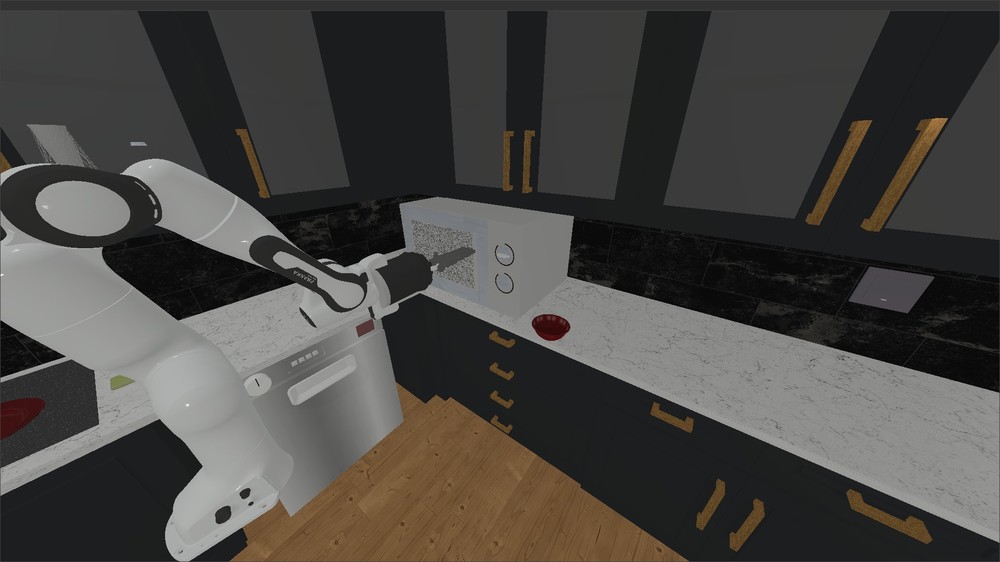}
\TaskLayoutStrip{TennisBallTransfer}{Core}{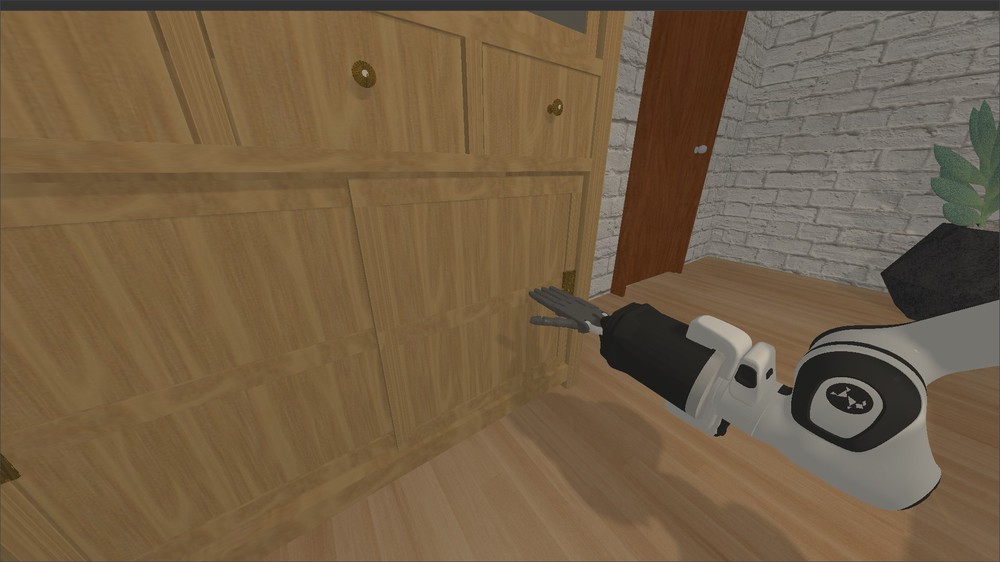}{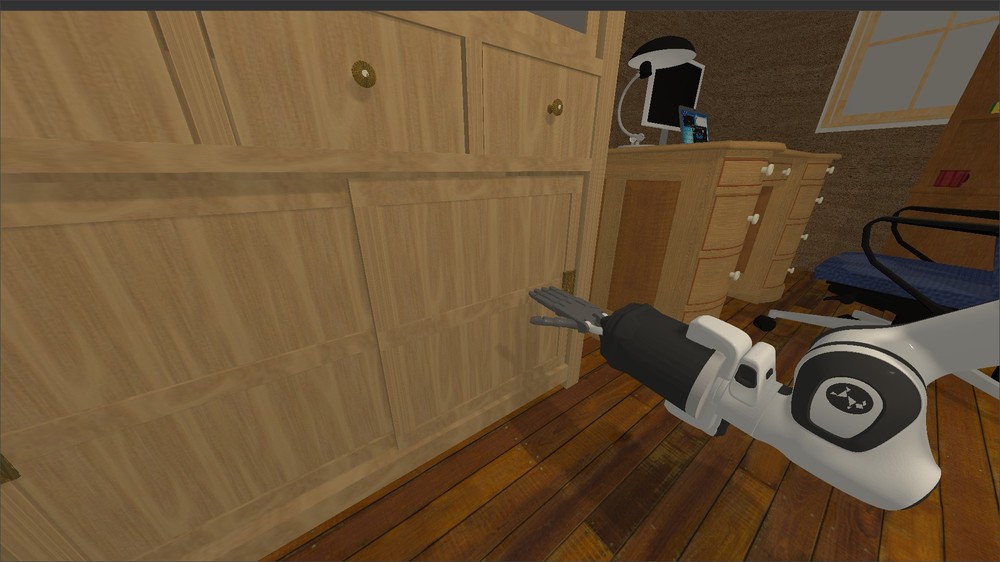}{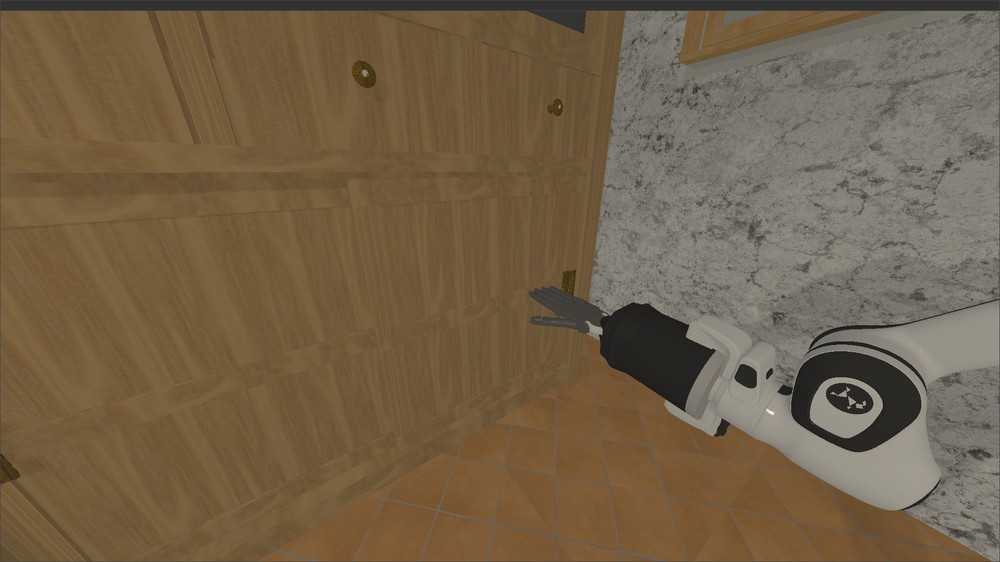}
\TaskLayoutStrip{BananaFromDrawer}{Core}{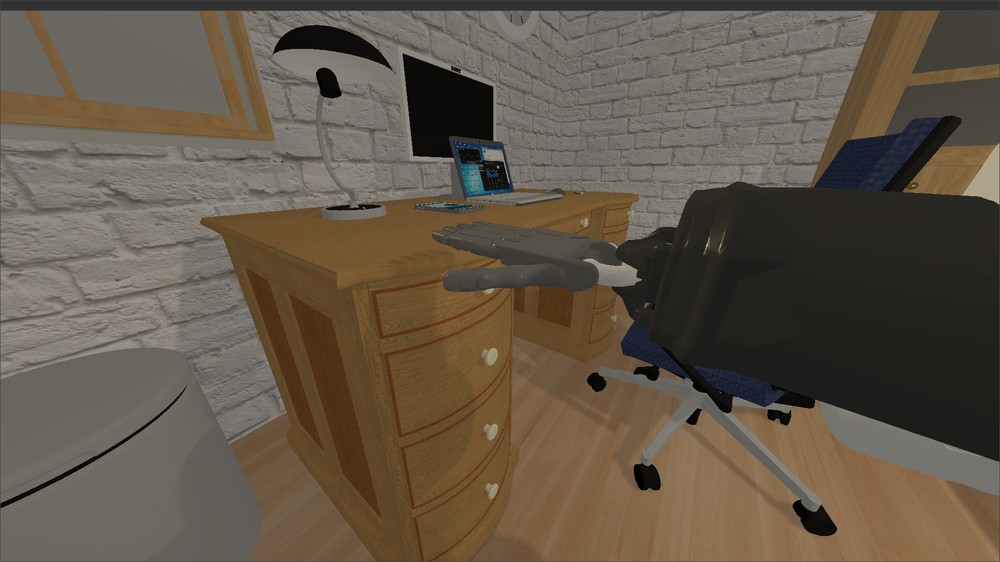}{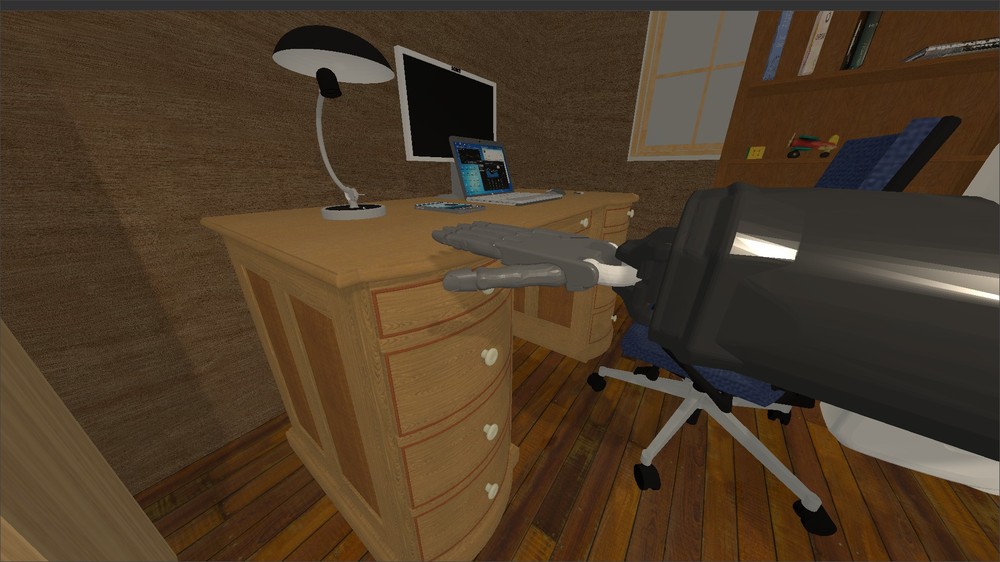}{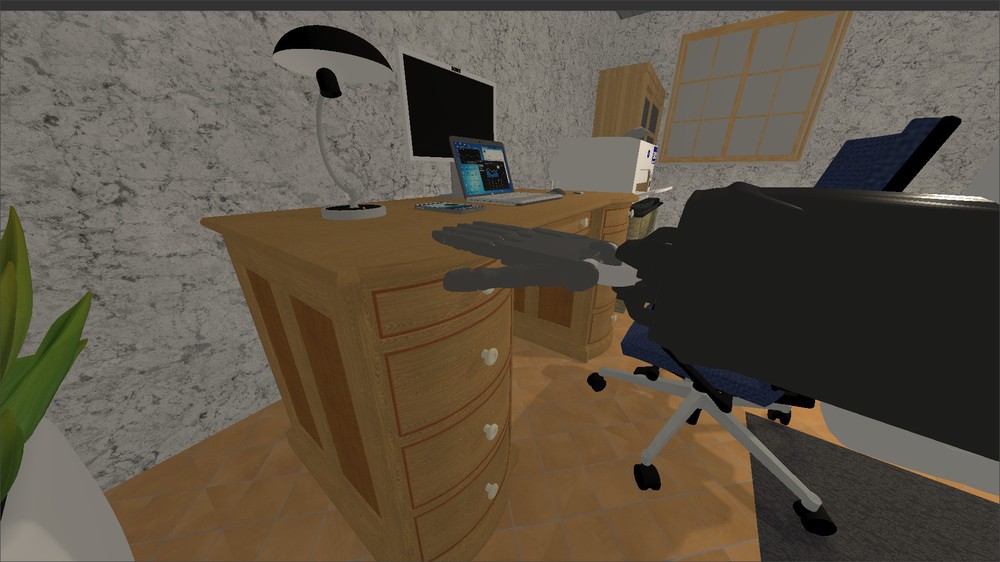}
\TaskLayoutStrip{CandleToCabinet}{Gen}{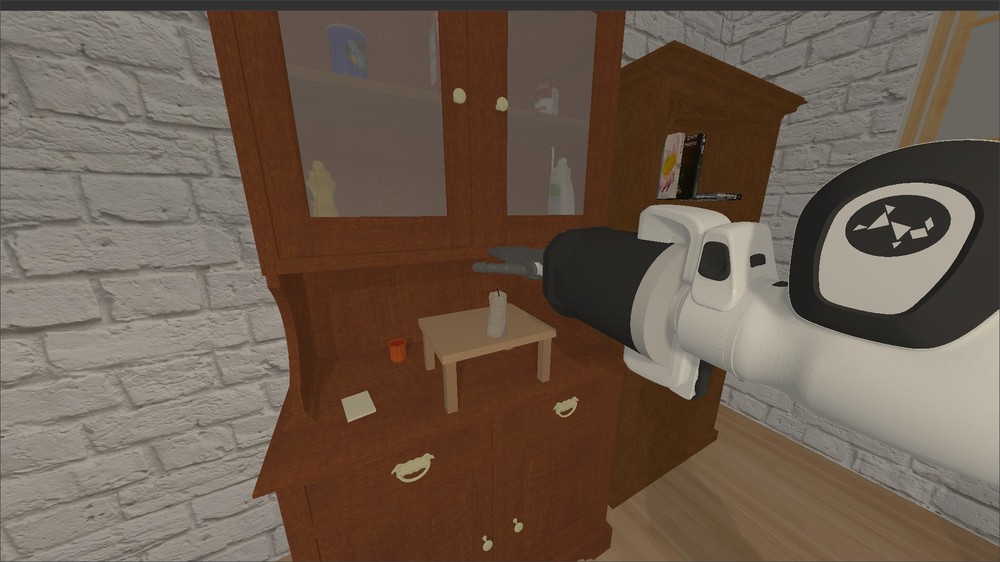}{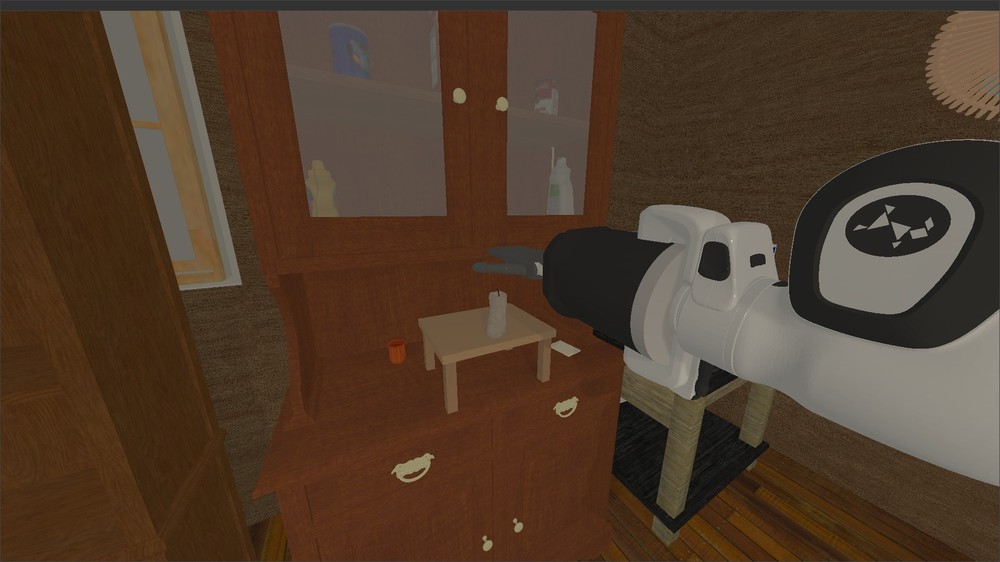}{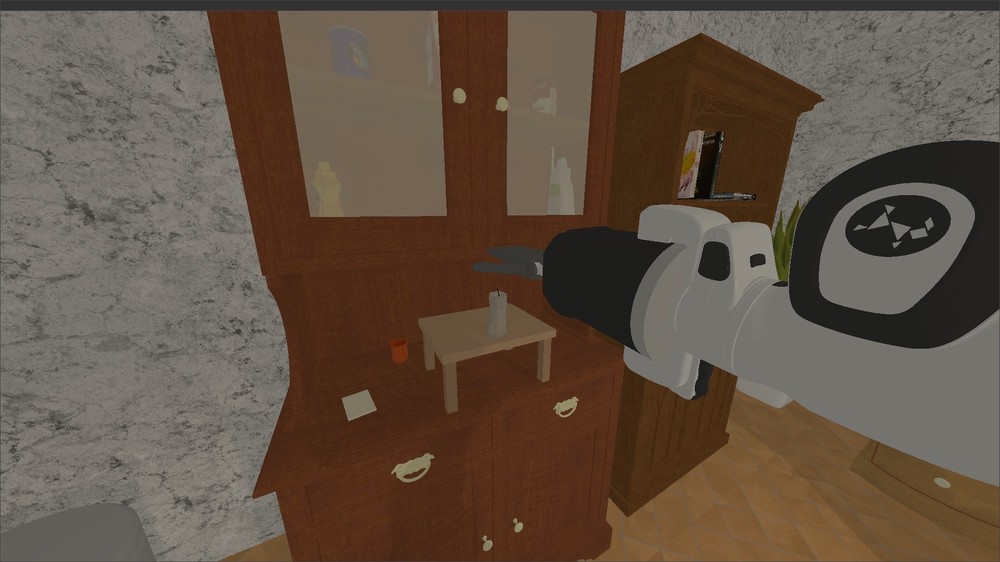}

\LayoutFamilyStart{Structured configuration}{TablewarePlacement}{Core}{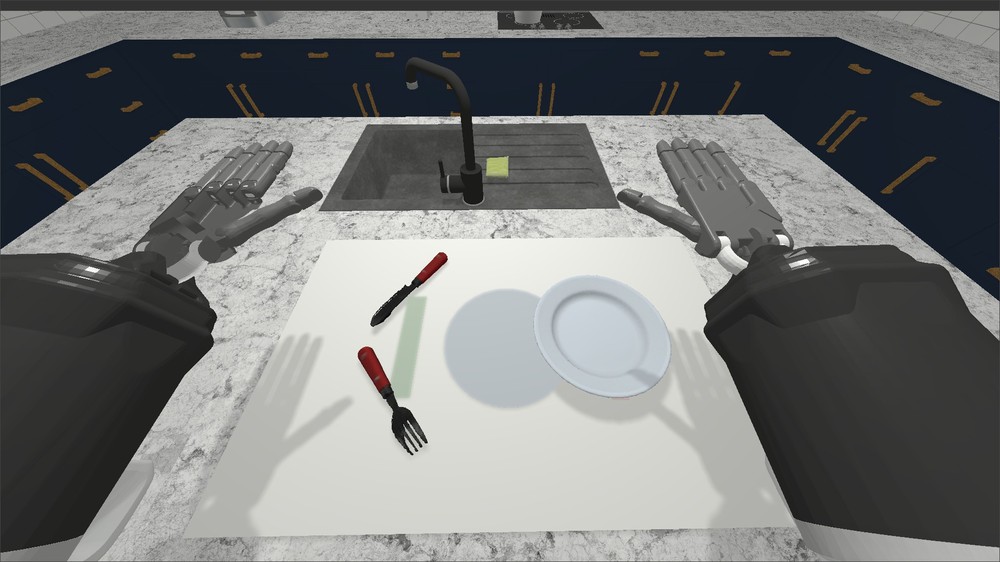}{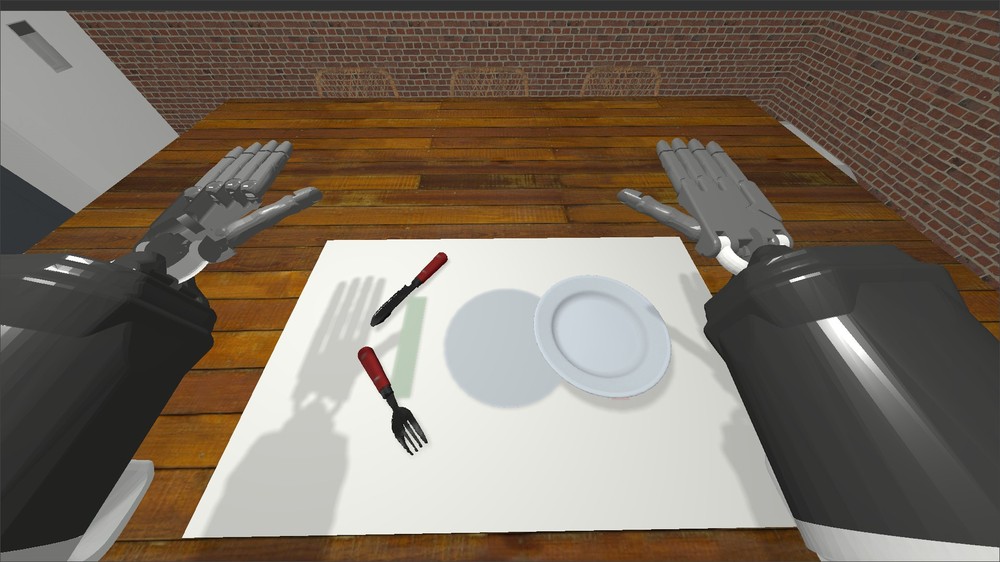}{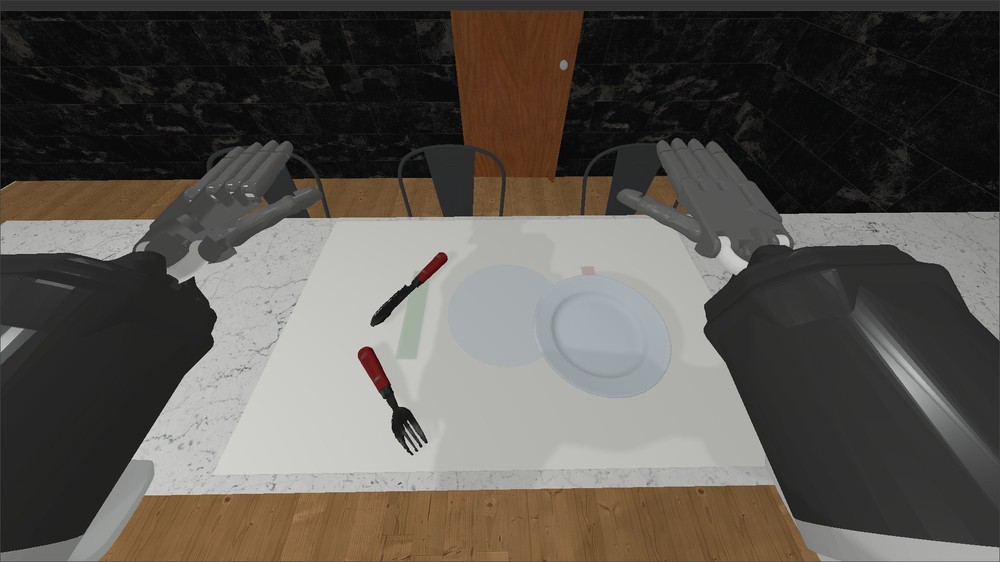}
\TaskLayoutStrip{HammerBlockAssembly}{Core}{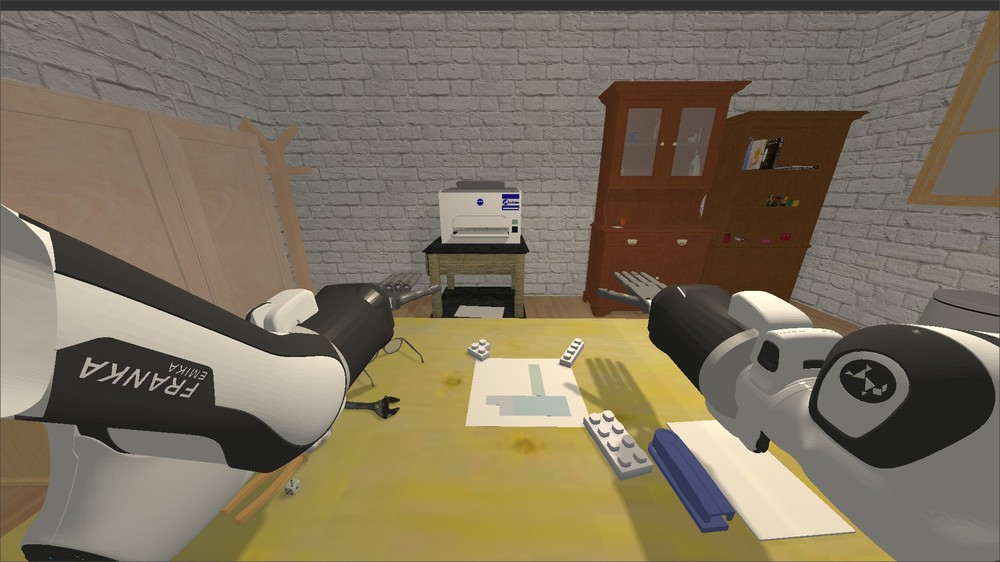}{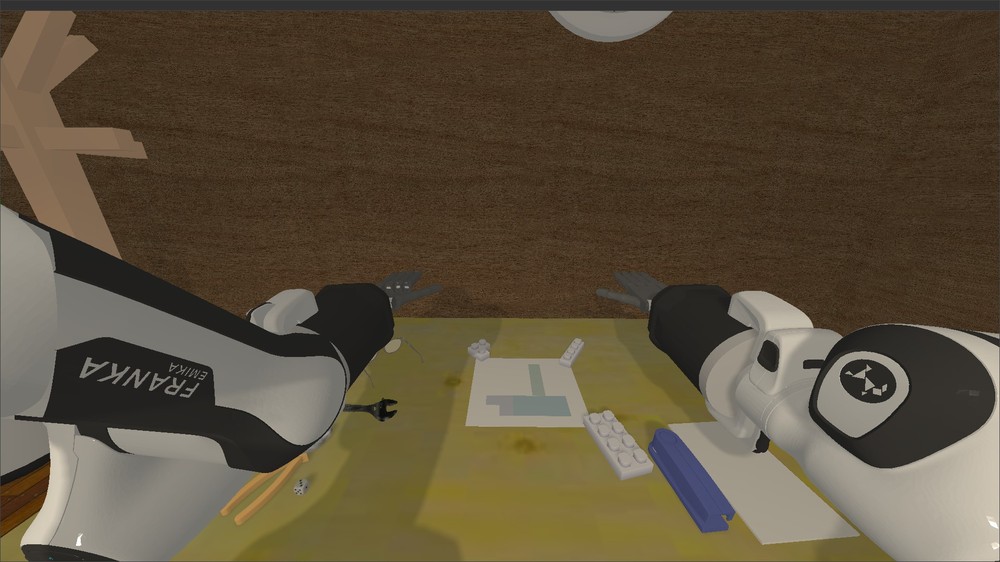}{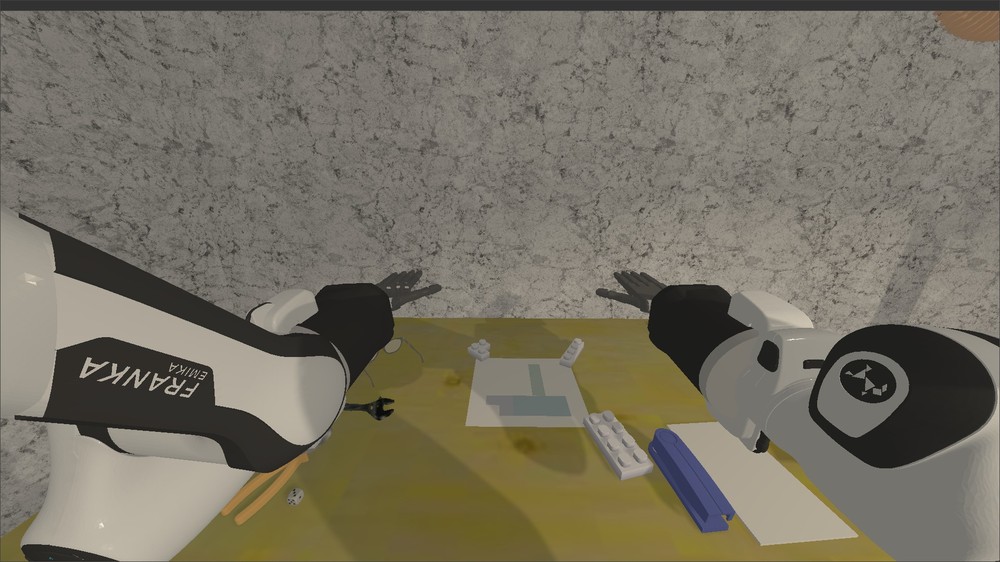}
\TaskLayoutStrip{SandwichAssembly}{Core}{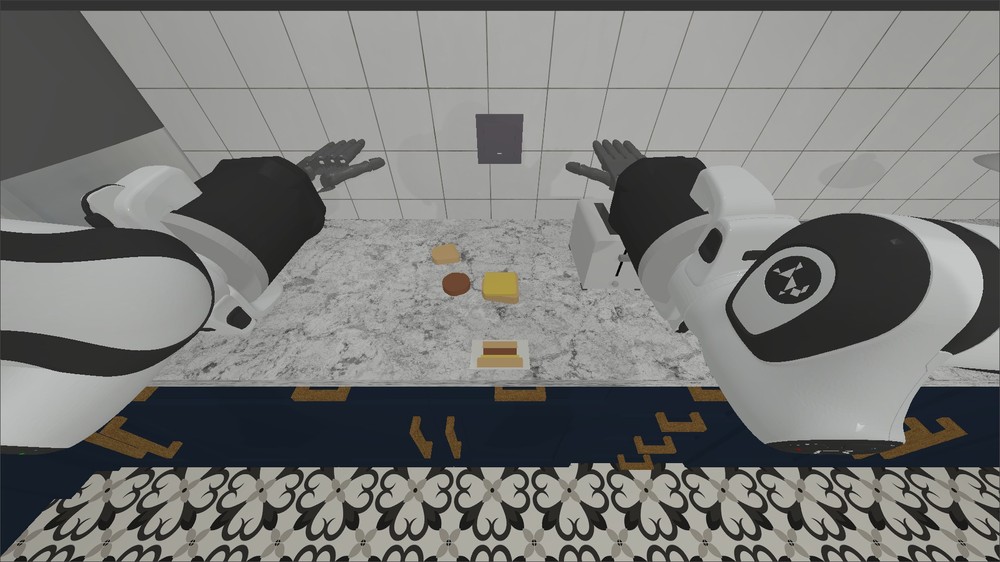}{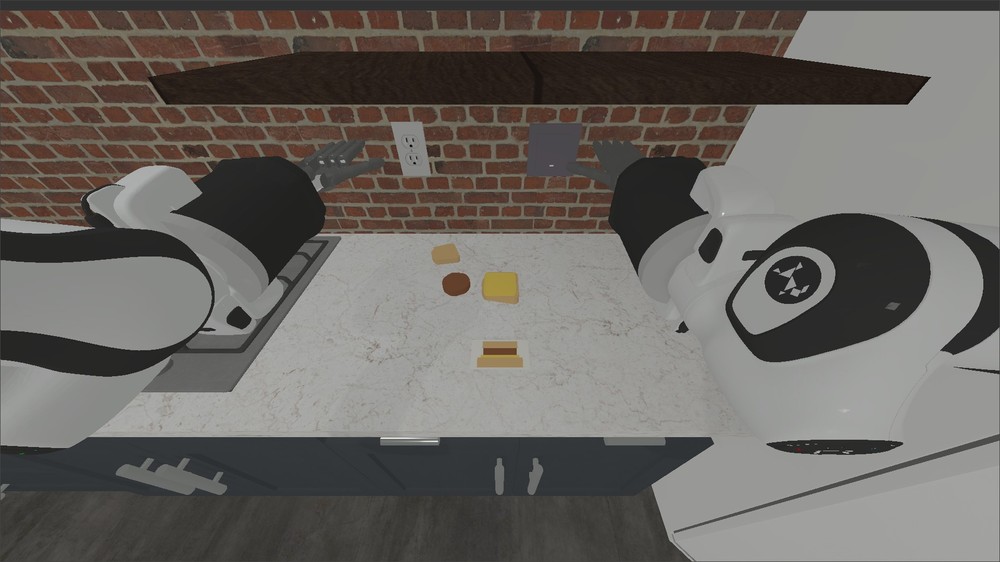}{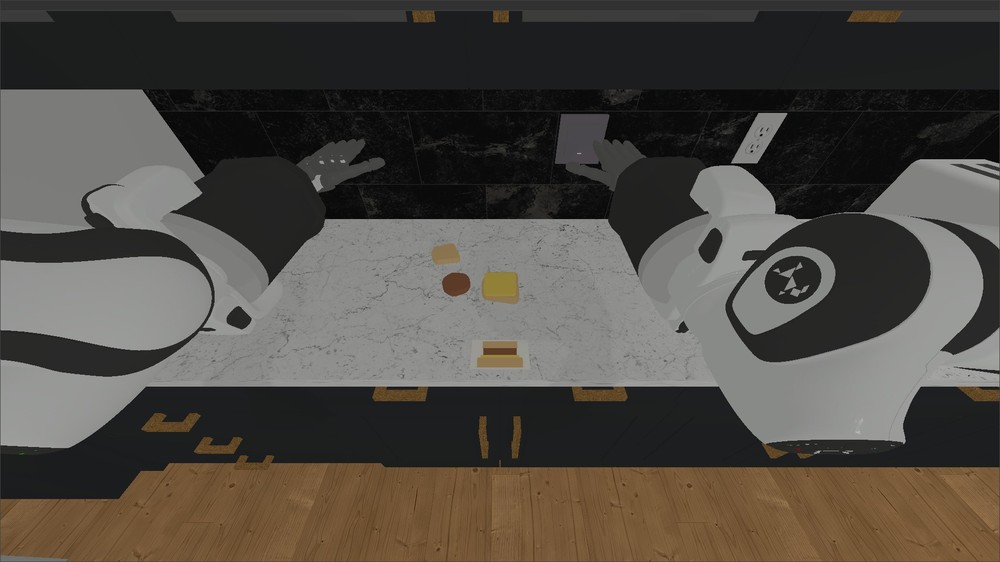}
\TaskLayoutStrip{RingTowerAssembly}{Core}{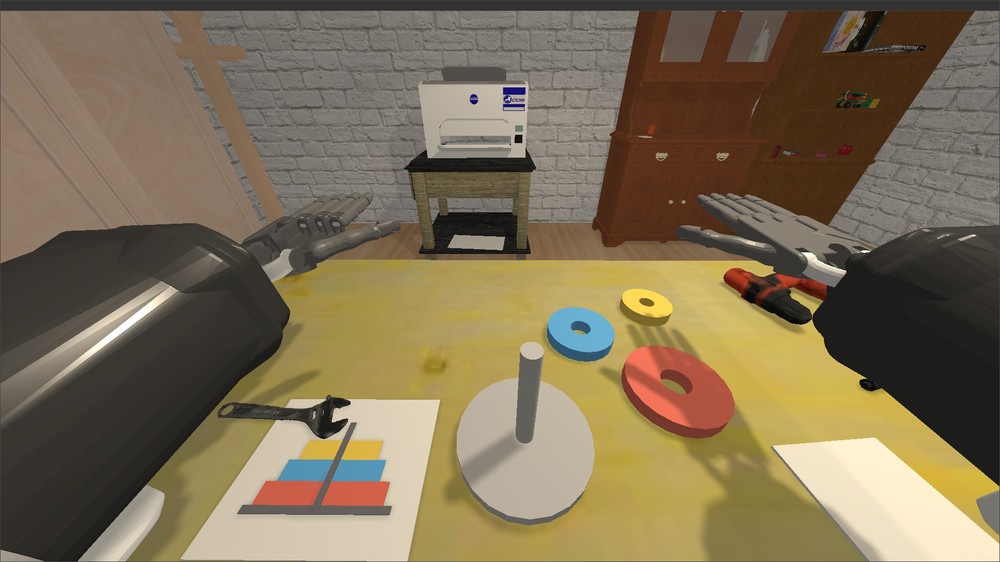}{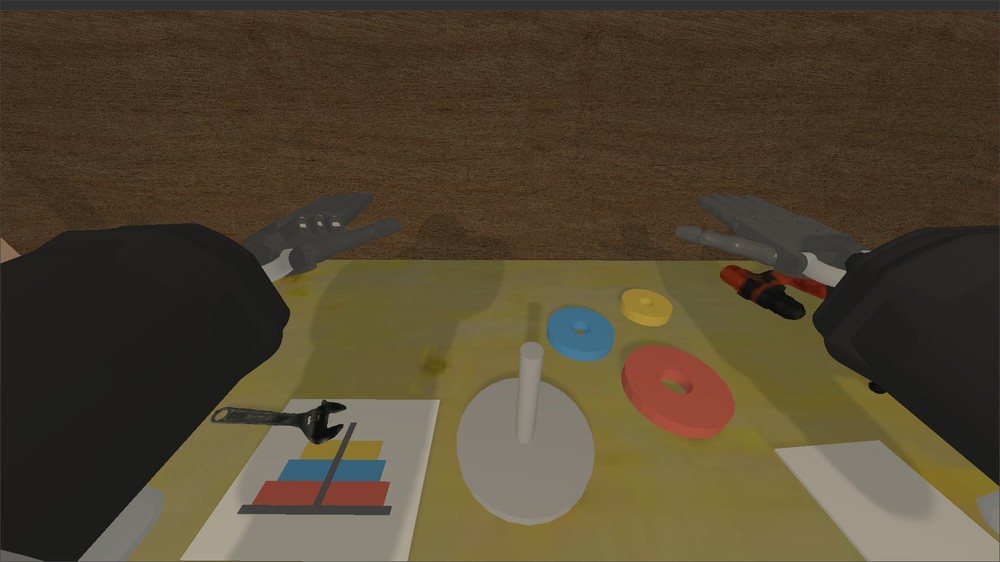}{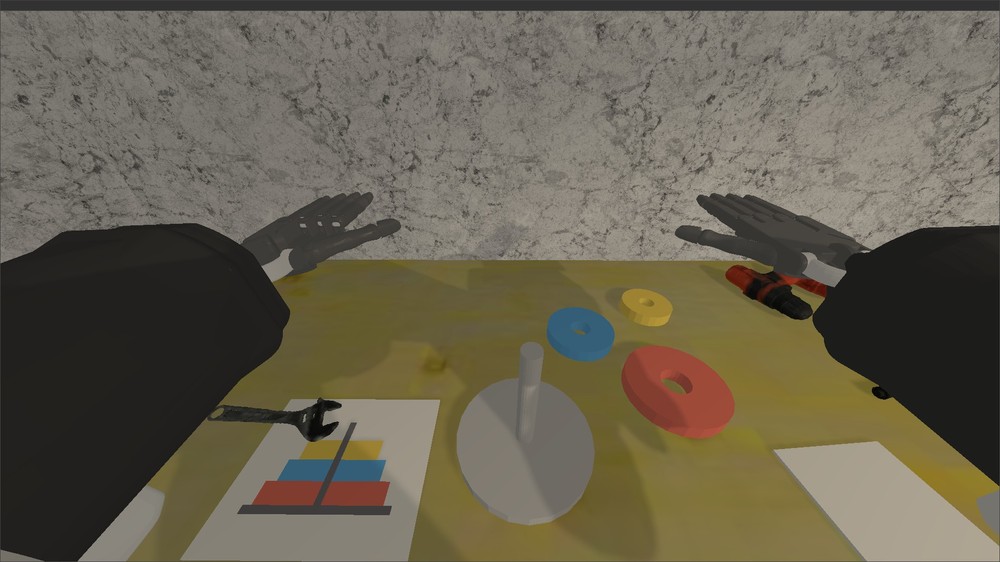}
\TaskLayoutStrip{BoxPacking}{Gen}{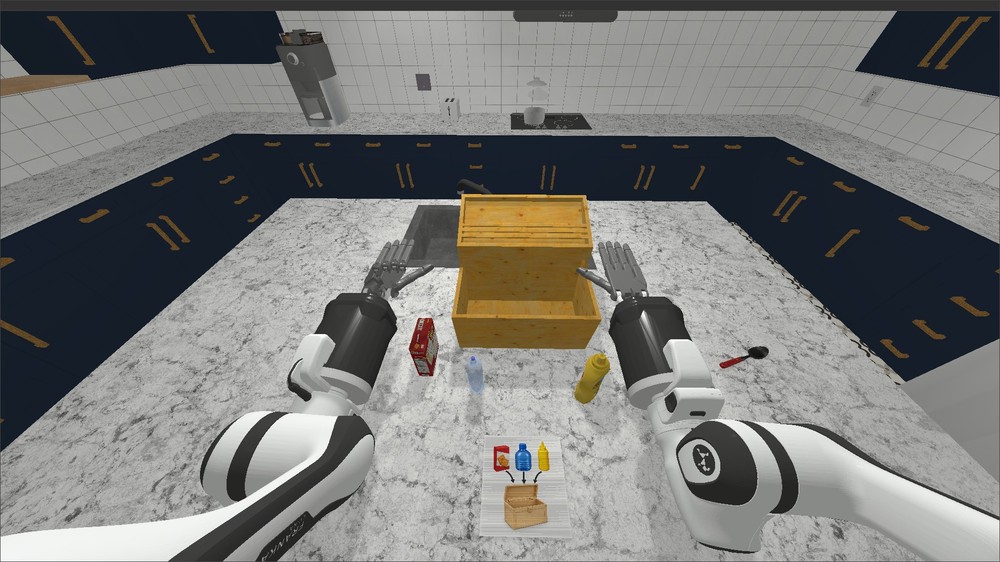}{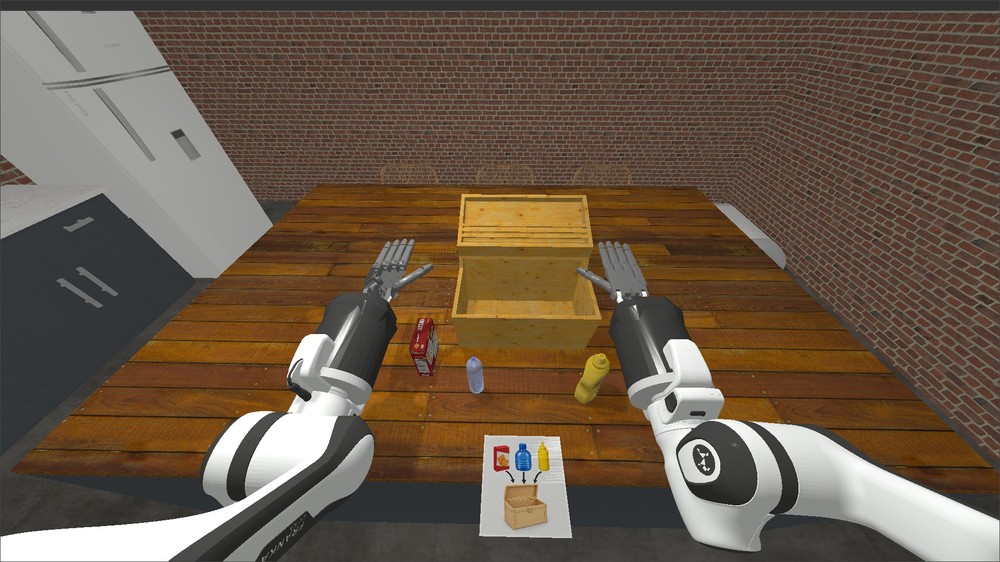}{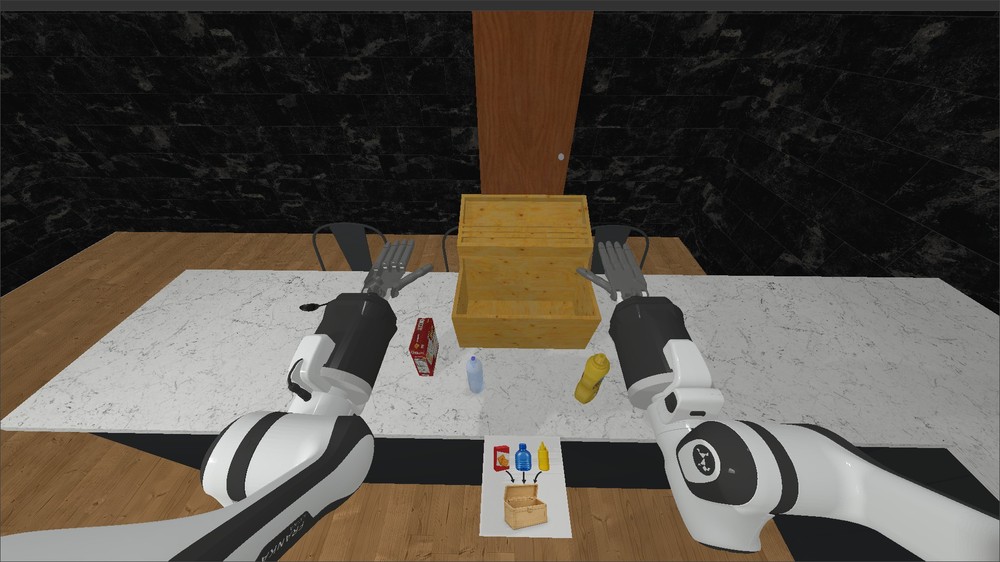}

\LayoutFamilyStart{Assembly}{PenCapping}{Core}{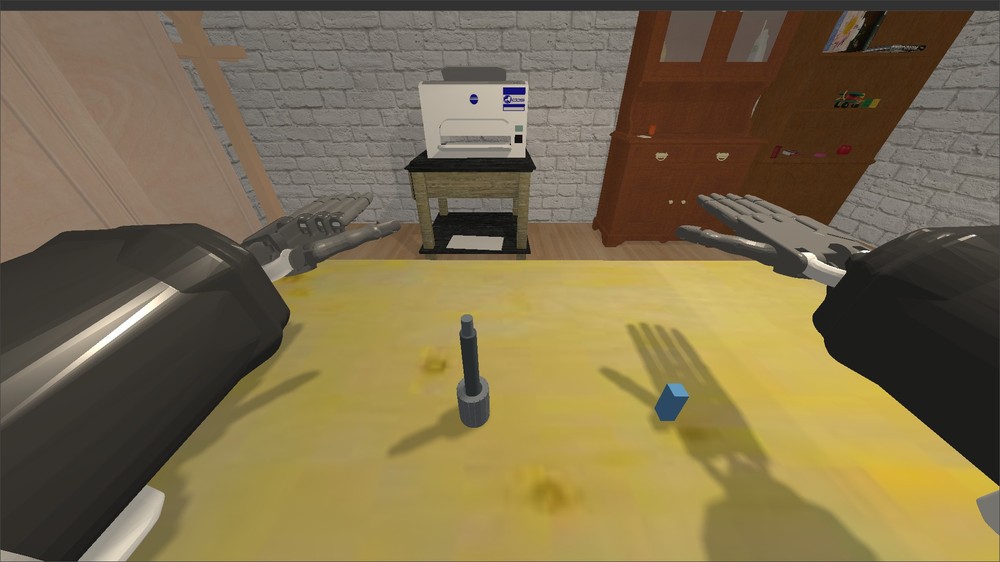}{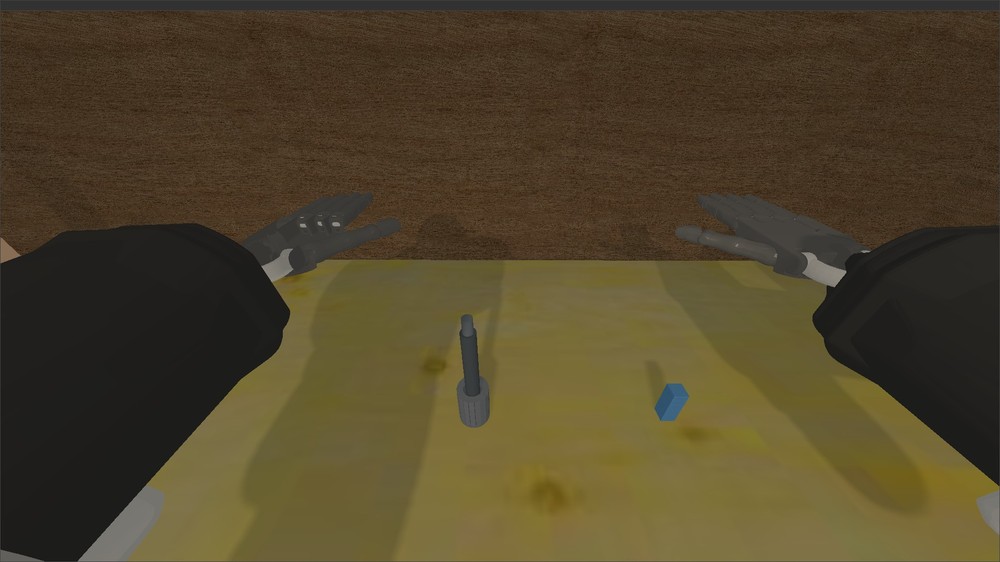}{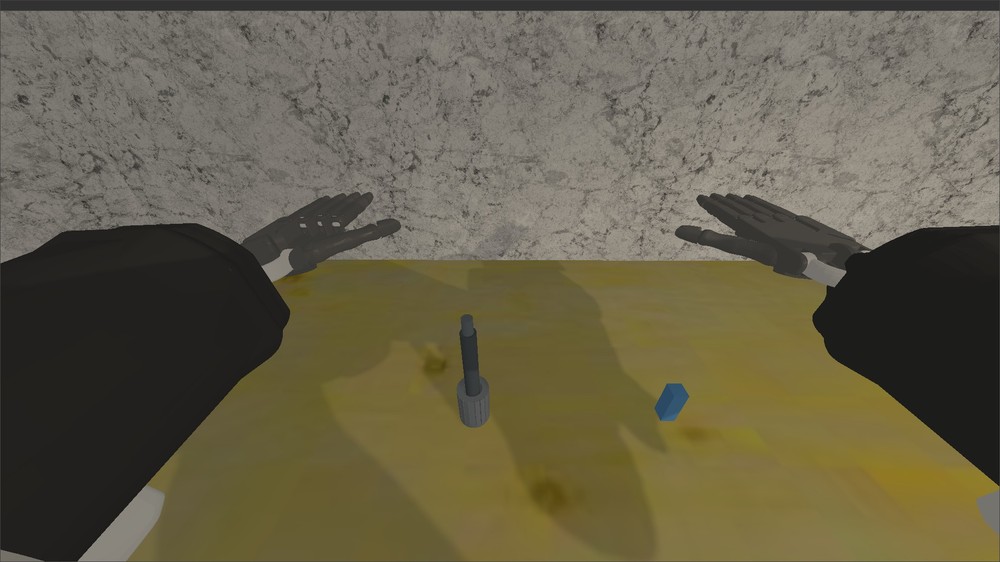}
\TaskLayoutStrip{CupSleeveFit}{Core}{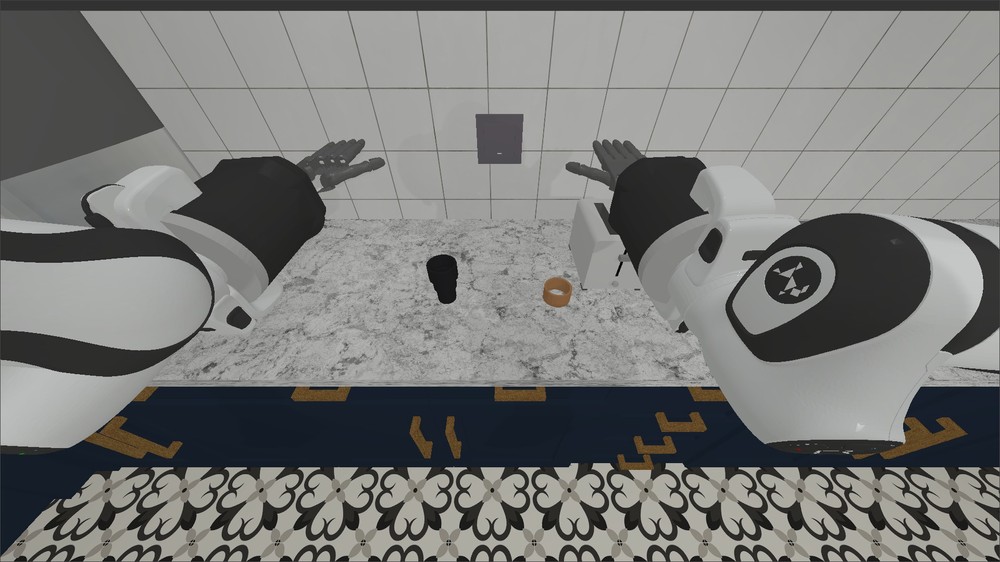}{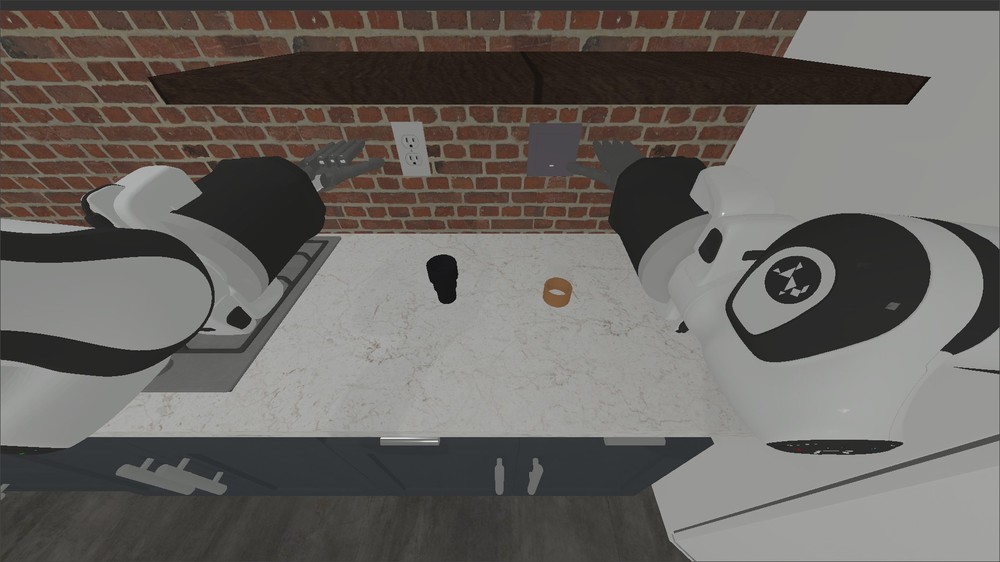}{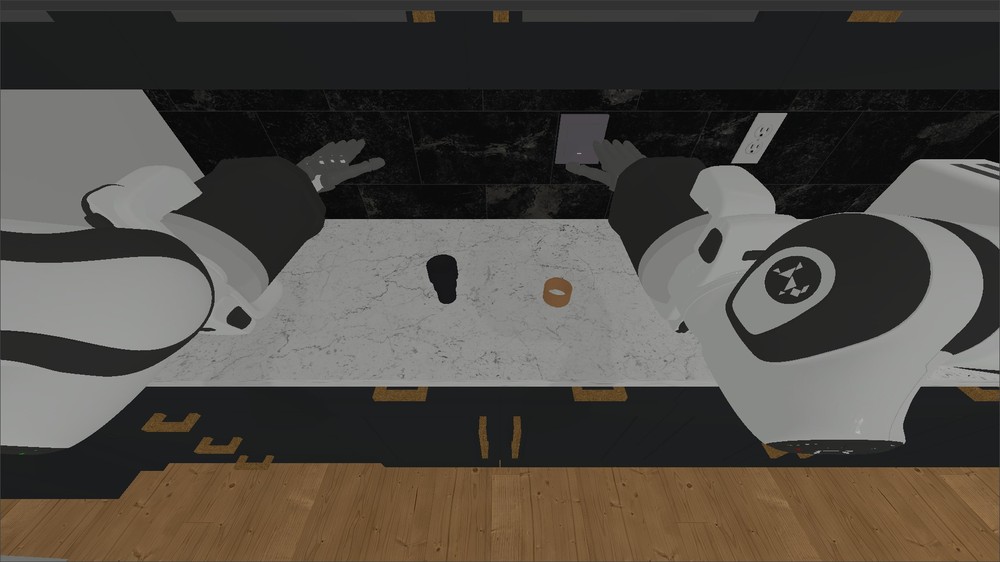}
\TaskLayoutStrip{BottleFunnelInsertion}{Core}{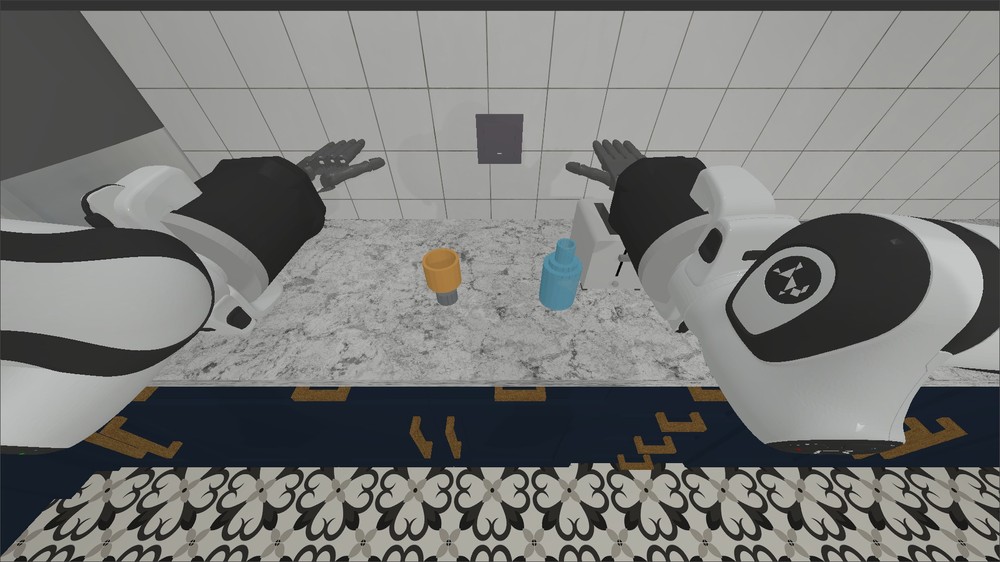}{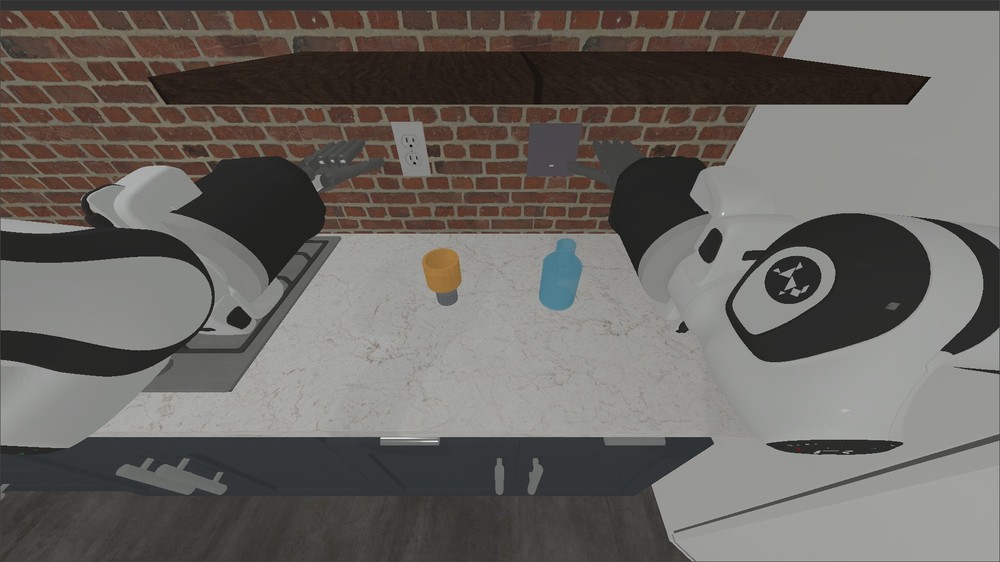}{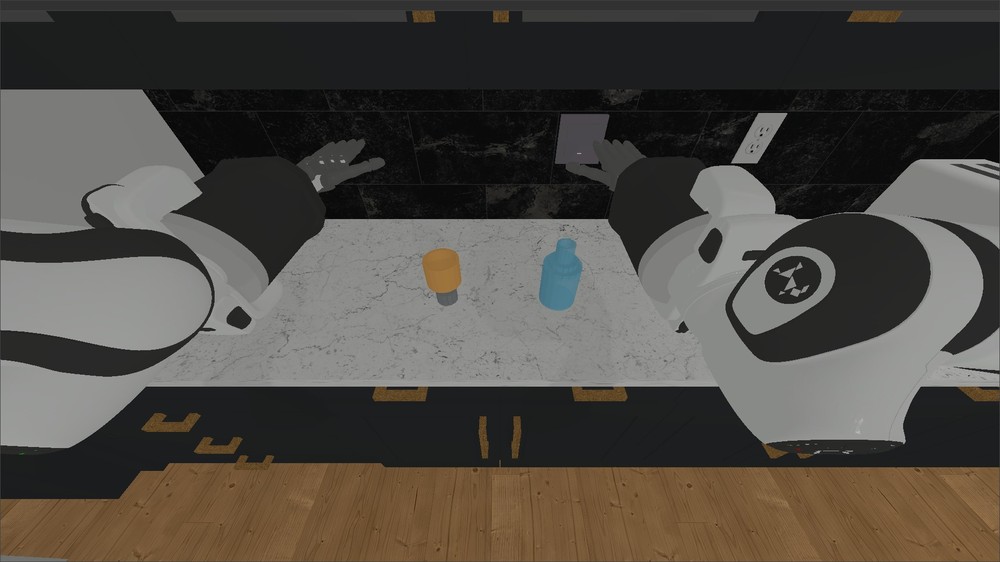}
\TaskLayoutStrip{PlugInsertion}{Core}{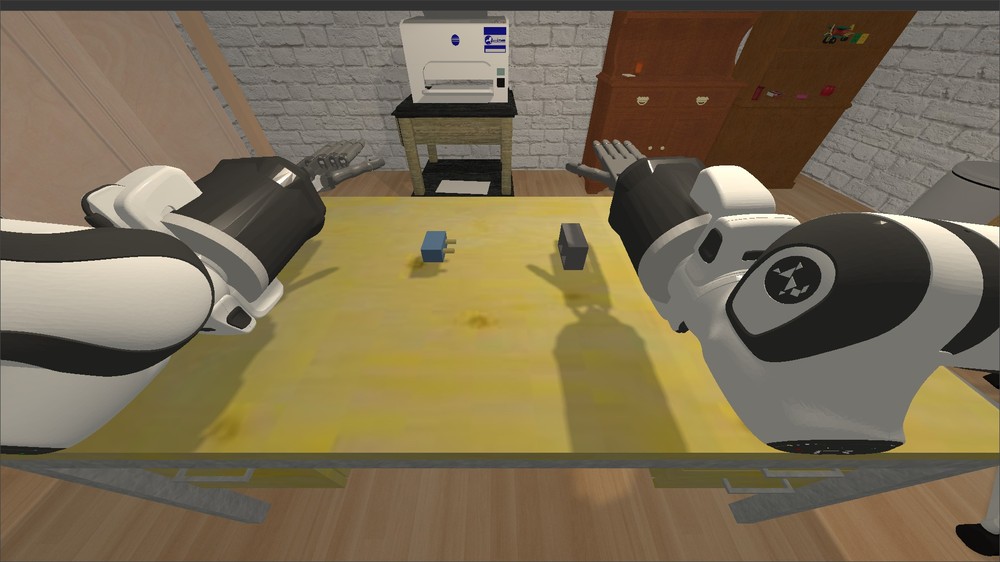}{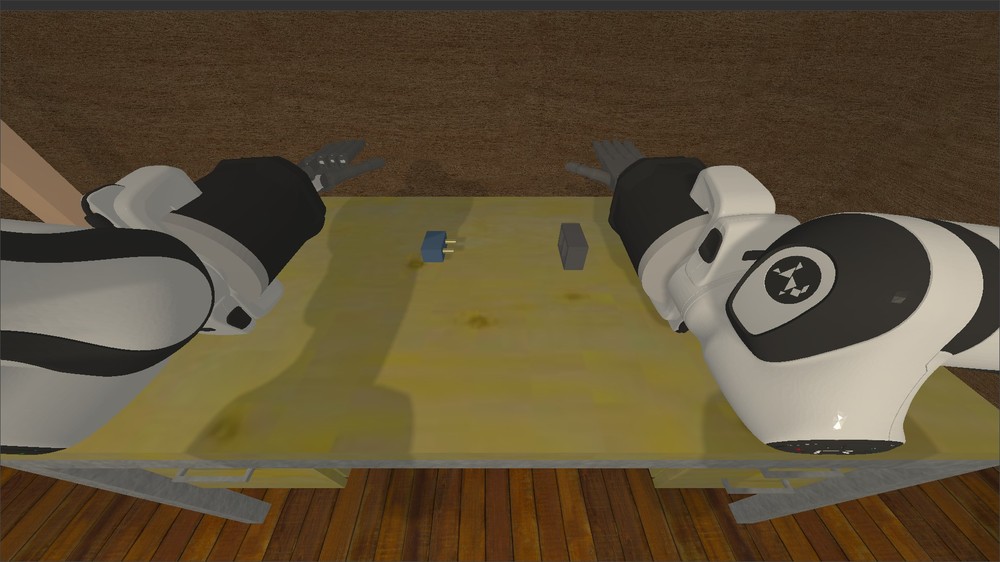}{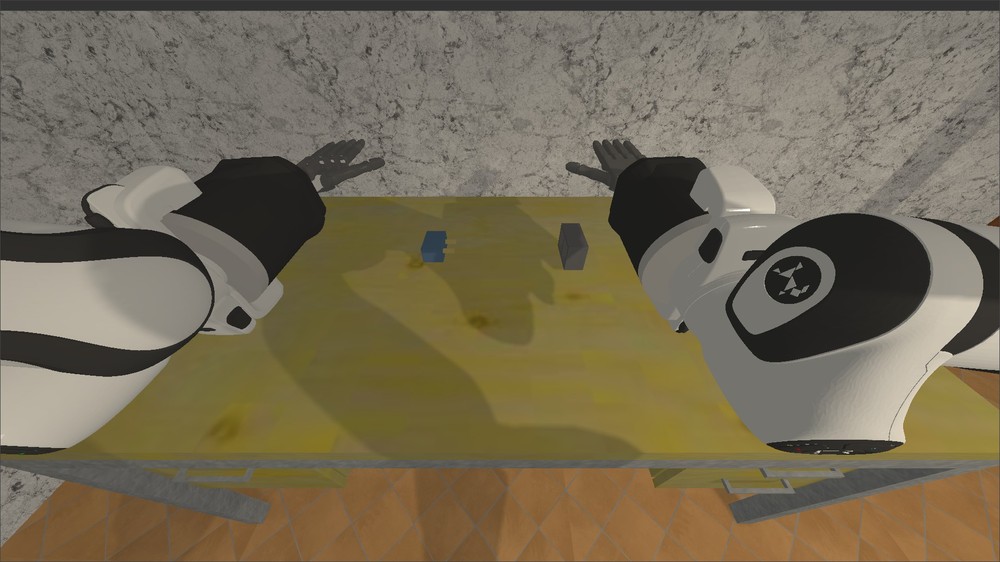}
\TaskLayoutStrip{ScissorsSheathing}{Gen}{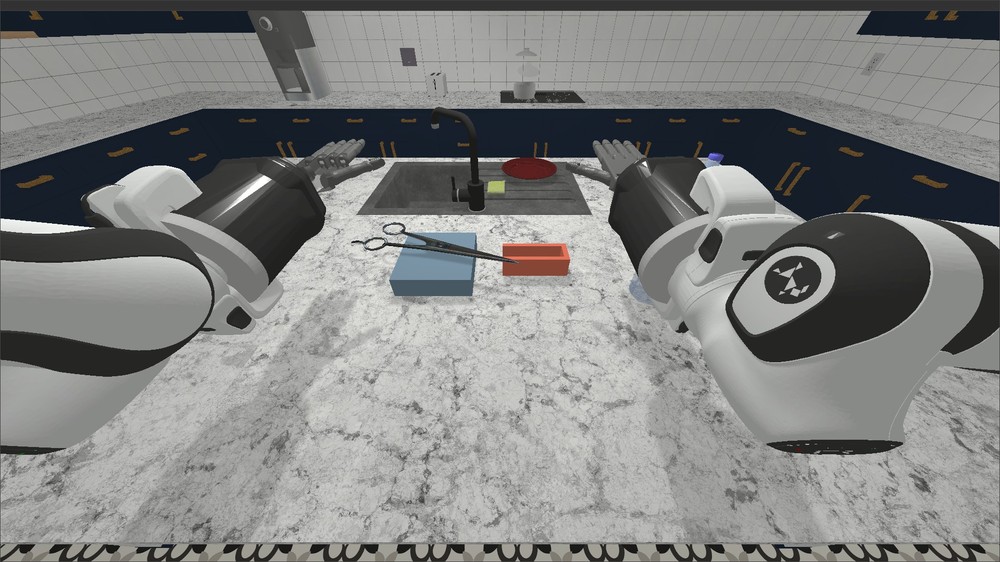}{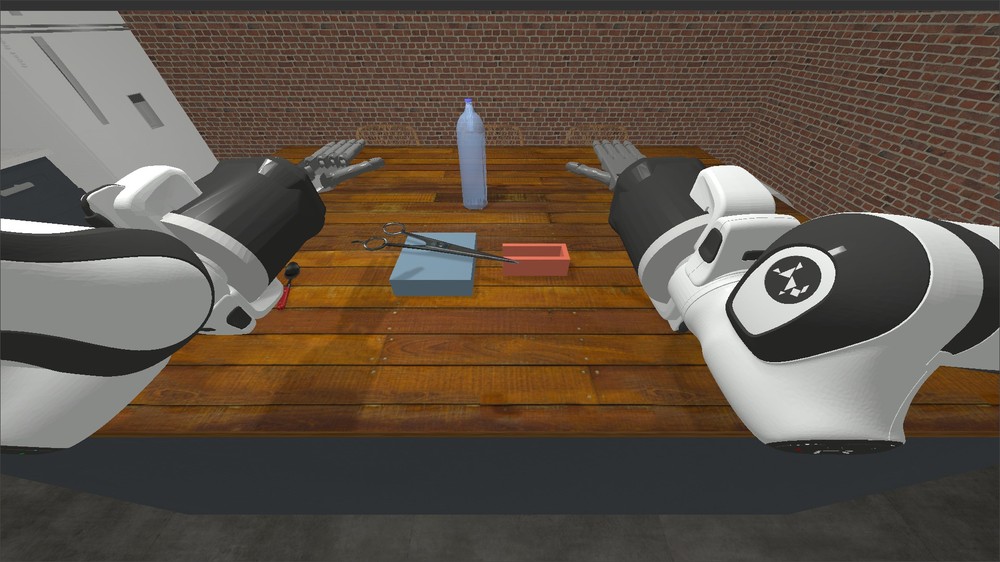}{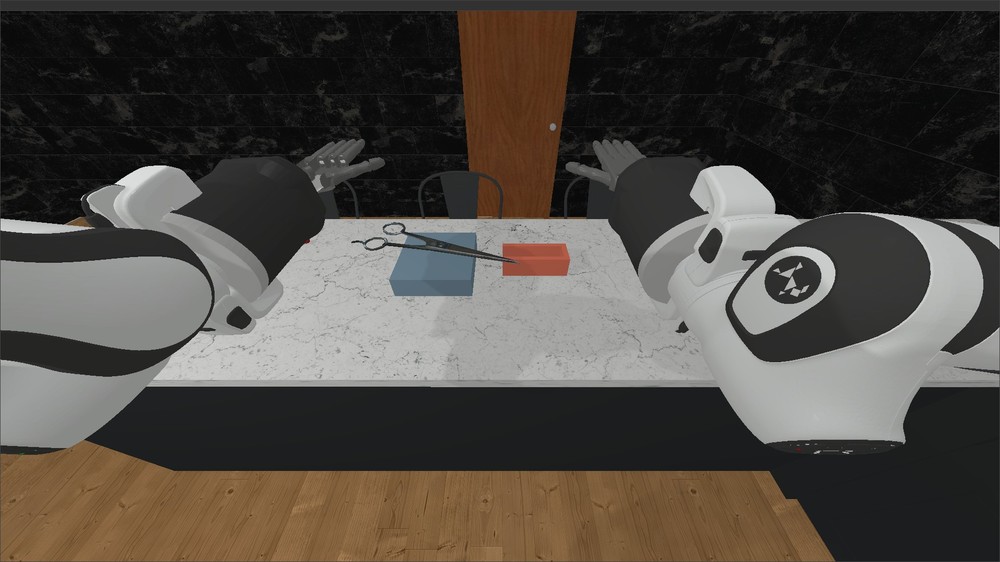}

\LayoutFamilyStart{Articulated configuration control}{LaptopClosing}{Core}{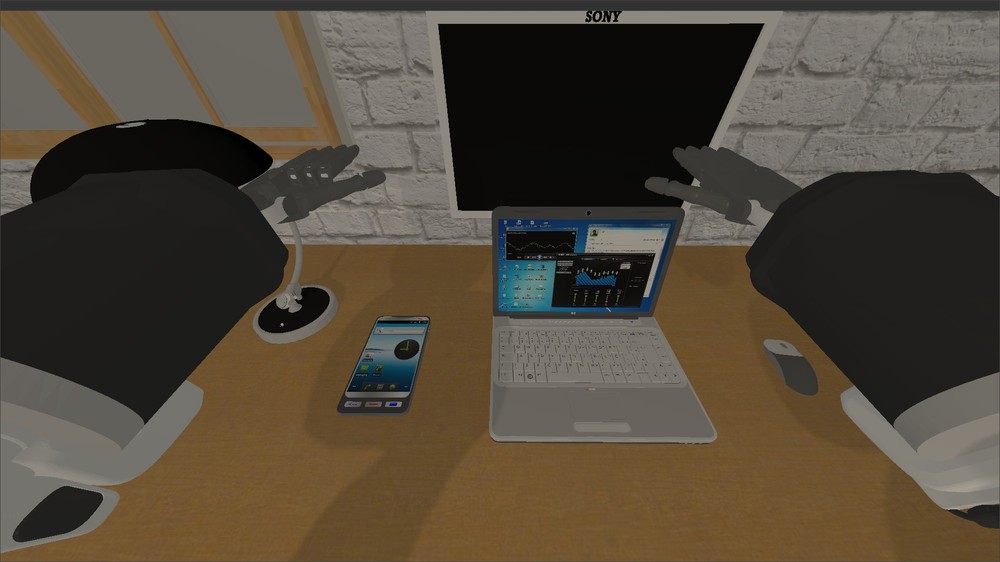}{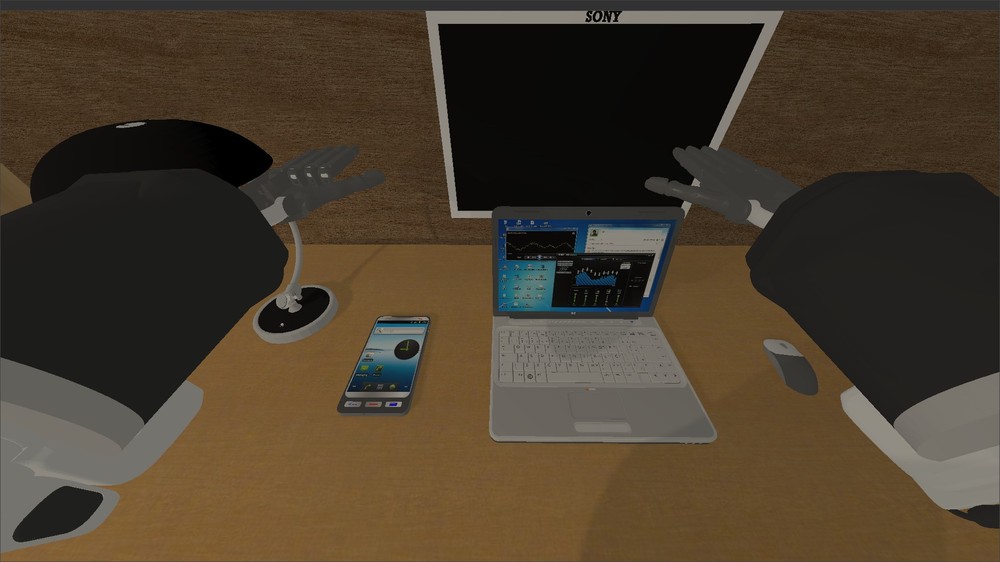}{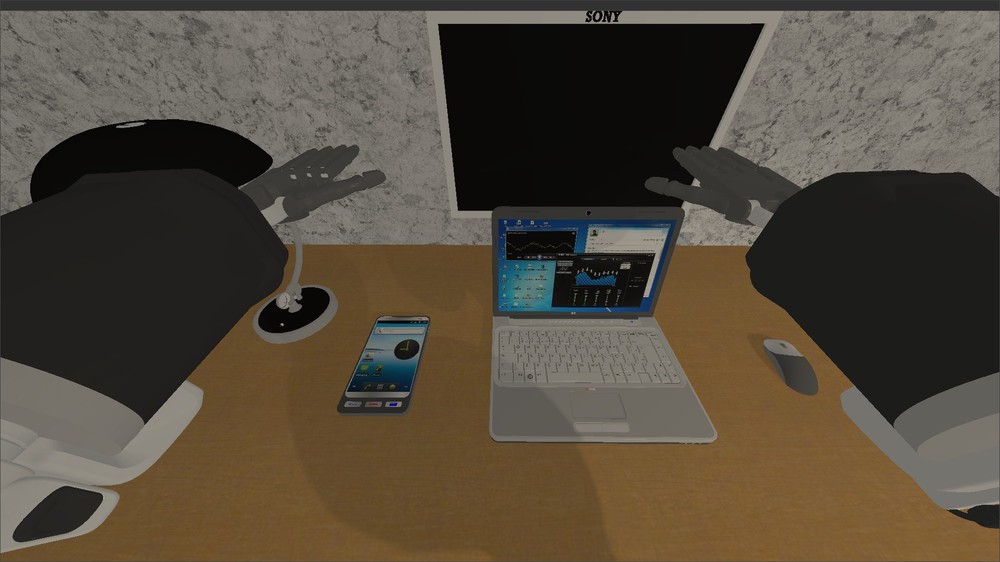}
\TaskLayoutStrip{StaplerPaperMark}{Core}{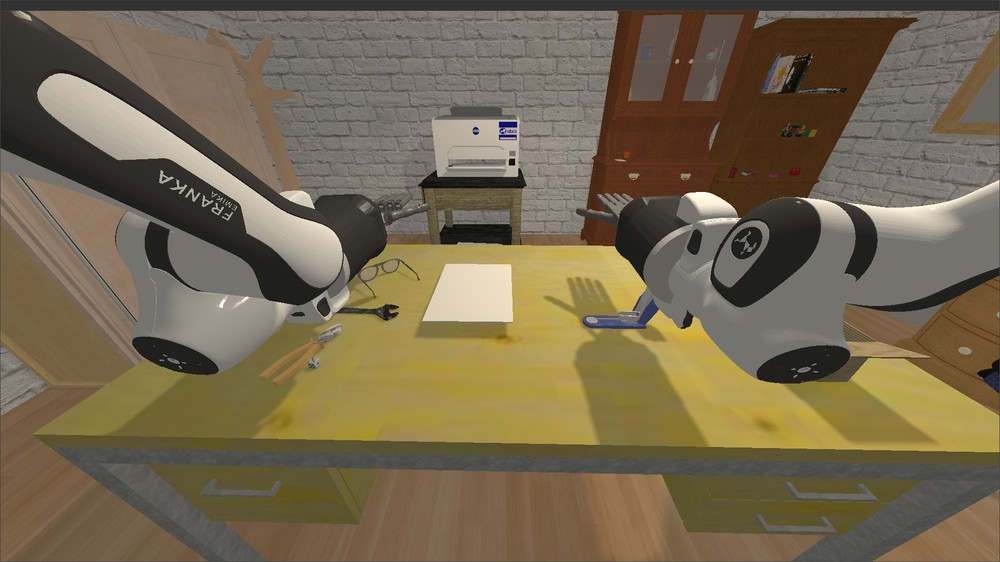}{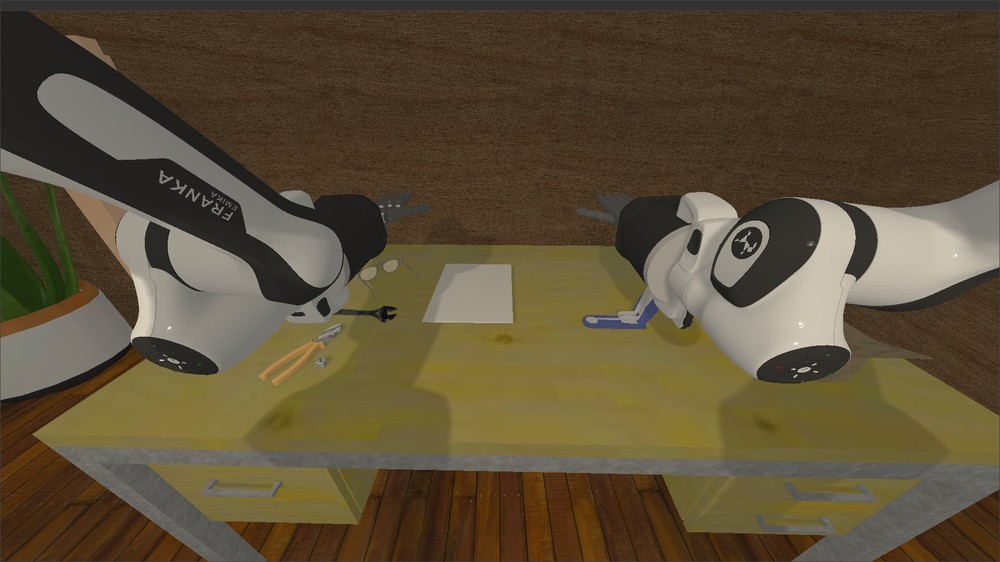}{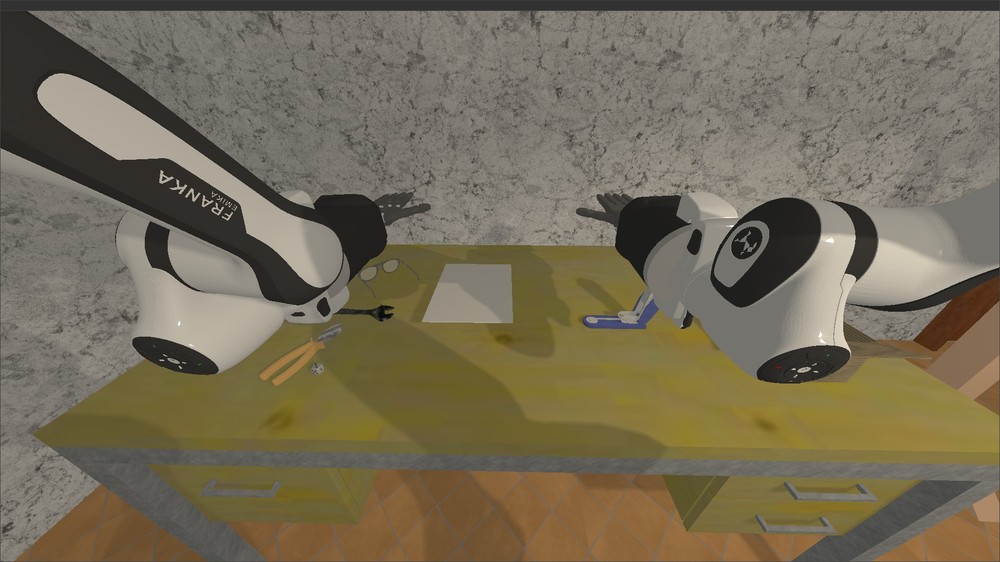}
\TaskLayoutStrip{ScissorsOpening}{Core}{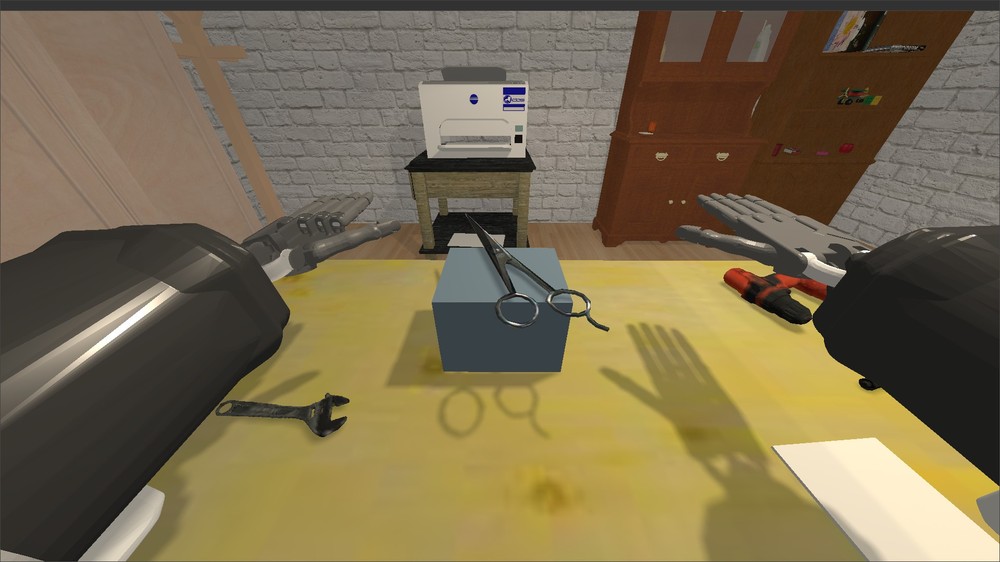}{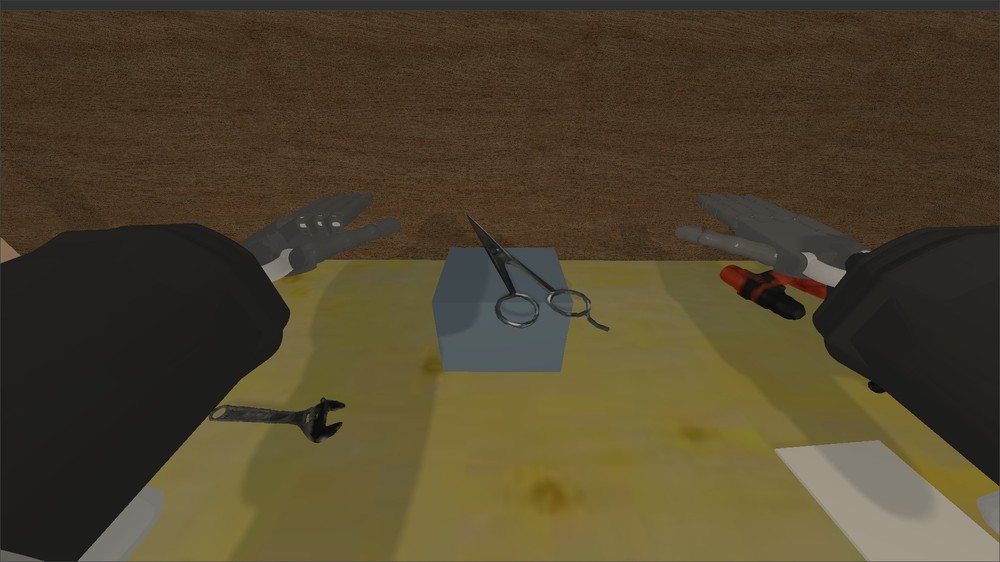}{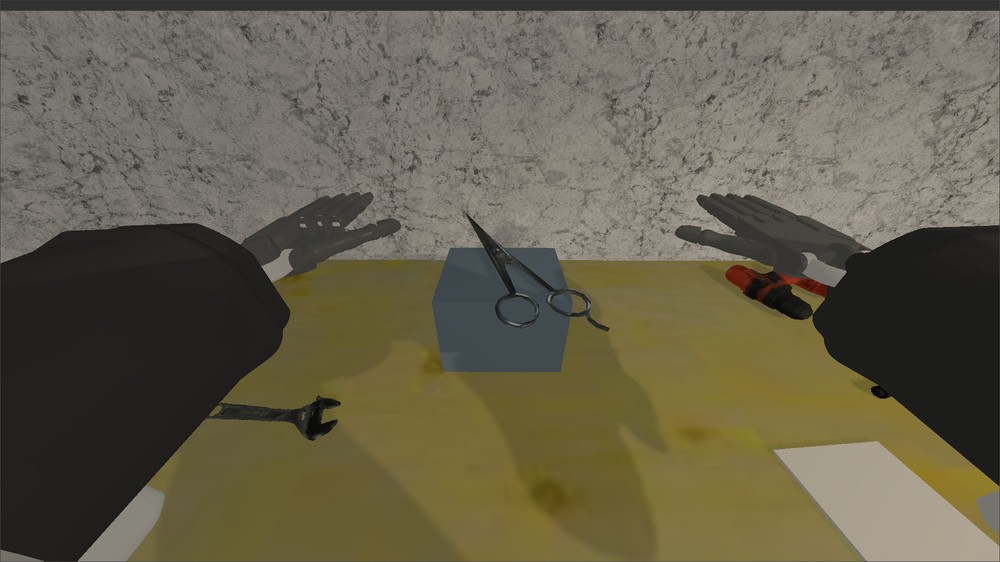}
\TaskLayoutStrip{BoxLidOpening}{Core}{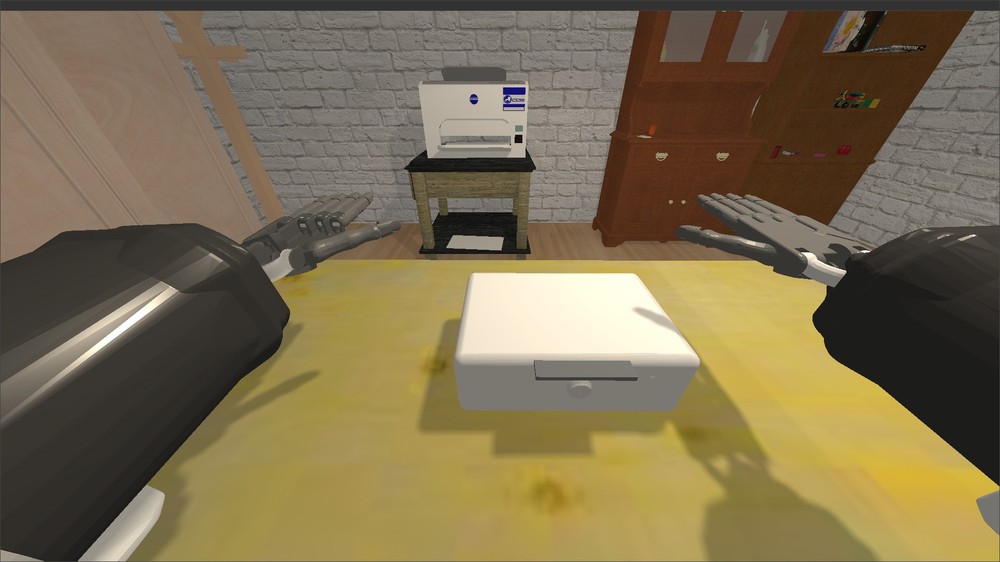}{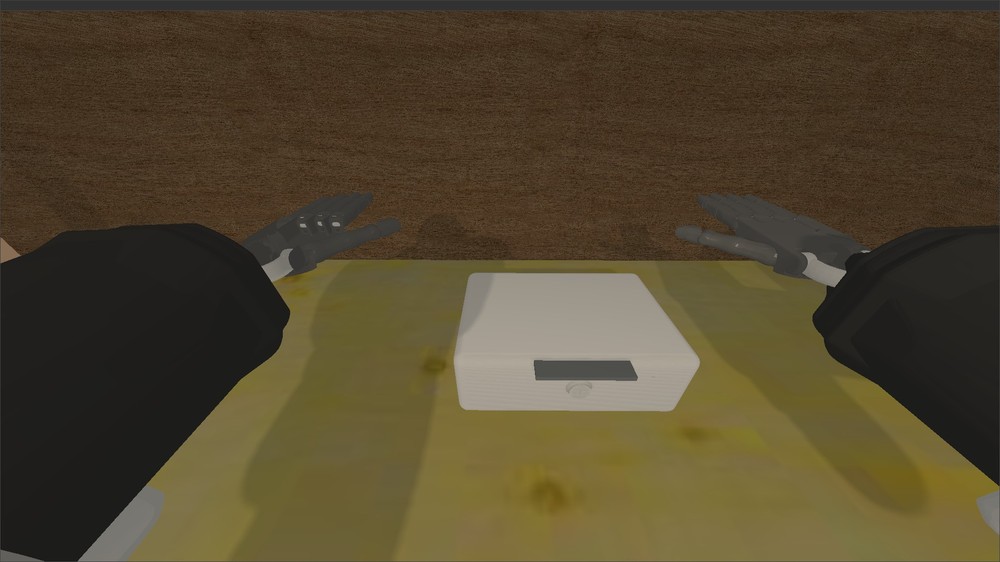}{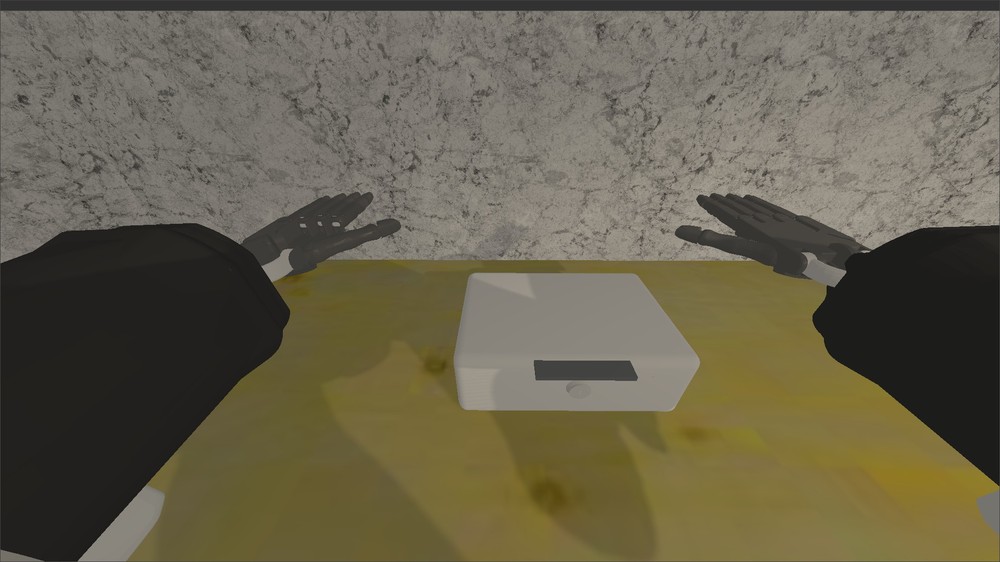}
\TaskLayoutStrip{GlassesFolding}{Gen}{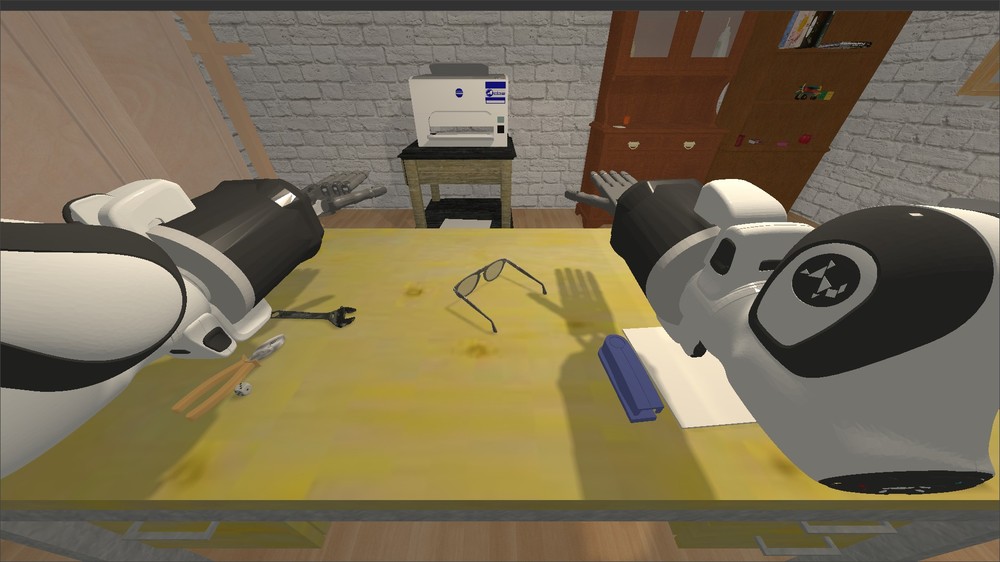}{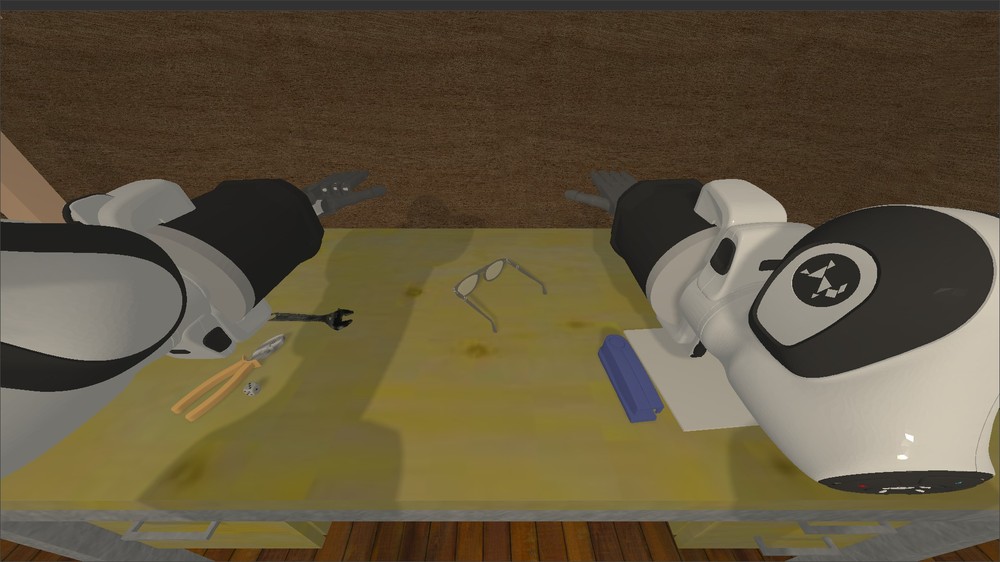}{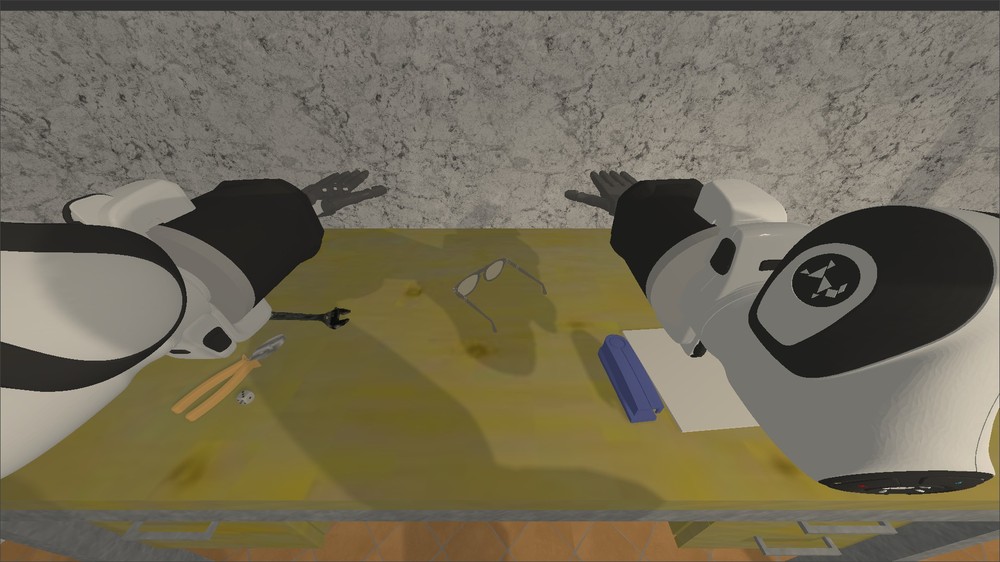}

\let\LayoutStripImage\undefined
\let\TaskLayoutStrip\undefined
\let\LayoutFamilyStart\undefined
\endgroup

\else

\newcommand{\LayoutImage}[1]{\includegraphics[width=0.174\textwidth]{figures/sceneLayout/#1}}
\newcommand{\LayoutHeader}[1]{\parbox{0.174\textwidth}{\centering\texttt{#1}}}

\begin{figure*}[p]
\centering
\begin{tabular}{@{}ccccc@{}}
\LayoutHeader{LampButton} & \LayoutHeader{WallSwitch} & \LayoutHeader{PhoneHomeButton} & \LayoutHeader{ToasterSlider} & \LayoutHeader{DisplayPowerButton}\\
\LayoutImage{LampButton_studio_l1.jpg} & \LayoutImage{WallSwitch_kitchen_l1.jpg} & \LayoutImage{PhoneHomeButton_studio_l1.jpg} & \LayoutImage{ToasterSlider_kitchen_l1.jpg} & \LayoutImage{DisplayPowerButton_studio_l1.jpg}\\
\LayoutImage{LampButton_studio_l2.jpg} & \LayoutImage{WallSwitch_kitchen_l2.jpg} & \LayoutImage{PhoneHomeButton_studio_l2.jpg} & \LayoutImage{ToasterSlider_kitchen_l2.jpg} & \LayoutImage{DisplayPowerButton_studio_l2.jpg}\\
\LayoutImage{LampButton_studio_l3.jpg} & \LayoutImage{WallSwitch_kitchen_l3.jpg} & \LayoutImage{PhoneHomeButton_studio_l3.jpg} & \LayoutImage{ToasterSlider_kitchen_l3.jpg} & \LayoutImage{DisplayPowerButton_studio_l3.jpg}
\end{tabular}
\caption{The three layouts for the Button-driven family.  The first four columns are Core tasks and the separated final column is the Gen task.}
\label{fig:layouts-button}
\end{figure*}

\begin{figure*}[p]
\centering
\begin{tabular}{@{}ccccc@{}}
\LayoutHeader{FaucetHandle} & \LayoutHeader{MouseWheel} & \LayoutHeader{CoffeeKnob} & \LayoutHeader{KettleLidTab} & \LayoutHeader{DishwasherKnob}\\
\LayoutImage{FaucetHandle_kitchen_l1.jpg} & \LayoutImage{MouseWheel_studio_l1.jpg} & \LayoutImage{CoffeeKnob_kitchen_l1.jpg} & \LayoutImage{KettleLidTab_kitchen_l1.jpg} & \LayoutImage{DishwasherKnob_kitchen_l1.jpg}\\
\LayoutImage{FaucetHandle_kitchen_l2.jpg} & \LayoutImage{MouseWheel_studio_l2.jpg} & \LayoutImage{CoffeeKnob_kitchen_l2.jpg} & \LayoutImage{KettleLidTab_kitchen_l2.jpg} & \LayoutImage{DishwasherKnob_kitchen_l2.jpg}\\
\LayoutImage{FaucetHandle_kitchen_l3.jpg} & \LayoutImage{MouseWheel_studio_l3.jpg} & \LayoutImage{CoffeeKnob_kitchen_l3.jpg} & \LayoutImage{KettleLidTab_kitchen_l3.jpg} & \LayoutImage{DishwasherKnob_kitchen_l3.jpg}
\end{tabular}
\caption{The three layouts for the Rotational manipulation family.}
\label{fig:layouts-rotational}
\end{figure*}

\begin{figure*}[p]
\centering
\begin{tabular}{@{}ccccc@{}}
\LayoutHeader{CupFromCabinet} & \LayoutHeader{MicrowaveBowl} & \LayoutHeader{TennisBallTransfer} & \LayoutHeader{BananaFromDrawer} & \LayoutHeader{CandleToCabinet}\\
\LayoutImage{CupFromCabinet_kitchen_l1.jpg} & \LayoutImage{MicrowaveBowl_kitchen_l1.jpg} & \LayoutImage{TennisBallTransfer_studio_l1.jpg} & \LayoutImage{BananaFromDrawer_studio_l1.jpg} & \LayoutImage{CandleToCabinet_studio_l1.jpg}\\
\LayoutImage{CupFromCabinet_kitchen_l2.jpg} & \LayoutImage{MicrowaveBowl_kitchen_l2.jpg} & \LayoutImage{TennisBallTransfer_studio_l2.jpg} & \LayoutImage{BananaFromDrawer_studio_l2.jpg} & \LayoutImage{CandleToCabinet_studio_l2.jpg}\\
\LayoutImage{CupFromCabinet_kitchen_l3.jpg} & \LayoutImage{MicrowaveBowl_kitchen_l3.jpg} & \LayoutImage{TennisBallTransfer_studio_l3.jpg} & \LayoutImage{BananaFromDrawer_studio_l3.jpg} & \LayoutImage{CandleToCabinet_studio_l3.jpg}
\end{tabular}
\caption{The three layouts for the Access-and-transfer family.}
\label{fig:layouts-access}
\end{figure*}

\begin{figure*}[p]
\centering
\begin{tabular}{@{}ccccc@{}}
\LayoutHeader{TablewarePlacement} & \LayoutHeader{HammerBlockAssembly} & \LayoutHeader{SandwichAssembly} & \LayoutHeader{RingTowerAssembly} & \LayoutHeader{BoxPacking}\\
\LayoutImage{TablewarePlacement_kitchen_l1.jpg} & \LayoutImage{HammerBlockAssembly_studio_l1.jpg} & \LayoutImage{SandwichAssembly_kitchen_l1.jpg} & \LayoutImage{RingTowerAssembly_studio_l1.jpg} & \LayoutImage{BoxPacking_kitchen_l1.jpg}\\
\LayoutImage{TablewarePlacement_kitchen_l2.jpg} & \LayoutImage{HammerBlockAssembly_studio_l2.jpg} & \LayoutImage{SandwichAssembly_kitchen_l2.jpg} & \LayoutImage{RingTowerAssembly_studio_l2.jpg} & \LayoutImage{BoxPacking_kitchen_l2.jpg}\\
\LayoutImage{TablewarePlacement_kitchen_l3.jpg} & \LayoutImage{HammerBlockAssembly_studio_l3.jpg} & \LayoutImage{SandwichAssembly_kitchen_l3.jpg} & \LayoutImage{RingTowerAssembly_studio_l3.jpg} & \LayoutImage{BoxPacking_kitchen_l3.jpg}
\end{tabular}
\caption{The three layouts for the Structured configuration family.}
\label{fig:layouts-structured}
\end{figure*}

\begin{figure*}[p]
\centering
\begin{tabular}{@{}ccccc@{}}
\LayoutHeader{PenCapping} & \LayoutHeader{CupSleeveFit} & \LayoutHeader{BottleFunnelInsertion} & \LayoutHeader{PlugInsertion} & \LayoutHeader{ScissorsSheathing}\\
\LayoutImage{PenCapping_studio_l1.jpg} & \LayoutImage{CupSleeveFit_kitchen_l1.jpg} & \LayoutImage{BottleFunnelInsertion_kitchen_l1.jpg} & \LayoutImage{PlugInsertion_studio_l1.jpg} & \LayoutImage{ScissorsSheathing_kitchen_l1.jpg}\\
\LayoutImage{PenCapping_studio_l2.jpg} & \LayoutImage{CupSleeveFit_kitchen_l2.jpg} & \LayoutImage{BottleFunnelInsertion_kitchen_l2.jpg} & \LayoutImage{PlugInsertion_studio_l2.jpg} & \LayoutImage{ScissorsSheathing_kitchen_l2.jpg}\\
\LayoutImage{PenCapping_studio_l3.jpg} & \LayoutImage{CupSleeveFit_kitchen_l3.jpg} & \LayoutImage{BottleFunnelInsertion_kitchen_l3.jpg} & \LayoutImage{PlugInsertion_studio_l3.jpg} & \LayoutImage{ScissorsSheathing_kitchen_l3.jpg}
\end{tabular}
\caption{The three layouts for the Assembly family.}
\label{fig:layouts-assembly}
\end{figure*}

\begin{figure*}[p]
\centering
\begin{tabular}{@{}ccccc@{}}
\LayoutHeader{LaptopClosing} & \LayoutHeader{StaplerPaperMark} & \LayoutHeader{ScissorsOpening} & \LayoutHeader{BoxLidOpening} & \LayoutHeader{GlassesFolding}\\
\LayoutImage{LaptopClosing_studio_l1.jpg} & \LayoutImage{StaplerPaperMark_studio_l1.jpg} & \LayoutImage{ScissorsOpening_studio_l1.jpg} & \LayoutImage{BoxLidOpening_studio_l1.jpg} & \LayoutImage{GlassesFolding_studio_l1.jpg}\\
\LayoutImage{LaptopClosing_studio_l2.jpg} & \LayoutImage{StaplerPaperMark_studio_l2.jpg} & \LayoutImage{ScissorsOpening_studio_l2.jpg} & \LayoutImage{BoxLidOpening_studio_l2.jpg} & \LayoutImage{GlassesFolding_studio_l2.jpg}\\
\LayoutImage{LaptopClosing_studio_l3.jpg} & \LayoutImage{StaplerPaperMark_studio_l3.jpg} & \LayoutImage{ScissorsOpening_studio_l3.jpg} & \LayoutImage{BoxLidOpening_studio_l3.jpg} & \LayoutImage{GlassesFolding_studio_l3.jpg}
\end{tabular}
\caption{The three layouts for the Articulated configuration control family.}
\label{fig:layouts-articulated}
\end{figure*}

\let\LayoutImage\undefined
\let\LayoutHeader\undefined

\fi
\ifSceneLayoutsByTaskRow\else\clearpage\fi

\subsection{Generalization axes}
\label{app:generalization-axes}

Apart from rotating each task through its three layouts across episodes, the evaluation protocol varies one factor at a time. A nominal episode applies no
continuous perturbation, and each generalized episode activates exactly one
of the following axes:
\begin{description}
\item[Camera.] The context camera is perturbed by up to $\pm7.5^\circ$ in
azimuth, $\pm5^\circ$ in elevation, $\pm10\%$ in viewing distance,
$\pm0.05$~m independently in each target coordinate, and $\pm5^\circ$ in
field of view.  The interaction camera uses the corresponding limits
$\pm5^\circ$, $\pm4^\circ$, $\pm8\%$, $\pm0.03$~m, and $\pm5^\circ$.
Wrist cameras remain rigidly attached to the wrists and are not independently
randomized.
\item[Lighting.] Ambient intensity is scaled in $[0.70,1.30]$ and direct-light
intensity in $[0.75,1.25]$.  A color parameter in $[-0.08,0.08]$ applies
coupled exponential red/blue gains while preserving non-negative RGB values
and the nominal light geometry.
\item[Physical.] Task-specific parameters alter quantities that affect contact
dynamics, such as mass, friction, damping, or joint properties.  Each task
declares valid bounds for the relevant objects; the sampled scene is rejected
if it violates task feasibility.
\item[Spatial.] Task-specific translational and rotational offsets alter object
poses and, where appropriate, robot placement or articulated initial states.
For evaluation, parameters whose names encode metric or angular offsets are
expanded to $1.5\times$ their collection range.
\end{description}

The nominal setting and the four generalization axes form five evaluation
conditions; the complete evaluation protocol is described in section~\ref{app:evaluation-protocol}.

\section{123-taxel tactile system}
\label{app:tactile}

\subsection{Taxel layout and observations}

Evo-DexBench places 123 virtual tactile taxels on each Shadow Hand. Each taxel is a finite rectangular patch attached to a parent link. It is described by its center, outward normal, in-plane tangent, and area, and is labeled by finger and anatomical region. Each taxel lies within approximately 1~mm of the visible hand surface.
The left-hand layout mirrors the calibrated right-hand layout. Both hands together therefore provide 246 channels, and corresponding channels on the two hands share the same anatomical meaning.

\Cref{tab:taxel-distribution} groups the taxels by hand part and by anatomical segment. There are 100 taxels on the fingers and 23 on the palm and sub-palm. The distal phalanges contain 62 taxels, about half of the total. This dense coverage captures fingertip and lateral contacts during button pressing, knob turning, and precision grasping. The palm taxels capture contacts during enveloping grasps and when
the palm provides stabilizing support.

\begin{table*}[ht]
\caption{Distribution of the 123 taxels on one Shadow Hand.}
\label{tab:taxel-distribution}
\centering
\small
\begin{tabular}{@{}lrrrrrrr@{}}
\toprule
Grouping & Thumb & Index & Middle & Ring & Little & Palm & Total \\
\midrule
By finger/area & 29 & 21 & 18 & 18 & 14 & 23 & 123 \\
\midrule
Region & Distal & Middle & Proximal & Palm & Sub-palm & & Total \\
By anatomy & 62 & 24 & 14 & 21 & 2 & & 123 \\
\bottomrule
\end{tabular}
\end{table*}

Each taxel reports one non-negative normal-force value in newtons.
The benchmark metadata versions the taxel geometry and channel
ordering used for training and evaluation.

\subsection{From contact impulses to tactile forces}

PhysX reports contact impulses at 120~Hz, while the policy actions and observations are recorded at 30~Hz. At each physics step, a contact is assigned to a taxel only if it (i) acts on the taxel’s parent link, (ii) falls within the rectangular patch expanded by a 1~mm assignment margin, (iii) lies within 10~mm of the patch plane, and (iv) is compressive with respect to the patch’s outward normal. If several patches satisfy these conditions, we distribute the contact’s vector impulse among them using spatial weights with finite support, so that patches closer to the contact receive larger weights.
The weights are normalized across the eligible patches, so the assigned impulses sum to the original vector impulse.

For each taxel, we sum the assigned impulses over the four physics
steps in a control period and divide by the full period duration to
obtain an average force vector. We then extract its compressive component
along the taxel normal and clamp negative values to zero. The resulting
observation is the average compressive normal force, in newtons, over
the policy's control period.

\subsection{Tactile inputs and force feedback in KineDex}

KineDex uses the 100 finger taxels on each hand as tactile input.
The thumb, index, middle, ring, and little fingers have 29, 21, 18,
18, and 14 channels, respectively. We zero-pad each finger to 29
channels, forming a $5\times29\times1$ tensor per hand for the shared
per-finger encoder. The remaining 23 palm and sub-palm channels are not fed to KineDex but remain available through the benchmark interface. At each time step, we sum the readings on each finger to obtain its scalar force target for
training.

Full KineDex outputs 30 benchmark motion values and five finger-force
targets per hand. At deployment, we clip the predicted targets to the
range observed during training. 
For each finger, the controller computes a flexion correction from the difference between the target and measured forces, using a gain of $0.005$~rad/N, and adds it to the corresponding motion command. The total correction is capped at $0.05$~rad per finger per control step. Finally, we clip the corrected joint commands to the environment’s joint limits.

Following the official KineDex ablation protocol, we train the No-Touch model
separately. It retains the tactile encoder but receives zeros during
both training and inference. Identity normalization keeps these
inputs at zero. The model predicts only the 30 motion values per hand, is trained without force targets, and is deployed without the force-feedback controller.
Both variants use the same visual inputs, proprioception,
demonstrations, backbone, optimizer, training budget, and evaluation
protocol. The comparison measures the combined effect of tactile
input, force supervision, and force feedback.

\section{Detailed data, training, and evaluation settings}
\label{app:training}

\subsection{Dataset and common policy contract}

Demonstrations are collected through the benchmark teleoperation interface and
stored in the LeRobot v3.0 format.  Each Core task
contains 50 successful trajectories and each Gen task contains eight.  At
30~Hz, a sample stores synchronized RGB observations, robot state, the command
applied by the demonstrator, the natural-language instruction, task-stage
state, and tactile measurements.  The simulator runs at
120~Hz, so four physics steps contribute to one policy observation and action.
Only successful trajectories are included in the training sets used in this
paper. In addition, to examine the soundness of the Core-Gen design, we collect 42 additional successful demonstrations for each of selected Gen tasks, bringing each of them to 50 demonstrations, the same number as each Core task. These additional trajectories are not part of the benchmark split and are used only in two supplementary studies. In the
Core--Gen role rotation (\cref{app:role-rotation}), DishwasherKnob
uses all 50 trajectories when it serves as Core, matching the other Core
tasks.  In the target-data
efficiency study (\cref{app:gen-data-efficiency}), the 4-, 8-, 16-, and
32-demonstration sets are nested subsets of these 50, and the
eight-demonstration set is identical to the benchmark Gen split.  

All policies use a common semantic action interface.  For a single arm--hand system,
the state contains a 7-D end-effector pose and 24 Shadow Hand joint positions
(31 dimensions), and the action contains a 6-D end-effector pose increment in
the robot-root frame followed by 24 absolute hand-joint targets (30
dimensions).  The bimanual interface concatenates the left and right vectors,
yielding 62-D state and 60-D action.  For single-hand tasks, visual input comprises context, interaction, and right-wrist RGB views. For bimanual tasks, it comprises context, left-wrist, and right-wrist views. The only exception is SmolVLA, which additionally retains the interaction view in bimanual tasks and thus uses four views.  Unless otherwise stated, state and action normalization statistics are
computed from the training data of each run.

\subsection{Policy implementations and adaptations}

We retain each method's released architecture, learning objective, and native
training implementation. All policies use the benchmark's LeRobot v3.0 data interface
and compact state/action convention. \Cref{tab:policy-integration} summarizes
the policy interfaces and compatibility adaptations.

\begin{table*}[ht]
\caption{Policy implementations and Evo-DexBench compatibility adaptations.}
\label{tab:policy-integration}
\centering
\small
\begin{tabular}{@{}T{0.10\linewidth}T{0.20\linewidth}T{0.28\linewidth}T{0.34\linewidth}@{}}
\toprule
Policy & Native model & Policy interface & Model-specific adaptation \\
\midrule
$\pi_{0.5}$ & OpenPI JAX; pretrained flow-matching VLA & Three RGB views, language, 31-D/62-D state; 50-step chunks & Pad state/action interfaces to 32-D/64-D and discard padded action dimensions \\
\midrule
SmolVLA & LeRobot 0.4.0; SmolVLM2-500M with flow-matching expert & Three/four RGB views, language, 31-D/62-D state; 50-step chunks & Pad to 32-D/64-D and expand four projection tensors for bimanual control \\
\midrule
QwenOFT & StarVLA; Qwen3-VL-4B-Instruct with an MLP action head & Three RGB views, language, 31-D/62-D state; 50-step chunks & Register the benchmark data and camera roles, use 30-D/60-D physical actions and q99 normalization; freeze the VLM \\
\midrule
DP-C & Convolutional Diffusion Policy with temporal U-Net & Two observation steps; 30-step external action horizon & Use a 32-step internal target for U-Net divisibility and discard the final two steps \\
\midrule
DP-T & Transformer Diffusion Policy & Two observation steps; 30-step action horizon & Map benchmark's compact state and action channels to the DP-T’s observation and action interface \\
\midrule
KineDex & Multi-view ResNet-18, per-finger tactile CNN, conditional U-Net & Two observation steps; 16-step horizon; motion plus five force targets per hand & Map 123 taxels to 100 finger taxels, expand the action head, and add bounded force feedback \\
\midrule
KineDex No-Touch & Backbone-matched KineDex variant & Visual/proprioceptive input; 30 motion values per hand & Feed zero tactile input and remove force supervision and force feedback \\
\bottomrule
\end{tabular}
\end{table*}

\paragraph{$\pi_{0.5}$ policy.}
All $\pi_{0.5}$ results in this paper use the official JAX implementation.
The adapter maps the three camera roles to OpenPI image keys, and prefixes the task
instruction with the benchmark control convention.  The native OpenPI padding
interface presents 32 dimensions for single-hand runs and 64 for bimanual runs;
only the semantic 30 or 60 action dimensions are executed.  Full model
parameters are optimized, and checkpoints are served directly from their JAX
Orbax representation.

For bimanual control, we concatenate the two 31-D states and the two 30-D
actions, and then append two padding entries to the state and four to the action, yielding 64-D vectors. We widen the pretrained action-input and action-output
projections, including the output bias, from 32 to 64 along their action axes.
We retain the original parameters in the first 32 positions and initialize the
additional positions for fine-tuning. At rollout, we take the first 60 predicted
action coordinates and split them into two 30-D arm--hand commands.

\paragraph{SmolVLA.}
We use the official lerobot/smolvla\_base implementation and retain its
SmolVLM2-500M-Video-Instruct backbone and flow-matching action expert.  The
vision encoder is frozen; the action expert and state projection are trained.
Single-hand runs consume the context, interaction, and right-wrist views with
31 state and 30 action dimensions. Only physical dimensions are normalized,
supervised, and executed.

For bimanual control, we add the left-wrist view and widen the state and
action interfaces from 32 to 64 dimensions. In the state-input and action-input
projections, we copy the pretrained weights for the 31 physical state
coordinates and 30 physical action coordinates into both arm--hand blocks,
scaling each input copy by $0.5$. In the action-output projection, we copy
the original 30 physical outputs, including their weights and biases, into
both output blocks at their original scale. The remaining padding entries
in the widened projections are zero. At rollout, the two 30-D output blocks
are executed as separate arm--hand commands.

\paragraph{QwenOFT.}
We use the official StarVLA QwenOFT implementation with the released
Qwen3-VL-4B-Instruct backbone.
The vision-language backbone is frozen, and only the MLP action head is
optimized. Each input contains a language instruction, robot state, and three
RGB views resized to $224\times224$: context, interaction, and right-wrist for
single-hand control; context, left-wrist, and right-wrist for bimanual control.
State/action dimensions are 31/30 and 62/60, respectively. The adapter preserves
the physical end-effector increments and absolute hand-joint targets and uses
StarVLA's q99 state/action normalization. The policy predicts 50 actions and
executes 10 before replanning.

\paragraph{DP-C and DP-T.}
Both variants use the official Stanford Diffusion Policy code path.  They share
the same observation and external action contract. At inference,
the models replan in receding-horizon fashion after executing eight predicted
steps.  When the training mixture changes, models are initialized from existing weights, while normalization statistics, optimizer state, and the
learning-rate schedule are reinitialized.

\paragraph{KineDex.}
The released KineDex architecture was originally designed for a different tactile array.  We configure its shared per-finger encoder according to the number of input channels and fingers, while preserving the visual encoder, diffusion backbone, and
training objective.  For the full variant (Full), the
per-hand action head is extended from the benchmark's 30 action dimensions to
35 by appending five force targets. The No-Touch variant follows the official
KineDex tactile-ablation protocol: the tactile input is set to zero, force-target supervision and force feedback are removed, and the model is trained as a separate model rather than having touch masked only at evaluation time. Full and No-Touch therefore use independently trained checkpoints.

\subsection{Optimization and transfer protocols}

\Cref{tab:training-hyperparameters} reports the settings used for the
paper results. QwenOFT's budget is specified in optimization steps,
while the other policies use epochs.
All training runs use seed 42.

\begin{table*}[ht]
\caption{Training hyperparameters. QwenOFT learning rates refer to the action head, and its batch size is global. KineDex No-Touch uses the same settings as Full KineDex.
$^{\dagger}$DisplayPowerButton and DishwasherKnob use 50 epochs; CandleToCabinet uses 100.
$^{\ddagger}$SmolVLA DisplayPowerButton, DishwasherKnob, and GlassesFolding use 100 epochs.}
\label{tab:training-hyperparameters}
\centering
\small
\begin{tabular}{@{}lrrrr@{}}
\toprule
Policy & Training budget & Peak LR & Minimum LR & Batch size \\
\midrule
$\pi_{0.5}$ & 25$^{\dagger}$ epochs & $5\!\times\!10^{-6}$ & $5\!\times\!10^{-7}$ & 32 \\
SmolVLA & 25$^{\ddagger}$ epochs & $1\!\times\!10^{-4}$ & $2.5\!\times\!10^{-6}$ & 32 \\
QwenOFT & 100{,}000 steps & $1\!\times\!10^{-4}$ & $5\!\times\!10^{-7}$ & 4 \\
DP-T & 600 epochs & $1\!\times\!10^{-4}$ & $1\!\times\!10^{-6}$ & 32 \\
DP-C & 600 epochs & $1\!\times\!10^{-4}$ & $1\!\times\!10^{-6}$ & 32 \\
KineDex & 600 epochs & $1\!\times\!10^{-4}$ & $1\!\times\!10^{-5}$ & 32 \\
\bottomrule
\end{tabular}
\end{table*}
For the standard $\pi_{0.5}$, SmolVLA, QwenOFT, DP-C, DP-T, and KineDex baselines, the Gen task contributes one half of each minibatch and the four Core tasks in its family
uniformly share the other half (12.5\% each).

The adaptation-strategy study fixes $\pi_{0.5}$, the target demonstrations, and
a 50-epoch target-stage budget.  \emph{Gen-only} trains from the released base
parameters using only the eight target demonstrations.  \emph{Core $\rightarrow$ Gen (final)} initializes target training from the terminal Core
checkpoint; \emph{Core $\rightarrow$ Gen (early)} instead uses the epoch-5 Core
checkpoint.  In both sequential conditions, target normalization and optimizer
state are reinitialized.  \emph{Core + Gen joint} trains on the 1:1 mixture for
the full schedule. The results are presented in the main text.

The Core-source relatedness study changes only the Core source paired with a
fixed Gen target.  Each mixed run maintains the 1:1 Core-Gen ratio and balances
the selected Core tasks within their half of the minibatch.  DisplayPowerButton
and DishwasherKnob use 50 epochs; CandleToCabinet uses 100.  The DishwasherKnob
replications with DP-T and KineDex follow the same source ratios.

\subsection{Evaluation protocol and uncertainty}
\label{app:evaluation-protocol}

\paragraph{Evaluation protocol.}
We use the scene layouts and generalization axes defined in
\cref{app:scenes}, with metrics as defined in {\ifdefined\SupplMainMetrics\SupplMainMetrics\else\cref{eq}\fi}.
Each task is evaluated over 100 episodes: 20 nominal episodes and 20 for each
of the camera, lighting, physical, and spatial variations. The three layouts of each task rotate across episodes and are assigned 34, 33, and 33 episodes, respectively, over the full evaluation.

Continuous parameters are sampled deterministically from seeds
$\{1001,1002,1003,1004,1005\}$. Within each generalized condition, every seed is used four times. All compared policies are evaluated under the same conditions, seeds, layouts, and reset schedule, so that evaluation sampling is controlled across policies.
An episode terminates on success or the registered task horizon. Model weights and normalization statistics are kept fixed during evaluation.

\paragraph{Uncertainty estimation.}
We estimate the uncertainty of the family-level and overall SRs shown in
{\ifdefined\SupplMainBaselineFigure\SupplMainBaselineFigure\else\cref{fig:baseline_performance}\fi} conditional on the evaluated policies
and task set. Let $\hat p_t$ be the observed SR for task $t$, estimated from $n=100$
episodes. For an equally weighted mean over $T$ fixed tasks, we use
\begin{equation}
  \bar p=\frac{1}{T}\sum_{t=1}^{T}\hat p_t,
  \qquad
  \mathrm{SE}(\bar p)=\frac{1}{T}
  \sqrt{\sum_{t=1}^{T}\frac{\hat p_t(1-\hat p_t)}{n}}.
\end{equation}
The 95\% normal-approximation interval is
$\bar p\pm1.96\,\mathrm{SE}(\bar p)$, truncated to $[0,1]$ and converted to
percentages for display.
\section{Detailed experimental results}
\label{app:results}

\providecommand{\Score}[2]{#1\, / \,#2}
\begingroup
\renewcommand{\Score}[2]{\makebox[2.65em][r]{#1}\,/\,\makebox[2.4em][l]{#2}}
\newcommand{\SupplDeltaPair}[2]{\makebox[1.8em][r]{#1}\,/\,\makebox[3.1em][l]{#2}}
\definecolor{supplgain}{RGB}{170,25,36}
\definecolor{supplloss}{RGB}{35,75,155}
\newcommand{\SupplGain}[1]{\textcolor{supplgain}{#1}}
\newcommand{\SupplLoss}[1]{\textcolor{supplloss}{#1}}

\subsection{Complete baseline evaluation results}

\Cref{tab:suppl-baseline-all} reports the SR and StageScore of all six policies on each of the 30 tasks, together with family, Core, Gen, and overall means.

\begin{table}[H]
    \caption{Baseline performance on all 30 tasks. Entries are SR (\%) / StageScore (0--100); $\dagger$ marks Gen tasks. Each mean weights the tasks in its group equally. Core and Gen means use 24 and 6 tasks, respectively. Bold indicates the highest value for each metric within a row.}
    \label{tab:suppl-baseline-all}
    \centering
    \small
    \setlength{\tabcolsep}{3pt}
    \renewcommand{\arraystretch}{0.96}
    \resizebox{\textwidth}{!}{%
    \begin{tabular}{@{}llcccccc@{}}
        \toprule
        Family & Task & $\pi_{0.5}$ & DP-T & DP-C & KineDex & SmolVLA & QwenOFT \\
        \midrule
        \multirow{6}{*}{\makecell[l]{Button-\\driven}} & LampButton & \Score{\textbf{49\%}}{\textbf{54.40}} & \Score{20\%}{23.90} & \Score{41\%}{47.60} & \Score{29\%}{35.60} & \Score{23\%}{29.30} & \Score{15\%}{18.60} \\
         & WallSwitch & \Score{44\%}{63.20} & \Score{31\%}{52.90} & \Score{\textbf{69\%}}{\textbf{81.90}} & \Score{63\%}{75.30} & \Score{42\%}{61.20} & \Score{66\%}{77.40} \\
         & PhoneHomeButton & \Score{\textbf{28\%}}{\textbf{33.70}} & \Score{5\%}{7.40} & \Score{10\%}{14.80} & \Score{16\%}{21.40} & \Score{15\%}{21.00} & \Score{3\%}{3.60} \\
         & ToasterSlider & \Score{41\%}{76.40} & \Score{44\%}{77.60} & \Score{60\%}{84.00} & \Score{\textbf{63\%}}{\textbf{85.20}} & \Score{34\%}{73.60} & \Score{18\%}{67.20} \\
         & DisplayPowerButton $^{\dagger}$ & \Score{20\%}{39.40} & \Score{\textbf{38\%}}{\textbf{56.60}} & \Score{11\%}{33.80} & \Score{15\%}{39.60} & \Score{18\%}{36.00} & \Score{1\%}{7.00} \\
         & \textit{Family mean} & \Score{36.4\%}{\textbf{53.42}} & \Score{27.6\%}{43.68} & \Score{\textbf{38.2\%}}{52.42} & \Score{37.2\%}{51.42} & \Score{26.4\%}{44.22} & \Score{20.6\%}{34.76} \\
        \midrule
        \multirow{6}{*}{\makecell[l]{Rotational\\manipulation}} & FaucetHandle & \Score{\textbf{38\%}}{\textbf{42.70}} & \Score{17\%}{20.80} & \Score{26\%}{30.80} & \Score{15\%}{19.80} & \Score{12\%}{22.20} & \Score{1\%}{10.00} \\
         & MouseWheel & \Score{\textbf{63\%}}{\textbf{81.50}} & \Score{42\%}{65.50} & \Score{43\%}{65.25} & \Score{57\%}{76.50} & \Score{62\%}{77.75} & \Score{17\%}{23.75} \\
         & CoffeeKnob & \Score{\textbf{52\%}}{\textbf{81.90}} & \Score{10\%}{49.65} & \Score{27\%}{66.55} & \Score{33\%}{69.40} & \Score{45\%}{72.60} & \Score{4\%}{34.55} \\
         & KettleLidTab & \Score{\textbf{86\%}}{\textbf{87.90}} & \Score{12\%}{49.00} & \Score{72\%}{82.70} & \Score{62\%}{75.25} & \Score{76\%}{84.10} & \Score{85\%}{85.80} \\
         & DishwasherKnob $^{\dagger}$ & \Score{46\%}{62.50} & \Score{43\%}{57.80} & \Score{\textbf{67\%}}{\textbf{75.10}} & \Score{55\%}{67.00} & \Score{42\%}{57.30} & \Score{21\%}{39.90} \\
         & \textit{Family mean} & \Score{\textbf{57.0\%}}{\textbf{71.30}} & \Score{24.8\%}{48.55} & \Score{47.0\%}{64.08} & \Score{44.4\%}{61.59} & \Score{47.4\%}{62.79} & \Score{25.6\%}{38.80} \\
        \midrule
        \multirow{6}{*}{\makecell[l]{Access-and-\\transfer}} & CupFromCabinet & \Score{\textbf{35\%}}{\textbf{63.00}} & \Score{3\%}{13.75} & \Score{1\%}{7.75} & \Score{1\%}{16.50} & \Score{0\%}{16.25} & \Score{0\%}{0.50} \\
         & MicrowaveBowl & \Score{\textbf{19\%}}{\textbf{37.50}} & \Score{0\%}{2.50} & \Score{0\%}{14.75} & \Score{2\%}{19.25} & \Score{1\%}{12.00} & \Score{0\%}{0.00} \\
         & TennisBallTransfer & \Score{\textbf{33\%}}{\textbf{54.95}} & \Score{4\%}{26.30} & \Score{3\%}{27.95} & \Score{10\%}{35.00} & \Score{9\%}{31.85} & \Score{0\%}{12.00} \\
         & BananaFromDrawer & \Score{\textbf{25\%}}{\textbf{53.35}} & \Score{0\%}{5.95} & \Score{0\%}{13.10} & \Score{2\%}{15.05} & \Score{1\%}{5.15} & \Score{0\%}{0.00} \\
         & CandleToCabinet $^{\dagger}$ & \Score{\textbf{3\%}}{\textbf{19.25}} & \Score{0\%}{0.75} & \Score{0\%}{16.50} & \Score{0\%}{3.25} & \Score{0\%}{1.50} & \Score{0\%}{0.25} \\
         & \textit{Family mean} & \Score{\textbf{23.0\%}}{\textbf{45.61}} & \Score{1.4\%}{9.85} & \Score{0.8\%}{16.01} & \Score{3.0\%}{17.81} & \Score{2.2\%}{13.35} & \Score{0.0\%}{2.55} \\
        \midrule
        \multirow{6}{*}{\makecell[l]{Structured\\configuration}} & TablewarePlacement & \Score{0\%}{9.90} & \Score{0\%}{2.10} & \Score{\textbf{2\%}}{\textbf{26.00}} & \Score{0\%}{3.30} & \Score{0\%}{12.30} & \Score{0\%}{0.00} \\
         & HammerBlockAssembly & \Score{\textbf{0\%}}{0.25} & \Score{\textbf{0\%}}{0.00} & \Score{\textbf{0\%}}{0.50} & \Score{\textbf{0\%}}{\textbf{0.75}} & \Score{\textbf{0\%}}{0.25} & \Score{\textbf{0\%}}{0.00} \\
         & SandwichAssembly & \Score{4\%}{18.80} & \Score{0\%}{8.40} & \Score{\textbf{9\%}}{\textbf{25.80}} & \Score{1\%}{11.40} & \Score{0\%}{7.20} & \Score{0\%}{0.00} \\
         & RingTowerAssembly & \Score{\textbf{0\%}}{\textbf{3.30}} & \Score{\textbf{0\%}}{0.30} & \Score{\textbf{0\%}}{1.20} & \Score{\textbf{0\%}}{1.50} & \Score{\textbf{0\%}}{0.00} & \Score{\textbf{0\%}}{0.00} \\
         & BoxPacking $^{\dagger}$ & \Score{\textbf{0\%}}{\textbf{2.40}} & \Score{\textbf{0\%}}{0.00} & \Score{\textbf{0\%}}{0.40} & \Score{\textbf{0\%}}{0.20} & \Score{\textbf{0\%}}{2.20} & \Score{\textbf{0\%}}{0.00} \\
         & \textit{Family mean} & \Score{0.8\%}{6.93} & \Score{0.0\%}{2.16} & \Score{\textbf{2.2\%}}{\textbf{10.78}} & \Score{0.2\%}{3.43} & \Score{0.0\%}{4.39} & \Score{0.0\%}{0.00} \\
        \midrule
        \multirow{6}{*}{\makecell[l]{Assembly}} & PenCapping & \Score{0\%}{\textbf{28.20}} & \Score{0\%}{15.00} & \Score{0\%}{22.20} & \Score{\textbf{1\%}}{25.20} & \Score{0\%}{19.60} & \Score{0\%}{10.80} \\
         & CupSleeveFit & \Score{3\%}{22.50} & \Score{0\%}{15.30} & \Score{2\%}{22.90} & \Score{\textbf{7\%}}{\textbf{28.00}} & \Score{0\%}{17.80} & \Score{0\%}{6.60} \\
         & BottleFunnelInsertion & \Score{2\%}{\textbf{37.20}} & \Score{0\%}{25.60} & \Score{\textbf{4\%}}{31.20} & \Score{2\%}{37.00} & \Score{\textbf{4\%}}{31.00} & \Score{0\%}{26.80} \\
         & PlugInsertion & \Score{7\%}{23.00} & \Score{4\%}{14.50} & \Score{\textbf{15\%}}{\textbf{35.50}} & \Score{8\%}{30.75} & \Score{3\%}{16.50} & \Score{0\%}{0.50} \\
         & ScissorsSheathing $^{\dagger}$ & \Score{1\%}{7.60} & \Score{1\%}{11.60} & \Score{\textbf{2\%}}{6.40} & \Score{0\%}{7.00} & \Score{0\%}{\textbf{14.00}} & \Score{1\%}{8.20} \\
         & \textit{Family mean} & \Score{2.6\%}{23.70} & \Score{1.0\%}{16.40} & \Score{\textbf{4.6\%}}{23.64} & \Score{3.6\%}{\textbf{25.59}} & \Score{1.4\%}{19.78} & \Score{0.2\%}{10.58} \\
        \midrule
        \multirow{6}{*}{\makecell[l]{Articulated\\configuration\\control}} & LaptopClosing & \Score{47\%}{60.50} & \Score{49\%}{59.20} & \Score{\textbf{85\%}}{\textbf{91.30}} & \Score{69\%}{75.90} & \Score{53\%}{59.90} & \Score{13\%}{27.70} \\
         & StaplerPaperMark & \Score{\textbf{42\%}}{\textbf{60.90}} & \Score{2\%}{26.90} & \Score{24\%}{50.60} & \Score{33\%}{54.70} & \Score{12\%}{37.00} & \Score{0\%}{22.60} \\
         & ScissorsOpening & \Score{23\%}{55.90} & \Score{6\%}{45.40} & \Score{\textbf{29\%}}{\textbf{58.60}} & \Score{26\%}{56.80} & \Score{15\%}{49.90} & \Score{0\%}{40.60} \\
         & BoxLidOpening & \Score{18\%}{41.10} & \Score{13\%}{39.10} & \Score{15\%}{39.60} & \Score{\textbf{26\%}}{\textbf{46.40}} & \Score{21\%}{40.80} & \Score{16\%}{43.30} \\
         & GlassesFolding $^{\dagger}$ & \Score{4\%}{14.00} & \Score{\textbf{6\%}}{16.80} & \Score{\textbf{6\%}}{\textbf{20.40}} & \Score{4\%}{16.40} & \Score{0\%}{5.20} & \Score{2\%}{10.00} \\
         & \textit{Family mean} & \Score{26.8\%}{46.48} & \Score{15.2\%}{37.48} & \Score{\textbf{31.8\%}}{\textbf{52.10}} & \Score{31.6\%}{50.04} & \Score{20.2\%}{38.56} & \Score{6.2\%}{28.84} \\
        \midrule
        \multicolumn{2}{l}{\textbf{Core mean}} & \Score{\textbf{27.5\%}}{\textbf{45.50}} & \Score{10.9\%}{26.96} & \Score{22.4\%}{39.27} & \Score{21.9\%}{38.16} & \Score{17.8\%}{33.30} & \Score{9.9\%}{21.35} \\
        \multicolumn{2}{l}{\textbf{Gen mean}} & \Score{12.3\%}{24.19} & \Score{\textbf{14.7\%}}{23.93} & \Score{14.3\%}{\textbf{25.43}} & \Score{12.3\%}{22.24} & \Score{10.0\%}{19.37} & \Score{4.2\%}{10.89} \\
        \multicolumn{2}{l}{\textbf{Overall mean}} & \Score{\textbf{24.4\%}}{\textbf{41.24}} & \Score{11.7\%}{26.35} & \Score{20.8\%}{36.51} & \Score{20.0\%}{34.98} & \Score{16.3\%}{30.52} & \Score{8.8\%}{19.26} \\
        \bottomrule
    \end{tabular}%
    }
\end{table}

\clearpage
\subsection{Complete Core-source relatedness results}

\Cref{tab:pi-relatedness-full} reports $\pi_{0.5}$ source conditions.  In each mixed condition, half of the training samples are drawn from the Gen demonstrations of the target and the other half from the Core demonstrations of the listed family.

In addition to the single-arm experiments in the main paper, we compare
Assembly and Articulated configuration control as Core sources for two
bimanual targets: ScissorsSheathing and GlassesFolding. For these bimanual targets, the cross-family comparison (\cref{tab:pi-relatedness-full}) is run at
50 epochs, and both the Core + Gen runs and the Gen-only reference use
this budget. The results in \cref{tab:suppl-baseline-all} use the default 25-epoch
budget (\cref{tab:training-hyperparameters}). For each bimanual target, the same-family source yields one more successful episode.

\begin{table*}[ht]
\caption{Complete $\pi_{0.5}$ Core-source relatedness results.  Entries are SR (\%) / StageScore; bold marks the strongest result for each target. $\dagger$ marks a Core source from the same task family as the Gen target. The last column shows changes relative to Gen-only training ($\Delta$SR in percentage points / $\Delta$StageScore).}
\label{tab:pi-relatedness-full}
\centering
\small
\setlength{\tabcolsep}{4pt}
\small
\begin{tabular}{@{}crlcc@{}}
\toprule
Gen target & Epochs & Training data & SR / StageScore & $\Delta$SR / $\Delta$StageScore \\
\midrule
\multirow{4}{*}{\makecell{DisplayPower\\Button}} & 50 & Gen-only & \Score{11\%}{29.70} & \textemdash \\
& 50 & Button-driven$^{\dagger}$ Core + Gen & \textbf{\Score{20\%}{39.40}} & \SupplDeltaPair{\SupplGain{+9}}{\SupplGain{+9.70}} \\
& 50 & Rotational manip. Core + Gen & \Score{17\%}{34.70} & \SupplDeltaPair{\SupplGain{+6}}{\SupplGain{+5.00}} \\
& 50 & Access-and-transfer Core + Gen & \Score{12\%}{33.00} & \SupplDeltaPair{\SupplGain{+1}}{\SupplGain{+3.30}} \\
\midrule
\multirow{4}{*}{\makecell{Dishwasher\\Knob}} & 50 & Gen-only & \Score{36\%}{55.10} & \textemdash \\
& 50 & Button-driven Core + Gen & \Score{40\%}{61.00} & \SupplDeltaPair{\SupplGain{+4}}{\SupplGain{+5.90}} \\
& 50 & Rotational manip.$^{\dagger}$ Core + Gen & \textbf{\Score{46\%}{62.50}} & \SupplDeltaPair{\SupplGain{+10}}{\SupplGain{+7.40}} \\
& 50 & Access-and-transfer Core + Gen & \Score{32\%}{52.40} & \SupplDeltaPair{\SupplLoss{-4}}{\SupplLoss{-2.70}} \\
\midrule
\multirow{4}{*}{\makecell{CandleTo\\Cabinet}} & 100 & Gen-only & \Score{0\%}{4.50} & \textemdash \\
& 100 & Button-driven Core + Gen & \Score{0\%}{7.75} & \SupplDeltaPair{0}{\SupplGain{+3.25}} \\
& 100 & Rotational manip. Core + Gen & \Score{1\%}{8.00} & \SupplDeltaPair{\SupplGain{+1}}{\SupplGain{+3.50}} \\
& 100 & Access-and-transfer$^{\dagger}$ Core + Gen & \textbf{\Score{3\%}{19.25}} & \SupplDeltaPair{\SupplGain{+3}}{\SupplGain{+14.75}} \\
\midrule
\multirow{3}{*}[\jot]{\makecell{Scissors\\Sheathing}}
 & 50 & Gen-only & \Score{1\%}{5.40} & \textemdash \\
 & 50 & Assembly$^{\dagger}$ Core + Gen & \textbf{\Score{9\%}{17.20}} & \SupplDeltaPair{\SupplGain{+8}}{\SupplGain{+11.80}} \\
 & 50 & \makecell[l]{Articulated control Core + Gen}
 & \Score{8\%}{17.00} & \SupplDeltaPair{\SupplGain{+7}}{\SupplGain{+11.60}} \\
\midrule
\multirow{3}{*}[\jot]{\makecell{Glasses\\Folding}}
 & 50 & Gen-only & \Score{1\%}{7.40} & \textemdash \\
 & 50 & Assembly Core + Gen & \Score{1\%}{\textbf{8.20}} & \SupplDeltaPair{0}{\SupplGain{+0.80}} \\
 & 50 & \makecell[l]{Articulated control$^{\dagger}$ Core + Gen}
 & \Score{\textbf{2\%}}{8.00} & \SupplDeltaPair{\SupplGain{+1}}{\SupplGain{+0.60}} \\
\bottomrule
\end{tabular}
\end{table*}

\subsection{Core-set breadth and composition}

For each Gen target, \cref{tab:core-breadth-composition} compares three Core
sources: all four Core tasks of the same family, and two two-task subsets
that split these four tasks into complementary pairs. All settings use the
same Gen demonstrations and epoch budget and differ only in the Core source.

\begin{table*}[ht]
\caption{Core-set breadth/composition ablation for $\pi_{0.5}$. Entries are SR (\%) / StageScore; bold marks the best result for each target. The last column shows changes relative to Gen-only training ($\Delta$SR in percentage points / $\Delta$StageScore).}
\label{tab:core-breadth-composition}
\centering
\small
\setlength{\tabcolsep}{4pt}
\begin{tabular}{@{}clcc@{}}
\toprule
Gen target & Training data & SR / StageScore & $\Delta$SR / $\Delta$StageScore \\
\midrule
\multirow{4}{*}{\makecell{DisplayPower\\Button}} & Gen-only & \Score{11\%}{29.70} & \textemdash \\
& LampButton + ToasterSlider + Gen & \Score{17\%}{32.90} & \SupplDeltaPair{\SupplGain{+6}}{\SupplGain{+3.20}} \\
& WallSwitch + PhoneHomeButton + Gen & \Score{\textbf{20\%}}{39.30} & \SupplDeltaPair{\SupplGain{+9}}{\SupplGain{+9.60}} \\
& \textbf{Four matched Core} + Gen & \textbf{\Score{20\%}{39.40}} & \SupplDeltaPair{\SupplGain{+9}}{\SupplGain{+9.70}} \\
\midrule
\multirow{4}{*}{\makecell{Dishwasher\\Knob}} & Gen-only & \Score{36\%}{55.10} & \textemdash \\
& MouseWheel + CoffeeKnob + Gen & \Score{35\%}{53.60} & \SupplDeltaPair{\SupplLoss{-1}}{\SupplLoss{-1.50}} \\
& FaucetHandle + KettleLidTab + Gen & \Score{37\%}{60.70} & \SupplDeltaPair{\SupplGain{+1}}{\SupplGain{+5.60}} \\
& \textbf{Four matched Core} + Gen & \textbf{\Score{46\%}{62.50}} & \SupplDeltaPair{\SupplGain{+10}}{\SupplGain{+7.40}} \\
\midrule
\multirow{4}{*}{\makecell{CandleTo\\Cabinet}} & Gen-only & \Score{0\%}{4.50} & \textemdash \\
& CupFromCabinet + BananaFromDrawer + Gen & \Score{2\%}{8.50} & \SupplDeltaPair{\SupplGain{+2}}{\SupplGain{+4.00}} \\
& MicrowaveBowl + TennisBallTransfer + Gen & \Score{1\%}{9.25} & \SupplDeltaPair{\SupplGain{+1}}{\SupplGain{+4.75}} \\
& \textbf{Four matched Core} + Gen & \textbf{\Score{3\%}{19.25}} & \SupplDeltaPair{\SupplGain{+3}}{\SupplGain{+14.75}} \\
\bottomrule
\end{tabular}
\end{table*}

\subsection{Rotation of Core and Gen roles}
\label{app:role-rotation}

We rotate the Gen role across all five tasks in the Rotational manipulation family. For each target, we compare Gen-only training with joint training that uses the other four tasks as Core under the same joint-training protocol. This comparison tests whether within-family transfer extends to tasks originally assigned to Core. Across the five targets, joint training raises mean SR from 36.4\% to 38.4\% and mean StageScore from 53.99 to 58.28. SR improves on three targets, is unchanged on one, and declines on one. StageScore improves on four targets. These results show that tasks originally assigned to Core can also benefit from within-family transfer, with gains varying by target.

\begin{table}[!ht]
\centering
\small
\caption{Role rotation in the Rotational manipulation family. Each task serves once as Gen, with the other four tasks serving as Core in joint training. Entries are SR (\%) / StageScore (0--100). $\dagger$ marks the original Gen task. The last column shows changes from Gen-only (SR in percentage points). The mean weights all five targets equally.}
\label{tab:core-gen-role-rotation}
\begin{tabular}{@{}lccc@{}}
\toprule
Gen target & Gen-only & Core + Gen joint & $\Delta$SR / $\Delta$StageScore \\
\midrule
FaucetHandle    & \Score{26\%}{36.20} & \Score{32\%}{45.80} & \SupplDeltaPair{\SupplGain{+6}}{\SupplGain{+9.60}} \\
MouseWheel      & \Score{44\%}{62.50} & \Score{28\%}{54.50} & \SupplDeltaPair{\SupplLoss{-16}}{\SupplLoss{-8.00}} \\
CoffeeKnob      & \Score{12\%}{49.70} & \Score{12\%}{51.40} & \SupplDeltaPair{0}{\SupplGain{+1.70}} \\
KettleLidTab    & \Score{64\%}{66.45} & \Score{74\%}{77.20} & \SupplDeltaPair{\SupplGain{+10}}{\SupplGain{+10.75}} \\
DishwasherKnob $^{\dagger}$ & \Score{36\%}{55.10} & \Score{46\%}{62.50} & \SupplDeltaPair{\SupplGain{+10}}{\SupplGain{+7.40}} \\
\midrule
\textbf{Mean}   & \Score{36.4\%}{53.99} & \Score{38.4\%}{58.28} & \SupplDeltaPair{\SupplGain{+2.0}}{\SupplGain{+4.29}} \\
\bottomrule
\end{tabular}
\end{table}

\subsection{Complete target-data efficiency results}
\label{app:gen-data-efficiency}

Table~\ref{tab:gen-data-efficiency-full} gives the per-task results underlying
\Cref{fig:suppl-data-efficiency}. For each of the five Gen targets other than
CandleToCabinet from the Access-and-transfer family, $\pi_{0.5}$
is trained with 4, 8, 16, 32, or 50 target demonstrations. The mixed setting
uses eight target demonstrations and 200 demonstrations from the four matched
Core tasks, sampled at a Core:Gen ratio of 1:1. All settings are evaluated for 100 episodes per task, and the mean uses the same five targets.

\begin{table}[ht]
\caption{Target-data efficiency of $\pi_{0.5}$. Entries are SR (\%) / StageScore (0--100). Means weight the five tasks equally.}
\label{tab:gen-data-efficiency-full}
\centering
\footnotesize
\setlength{\tabcolsep}{3pt}
\small
\resizebox{\textwidth}{!}{%
\begin{tabular}{@{}lcccccc@{}}
\toprule
Task & 4 Gen & 8 Gen & 16 Gen & 32 Gen & 50 Gen & Core + 8 Gen \\
\midrule
DisplayPowerButton & \Score{2\%}{20.30} & \Score{11\%}{29.70} & \Score{15\%}{31.70} & \Score{20\%}{42.50} & \Score{29\%}{51.50} & \Score{20\%}{39.40} \\
DishwasherKnob & \Score{19\%}{37.60} & \Score{36\%}{55.10} & \Score{39\%}{58.20} & \Score{45\%}{65.10} & \Score{72\%}{80.70} & \Score{46\%}{62.50} \\
BoxPacking & \Score{0\%}{0.00} & \Score{0\%}{2.00} & \Score{0\%}{4.40} & \Score{0\%}{8.60} & \Score{0\%}{8.55} & \Score{0\%}{2.40} \\
ScissorsSheathing & \Score{0\%}{1.60} & \Score{1\%}{5.40} & \Score{2\%}{11.20} & \Score{1\%}{10.20} & \Score{3\%}{18.00} & \Score{1\%}{7.60} \\
GlassesFolding & \Score{1\%}{6.20} & \Score{1\%}{7.40} & \Score{2\%}{8.20} & \Score{6\%}{24.80} & \Score{8\%}{30.00} & \Score{4\%}{14.00} \\
\midrule
\textbf{Mean (5 tasks)} & \Score{4.4\%}{13.14} & \Score{9.8\%}{19.92} & \Score{11.6\%}{22.74} & \Score{14.4\%}{30.24} & \Score{22.4\%}{37.75} & \Score{14.2\%}{25.18} \\
\bottomrule
\end{tabular}%
}
\end{table}

\begin{figure}[ht]
\centering
\includegraphics[width=\linewidth]{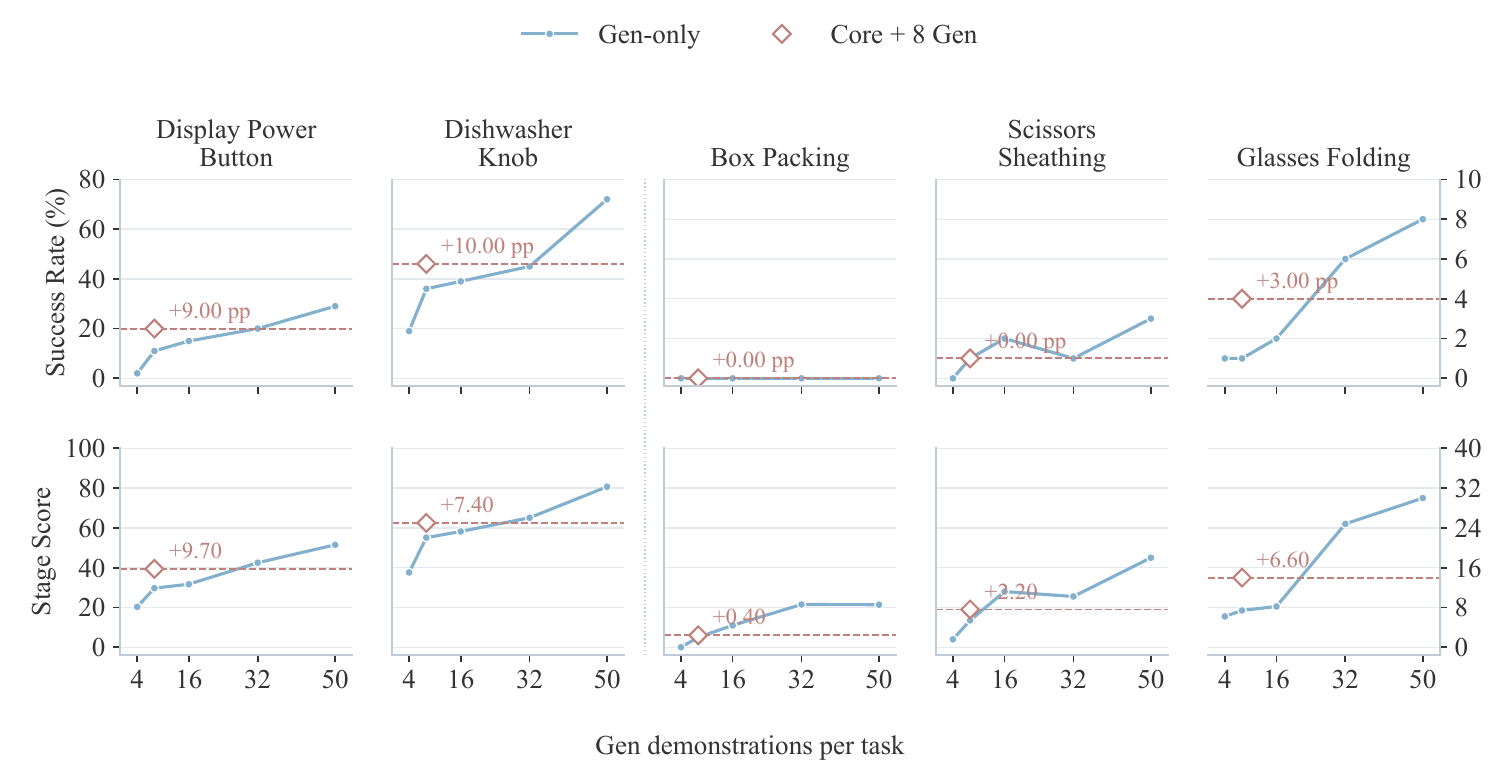}
\caption{Target-data efficiency of $\pi_{0.5}$ for each of the five Gen targets. The top row shows SR and the bottom row shows StageScore. Each column compares Gen-only training at five demonstration budgets with matched Core training plus eight Gen demonstrations.}
\label{fig:suppl-data-efficiency}
\end{figure}

\subsection{Complete tactile ablation results}

\Cref{tab:tactile-taskwise} reports paired Full and No-Touch KineDex results on all 30 tasks, including family and overall means. Differences are Full minus No-Touch, with positive values favoring Full. 

\begingroup
\small
\begin{longtable}{@{}lccrr@{}}
\caption{Complete KineDex tactile ablation. Entries are SR (\%) / StageScore (0--100). Differences are Full minus No-Touch. $\Delta$SR is in percentage points.}\label{tab:tactile-taskwise}\\
\toprule
Task & No-Touch & Full & $\Delta$SR & $\Delta$StageScore \\
\midrule
\endfirsthead
\multicolumn{5}{@{}l}{\textbf{Table~\thetable{} continued from the previous page}}\\
\toprule
Task & No-Touch & Full & $\Delta$SR & $\Delta$StageScore \\
\midrule
\endhead
\midrule
\multicolumn{5}{r@{}}{Continued on the next page}\\
\endfoot
\bottomrule
\endlastfoot
\multicolumn{5}{@{}l}{\textit{Button-driven}} \\
LampButton & \Score{31\%}{31.40} & \Score{29\%}{35.60} & \SupplLoss{-2.00} & \SupplGain{+4.20} \\
WallSwitch & \Score{54\%}{69.60} & \Score{63\%}{75.30} & \SupplGain{+9.00} & \SupplGain{+5.70} \\
PhoneHomeButton & \Score{10\%}{18.70} & \Score{16\%}{21.40} & \SupplGain{+6.00} & \SupplGain{+2.70} \\
ToasterSlider & \Score{61\%}{82.80} & \Score{63\%}{85.20} & \SupplGain{+2.00} & \SupplGain{+2.40} \\
DisplayPowerButton & \Score{19\%}{37.00} & \Score{15\%}{39.60} & \SupplLoss{-4.00} & \SupplGain{+2.60} \\
\textit{Family mean} & \Score{35.0\%}{47.90} & \Score{37.2\%}{51.42} & \SupplGain{+2.20} & \SupplGain{+3.52} \\
\midrule
\multicolumn{5}{@{}l}{\textit{Rotational manipulation}} \\
FaucetHandle & \Score{15\%}{16.30} & \Score{15\%}{19.80} & 0.00 & \SupplGain{+3.50} \\
MouseWheel & \Score{46\%}{67.00} & \Score{57\%}{76.50} & \SupplGain{+11.00} & \SupplGain{+9.50} \\
CoffeeKnob & \Score{25\%}{63.90} & \Score{33\%}{69.40} & \SupplGain{+8.00} & \SupplGain{+5.50} \\
KettleLidTab & \Score{54\%}{70.70} & \Score{62\%}{75.25} & \SupplGain{+8.00} & \SupplGain{+4.55} \\
DishwasherKnob & \Score{60\%}{73.50} & \Score{55\%}{67.00} & \SupplLoss{-5.00} & \SupplLoss{-6.50} \\
\textit{Family mean} & \Score{40.0\%}{58.28} & \Score{44.4\%}{61.59} & \SupplGain{+4.40} & \SupplGain{+3.31} \\
\midrule
\multicolumn{5}{@{}l}{\textit{Access-and-transfer}} \\
CupFromCabinet & \Score{2\%}{17.00} & \Score{1\%}{16.50} & \SupplLoss{-1.00} & \SupplLoss{-0.50} \\
MicrowaveBowl & \Score{0\%}{11.00} & \Score{2\%}{19.25} & \SupplGain{+2.00} & \SupplGain{+8.25} \\
TennisBallTransfer & \Score{8\%}{31.25} & \Score{10\%}{35.00} & \SupplGain{+2.00} & \SupplGain{+3.75} \\
BananaFromDrawer & \Score{0\%}{9.45} & \Score{2\%}{15.05} & \SupplGain{+2.00} & \SupplGain{+5.60} \\
CandleToCabinet & \Score{2\%}{11.25} & \Score{0\%}{3.25} & \SupplLoss{-2.00} & \SupplLoss{-8.00} \\
\textit{Family mean} & \Score{2.4\%}{15.99} & \Score{3.0\%}{17.81} & \SupplGain{+0.60} & \SupplGain{+1.82} \\
\midrule
\multicolumn{5}{@{}l}{\textit{Structured configuration}} \\
TablewarePlacement & \Score{0\%}{2.40} & \Score{0\%}{3.30} & 0.00 & \SupplGain{+0.90} \\
HammerBlockAssembly & \Score{0\%}{0.50} & \Score{0\%}{0.75} & 0.00 & \SupplGain{+0.25} \\
SandwichAssembly & \Score{1\%}{13.40} & \Score{1\%}{11.40} & 0.00 & \SupplLoss{-2.00} \\
RingTowerAssembly & \Score{0\%}{0.90} & \Score{0\%}{1.50} & 0.00 & \SupplGain{+0.60} \\
BoxPacking & \Score{0\%}{0.00} & \Score{0\%}{0.20} & 0.00 & \SupplGain{+0.20} \\
\textit{Family mean} & \Score{0.2\%}{3.44} & \Score{0.2\%}{3.43} & 0.00 & \SupplLoss{-0.01} \\
\midrule
\multicolumn{5}{@{}l}{\textit{Assembly}} \\
PenCapping & \Score{1\%}{24.40} & \Score{1\%}{25.20} & 0.00 & \SupplGain{+0.80} \\
CupSleeveFit & \Score{8\%}{31.10} & \Score{7\%}{28.00} & \SupplLoss{-1.00} & \SupplLoss{-3.10} \\
BottleFunnelInsertion & \Score{1\%}{35.80} & \Score{2\%}{37.00} & \SupplGain{+1.00} & \SupplGain{+1.20} \\
PlugInsertion & \Score{6\%}{25.75} & \Score{8\%}{30.75} & \SupplGain{+2.00} & \SupplGain{+5.00} \\
ScissorsSheathing & \Score{0\%}{3.80} & \Score{0\%}{7.00} & 0.00 & \SupplGain{+3.20} \\
\textit{Family mean} & \Score{3.2\%}{24.17} & \Score{3.6\%}{25.59} & \SupplGain{+0.40} & \SupplGain{+1.42} \\
\midrule
\multicolumn{5}{@{}l}{\textit{Articulated configuration control}} \\
LaptopClosing & \Score{72\%}{77.70} & \Score{69\%}{75.90} & \SupplLoss{-3.00} & \SupplLoss{-1.80} \\
StaplerPaperMark & \Score{18\%}{47.10} & \Score{33\%}{54.70} & \SupplGain{+15.00} & \SupplGain{+7.60} \\
ScissorsOpening & \Score{36\%}{62.80} & \Score{26\%}{56.80} & \SupplLoss{-10.00} & \SupplLoss{-6.00} \\
BoxLidOpening & \Score{18\%}{39.90} & \Score{26\%}{46.40} & \SupplGain{+8.00} & \SupplGain{+6.50} \\
GlassesFolding & \Score{3\%}{5.00} & \Score{4\%}{16.40} & \SupplGain{+1.00} & \SupplGain{+11.40} \\
\textit{Family mean} & \Score{29.4\%}{46.50} & \Score{31.6\%}{50.04} & \SupplGain{+2.20} & \SupplGain{+3.54} \\
\midrule
\textbf{Overall mean} & \Score{18.4\%}{32.71} & \Score{20.0\%}{34.98} & \SupplGain{+1.63} & \SupplGain{+2.27} \\
\end{longtable}
\endgroup

\endgroup

\end{document}